%% file: ClinMPO_NBE_revised.tex
\documentclass[pdflatex,sn-nature]{sn-jnl}

\usepackage[T1]{fontenc}
\usepackage{newtxtext,newtxmath}
\usepackage{graphicx}
\usepackage{amsmath}
\usepackage{booktabs}
\usepackage{longtable}
\usepackage{tabularx}
\usepackage{array}
\usepackage{ragged2e}
\usepackage{multirow}
\usepackage{calc}
\usepackage{needspace}
\usepackage{setspace}
\usepackage{makecell}
\usepackage{microtype}
\usepackage{titlesec}
\usepackage{caption}
\usepackage{etoolbox}
\usepackage{xcolor}
\expandafter\def\csname suppnum@supp-fig-baseline\endcsname{1}
\expandafter\def\csname suppnum@supp-fig-clinmpo\endcsname{2}
\expandafter\def\csname suppnum@supp-fig-dataset-construction\endcsname{3}
\expandafter\def\csname suppnum@supp-fig-student-interface\endcsname{4}
\expandafter\def\csname suppnum@supp-fig-student-evaluation\endcsname{5}
\expandafter\def\csname suppnum@supp-fig-clinician-evaluation\endcsname{6}
\expandafter\def\csname suppnum@supp-fig-clinician-model-a-interface\endcsname{7}
\expandafter\def\csname suppnum@supp-fig-clinician-model-b-interface\endcsname{8}
\expandafter\def\csname suppnum@supp-fig-clinrm-validation\endcsname{9}
\expandafter\def\csname suppnum@supp-fig-category-effects\endcsname{10}
\expandafter\def\csname suppnum@supp-fig-training-category-heatmaps\endcsname{11}
\expandafter\def\csname suppnum@supp-fig-test-set-category-cross-count\endcsname{12}
\expandafter\def\csname suppnum@supp-fig-training-set-category-cross-count\endcsname{13}
\expandafter\def\csname suppnum@supp-tab-provenance\endcsname{1}
\expandafter\def\csname suppnum@supp-tab-cpts\endcsname{2}
\expandafter\def\csname suppnum@supp-tab-expert-interrater-reliability\endcsname{3}
\expandafter\def\csname suppnum@supp-tab-training-configuration\endcsname{4}
\expandafter\def\csname suppnum@supp-tab-competencies\endcsname{5}
\expandafter\def\csname suppnum@supp-tab-icd11\endcsname{6}
\expandafter\def\csname suppnum@supp-tab-all-model-accuracy\endcsname{7}
\expandafter\def\csname suppnum@supp-tab-unplotted-categories\endcsname{8}
\expandafter\def\csname suppnum@supp-tab-unplotted-category-accuracy\endcsname{9}
\expandafter\def\csname suppnum@supp-tab-pre-split-model-accuracy\endcsname{10}

\newcommand{\suppfigref}[1]{Supplementary Fig.~\csname suppnum@#1\endcsname}
\newcommand{\supptableref}[1]{Supplementary Table~\csname suppnum@#1\endcsname}
\newcommand{\suppnoteref}[2]{Supplementary Note~#2}
\newcommand{\clinmporepo}{\url{https://github.com/PAI-CUHK/ClinMPO}}

\titleformat{\section}{\fontsize{14}{18}\selectfont\bfseries}{}{0pt}{}
\titleformat{\subsection}{\normalsize\bfseries}{}{0pt}{}
\titleformat{\subsubsection}{\normalsize\bfseries\itshape}{}{0pt}{}
\titlespacing*{\section}{0pt}{36pt}{2pt}
\titlespacing*{\subsection}{0pt}{18pt}{6pt}
\titlespacing*{\subsubsection}{0pt}{18pt}{6pt}

\DeclareCaptionLabelSeparator{pipe}{\enspace|\enspace}
\DeclareCaptionFont{referencefigure}{\fontsize{10}{12}\selectfont}
\renewcommand{\figurename}{Fig.}

\makeatletter
\def\keywords#1{\ifx#1\empty\else\def\@keywords{\par\addvspace{3pt}{\keywordfont{\bfseries\keywordname:} #1\par}}\fi}
\patchcmd{\@maketitle}{\else\vskip21pt\fi}{\else\vskip6pt\fi}{}{}
\patchcmd{\@maketitle}{\removelastskip\vskip36pt\vskip0pt}{\removelastskip\vskip0pt}{}{}
\patchcmd{\@maketitle}
  {\ifequalcont{\par$^{\dagger}$\@equalconttext\par}\fi
   \removelastskip\vskip24pt}
  {\ifequalcont{\par$^{\dagger}$\@equalconttext\par}\fi
   \par$^{\ddagger}$Work done during an internship at The Chinese University of Hong Kong.\par
   \removelastskip\vskip8pt}
  {}{}
\patchcmd{\@maketitle}{\removelastskip\vskip20pt\nointerlineskip}{\removelastskip\vskip12pt\nointerlineskip}{}{}
\patchcmd{\@maketitle}{\vskip20pt%
        \artauthors}{\vskip12pt%
        \artauthors}{}{}
\patchcmd{\@maketitle}{{\vskip7pt\addressfont\auaddress}}{{\vskip4pt\addressfont\auaddress}}{}{}
\def\Titlefont{\reset@font\fontsize{17bp}{22.5bp}\selectfont\titraggedcenter}
\def\Authorfont{\reset@font\fontsize{11bp}{13bp}\selectfont\boldmath\titraggedcenter}
\def\addressfont{\reset@font\fontsize{9bp}{10.3bp}\selectfont\titraggedcenter}
\makeatother

\DeclareGraphicsExtensions{.pdf,.png,.jpg,.jpeg}

\begin{document}

\title[Evidence-guided reinforcement learning]{An evidence-guided reinforcement learning method to improve psychiatric reasoning in small language models}

\makeatletter
\expandafter\def\csname X@internship\endcsname{\ensuremath{\ddagger}}
\makeatother

\author[1]{\fnm{Xinxin} \sur{Lin}}
\equalcont{These authors contributed equally to this work.}

\author[2,3]{\fnm{Guangxin} \sur{Dai}}
\equalcont{These authors contributed equally to this work.}

\author[4]{\fnm{Yi} \sur{Zhong}}
\equalcont{These authors contributed equally to this work.}

\author[1]{\fnm{Xiang} \sur{Li}}
\equalcont{These authors contributed equally to this work.}

\author[5]{\fnm{Xue} \sur{Xiao}}
\author[1]{\fnm{Jian} \sur{Liu}}
\author[6]{\fnm{Yixin} \sur{Zhang}}
\author[7]{\fnm{Lingming} \sur{Hu}}
\author[1,8]{\fnm{Zhengdong} \sur{Wu}}
\author[7,9]{\fnm{Yongbo} \sur{Zheng}}
\author[10,internship]{\fnm{Runchuan} \sur{Zhu}}
\author[11,internship]{\fnm{Ming} \sur{Zhao}}
\author[12]{\fnm{Huizi} \sur{Yu}}
\author[9]{\fnm{Yi} \sur{Zhang}}
\author[13]{\fnm{Fangting} \sur{Lu}}
\author[2,3]{\fnm{Shuo} \sur{Wu}}
\author[2,3]{\fnm{Jun} \sur{Zhao}}
\author[14]{\fnm{Ping} \sur{Yin}}
\author[1,8]{\fnm{Joey W. Y.} \sur{Chan}}
\author[1,8]{\fnm{Ngan Yin} \sur{Chan}}
\author[7]{\fnm{Yumei} \sur{Wang}}
\author[15]{\fnm{Lejin} \sur{Yang}}
\author[16]{\fnm{Yanqiu} \sur{Xing}}
\author[1,8]{\fnm{Sijing} \sur{Chen}}

\author*[1,8,17,18]{\fnm{Yun Kwok} \sur{Wing}}
\email{ykwing@cuhk.edu.hk}

\author*[4,9,19]{\fnm{Lin} \sur{Lu}}
\email{linlu@bjmu.edu.cn}

\author*[2,3]{\fnm{Xin} \sur{Ma}}
\email{maxin@sdu.edu.cn}

\author*[1,17]{\fnm{Lizhou} \sur{Fan}}
\email{leofan@cuhk.edu.hk}

\affil[1]{%
  \orgname{Department of Psychiatry, The Chinese University of Hong Kong},
  \orgaddress{\country{Hong Kong SAR, China}}}

\affil[2]{%
  \orgname{School of Control Science and Engineering, Shandong University},
  \orgaddress{\city{Jinan}, \state{Shandong}, \country{China}}}

\affil[3]{%
  \orgname{Key Engineering Research Center of Intelligent Unmanned System of Ministry of Education, Shandong University},
  \orgaddress{\city{Jinan}, \state{Shandong}, \country{China}}}

\affil[4]{%
  \orgname{Peking University Sixth Hospital, Peking University Institute of Mental Health, NHC Key Laboratory of Mental Health (Peking University), National Center for Mental Disorders, National Clinical Research Center for Mental Disorders (Peking University Sixth Hospital), Peking University},
  \orgaddress{\city{Beijing}, \country{China}}}

\affil[5]{%
  \orgname{Inspur Cloud Information Technology Co., Ltd.},
  \orgaddress{\city{Jinan}, \country{China}}}

\affil[6]{%
  \orgname{Department of Medicine, Shanghai Medical College Fudan University},
  \orgaddress{\city{Shanghai}, \country{China}}}

\affil[7]{%
  \orgname{Department of Psychology, Shandong Provincial Hospital Affiliated to Shandong First Medical University},
  \orgaddress{\city{Jinan}, \country{China}}}

\affil[8]{%
  \orgname{Li Chiu Kong Family Sleep Assessment Unit, Faculty of Medicine, The Chinese University of Hong Kong},
  \orgaddress{\country{Hong Kong SAR, China}}}

\affil[9]{%
  \orgname{Institute of Brain Science and Brain-inspired Research, Shandong First Medical University and Shandong Academy of Medical Sciences},
  \orgaddress{\city{Jinan}, \state{Shandong}, \country{China}}}

\affil[10]{%
  \orgname{Department of Electrical and Computer Engineering, National University of Singapore},
  \orgaddress{\city{Singapore}, \country{Singapore}}}

\affil[11]{%
  \orgname{School of Intelligence Science and Technology, Nanjing University},
  \orgaddress{\city{Suzhou, Jiangsu}, \country{China}}}

\affil[12]{%
  \orgname{Department of Medicine and Therapeutics, The Chinese University of Hong Kong},
  \orgaddress{\country{Hong Kong SAR, China}}}

\affil[13]{%
  \orgname{Department of Neurology, The First Affiliated Hospital of Anhui Medical University, Anhui Medical University},
  \orgaddress{\city{Hefei}, \postcode{230022}, \country{China}}}

\affil[14]{%
  \orgname{Hong Kong ICI Cloud Service Limited},
  \orgaddress{\country{Hong Kong SAR, China}}}

\affil[15]{%
  \orgname{Department of Psychology, Qilu Hospital of Shandong University},
  \orgaddress{\city{Jinan}, \country{China}}}

\affil[16]{%
  \orgname{Department of Geriatric Medicine, Qilu Hospital of Shandong University},
  \orgaddress{\city{Jinan}, \state{Shandong}, \country{China}}}

\affil[17]{%
  \orgname{Li Ka Shing Institute of Health Sciences, Faculty of Medicine, The Chinese University of Hong Kong},
  \orgaddress{\country{Hong Kong SAR, China}}}

\affil[18]{%
  \orgname{Gerald Choa Neuroscience Institute, The Chinese University of Hong Kong},
  \orgaddress{\country{Hong Kong SAR, China}}}

\affil[19]{%
  \orgname{National Institute on Drug Dependence and Beijing Key Laboratory of Innovative Intervention Techniques and Translational Research for Mental Disorders Rehabilitation, Peking University},
  \orgaddress{\city{Beijing}, \country{China}}}

\newcommand{\manuscriptabstract}{Privacy and computational constraints limit the use of large language models in psychiatry, while adapting small language models (SLMs) often requires substantial data and expert annotation. We developed ClinMPO, an evidence-guided reinforcement-learning framework guided by the psychiatrist-defined Clinical Psychiatry Thinking Strategy (CPTS). ClinMPO uses ClinRM, a reward model trained on 18,569 question--answer pairs from 4,474 psychiatry articles. We evaluated four Qwen3 sizes on 1,737 model-screened questions. ClinMPO outperformed Base, supervised fine-tuning and standard group relative policy optimization across scales. From responses by 300 senior pre-licensure medical students, we established the human baseline, a medical-student reference. The 4B model approached this baseline, whereas the 8B model surpassed it and ranked first among 31 models and post-training variants. ClinMPO improved performance across two complementary schemes covering ICD-11 diagnostic categories and psychiatric practice competencies. Blinded assessment by three clinicians showed improved rationale quality across CPTS criteria. These findings highlight how existing clinical evidence and specialist knowledge can be incorporated into the development of medical AI systems through evidence-guided learning.}

\maketitle

\clearpage
\noindent{\fontsize{14}{18}\selectfont\bfseries Abstract}\par
\vspace{9pt}
\manuscriptabstract\par
\clearpage

\begin{figure}[p]
\centering
\includegraphics[width=\textwidth,height=0.84\textheight,keepaspectratio]{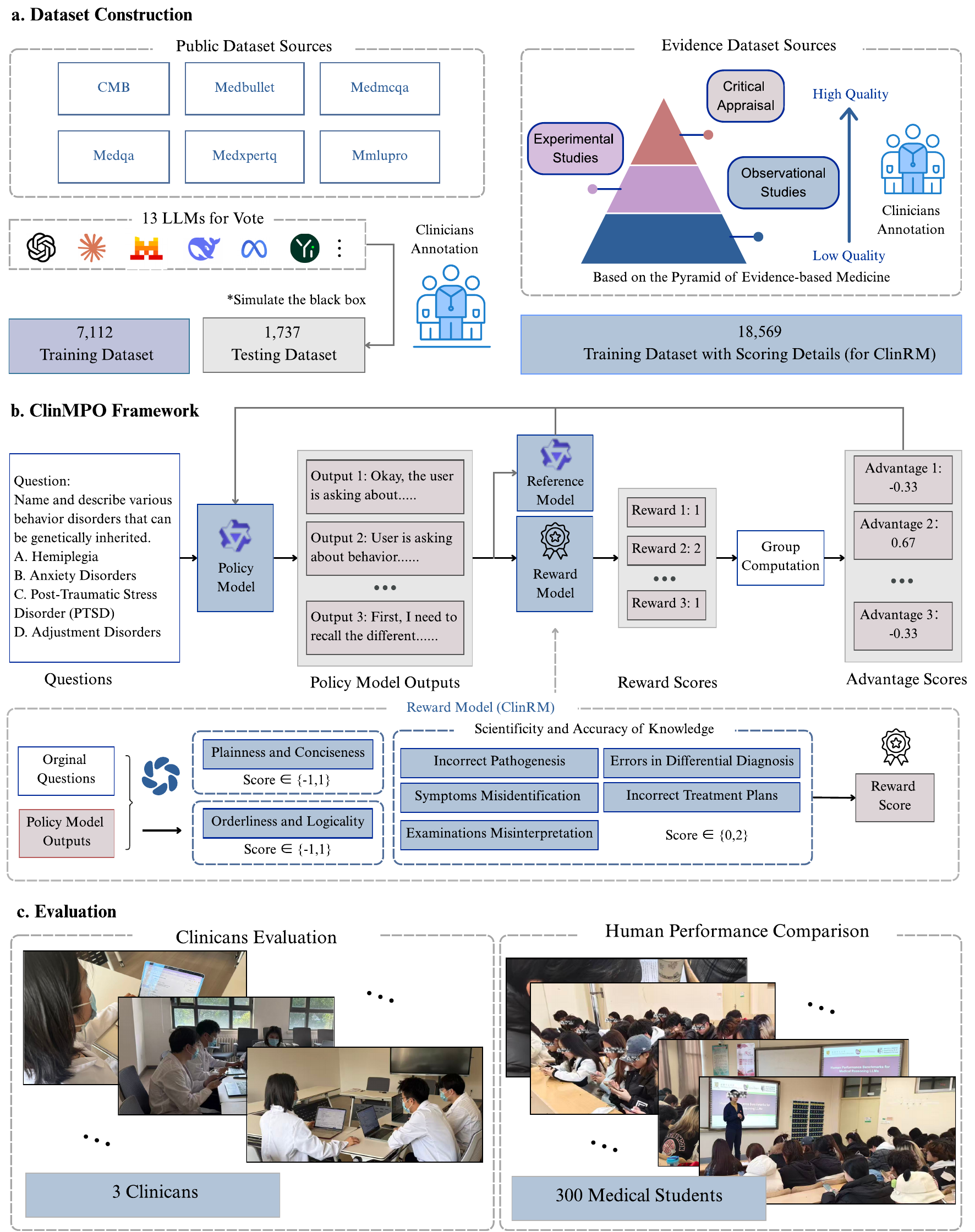}
\caption{\textbf{Overview of the ClinMPO framework. a} Data construction pipeline for Public QA Dataset and Evidence QA Dataset. \textbf{b} Illustration of the ClinMPO algorithm. Candidate responses are scored by a reward model (ClinRM) trained on the Evidence QA Dataset to apply the expert-defined CPTS criteria. Group-based reward and advantage calculations are then used to optimize the policy. \textbf{c} Model performance was evaluated using a refined clinical categorization framework and the human baseline from 300 medical students on a held-out test set, while generated rationales for 30\% of the test cases were independently assessed by three clinicians.}
\label{fig:1}
\end{figure}

\section{Introduction}\label{intro}

The global health workforce is projected to face a shortfall of 11 million health professionals by 2030, placing additional pressure on already unevenly distributed specialist services\cite{dai2026global}. This challenge is particularly acute in mental health: more than 1 billion people worldwide live with a mental health condition, yet access to care remains markedly unequal. In a 22-country WHO sample, service coverage for psychosis was 8\% in low-income countries and 54\% in high-income countries\cite{who2025mental,who2025atlas}. Across reporting countries, the median share of government health expenditure allocated to mental health was only 2.1\%, and the median specialized mental health workforce was 13.5 workers per 100,000 population\cite{who2025atlas}. Resource constraints are also evident in Shandong Province, China, where we recruited the cohort used to establish the human baseline, a medical-student reference for subsequent analyses. By the end of 2019, nearly 440,000 people with severe mental disorders, including schizophrenia, were registered for management in Shandong. Of these, 81\% lived in rural areas, and community health centers provided regular follow-up\cite{hu2022shandong}. The scale and distribution of this population underscore the need for clinical support tools that can operate within resource-constrained, privacy-sensitive health systems.

Psychiatric assessment requires language-mediated reasoning to integrate symptom narratives, mental-state findings, comorbidities, psychosocial context and treatment history\cite{harrison2017shorter,reed2019innovations}. Large language models (LLMs) can organize such information and support medical question answering, clinical text generation and diagnostic reasoning, but their computational demands and reliance on external infrastructure can impede privacy-preserving local deployment\cite{singhal2023large,singhal2025toward,tu2025conversational,obradovich2024opportunities}. With lower memory and infrastructure requirements, SLMs have therefore been proposed as a more deployable alternative to larger models, particularly where computational resources, cloud access and institutional governance constrain clinical implementation\cite{qin2026small,natbiomedeng2026critical}. Representative advances in medical and psychiatric SLMs include continued pretraining on biomedical corpora in BioMistral\cite{labrak2024biomistral} and domain-specific post-training in Psyche-R1, MentraSuite and PsychFound\cite{dai2026psyche,xiao2025mentrasuite,wang2026domain}. However, these approaches commonly depend on large curated datasets, extensive task-specific training resources or repeated human and expert annotation\cite{labrak2024biomistral,dai2026psyche,xiao2025mentrasuite,wang2026domain}. Such requirements are difficult to sustain in psychiatric settings, where specialist input is scarce and privacy constraints limit the aggregation and annotation of clinical data\cite{obradovich2024opportunities,zhang2025computational,malgaroli2025large}. Moreover, reducing model size alone does not ensure reliable multistep reasoning, and current psychiatric models remain vulnerable in diagnostic reasoning, treatment planning and longitudinal follow-up\cite{griot2025metacognition,fouda2026psychiatrybench}. A psychiatric SLM therefore requires a training strategy that improves reasoning ability while reducing dependence on repeated, response-level annotation by scarce experts.

To address this need, we developed ClinMPO, an evidence-guided reinforcement-learning framework for improving psychiatric reasoning in SLMs while limiting dependence on exhaustive expert annotation. Five psychiatry experts developed the Clinical Psychiatry Thinking Strategy (CPTS), which distinguishes clinical accuracy and completeness from clarity and concision and from organization and logical coherence. ClinRM learned to apply these criteria from 18,569 question--answer pairs derived from 4,474 psychiatry articles, thereby converting a one-time expert-defined rubric and literature-based evidence into a scalable reward signal. During policy optimization, ClinMPO uses ClinRM to score candidate responses and provide a criterion-resolved reward, without requiring psychiatrists to annotate each generated response. Policy training used 7,112 curated psychiatric questions drawn from six public medical question-answering datasets. These questions covered diagnostic and management reasoning across ICD-11 diagnostic categories and psychiatric practice competencies. In contrast to generic preference or outcome-only optimization, ClinMPO was designed to make evidence use and clinically consequential reasoning errors explicit components of the training objective.

We evaluated ClinMPO across four Qwen3 model sizes\cite{yang2025qwen3}, comparing each model with its corresponding Base, supervised fine-tuning (SFT), and standard group relative policy optimization (GRPO) variants. Evaluation included overall and category-level accuracy, the human baseline from 300 senior pre-licensure medical students and blinded assessment of generated rationales by three clinicians. We also examined generalization on non-psychiatric medical benchmarks and transfer to clinical summarization, dialogue-to-note generation and open-ended medical question answering. ClinMPO was the best-performing post-training method at every tested scale, with the largest gain over Base observed for the 4B model and the highest overall performance observed for the 8B model. Clinician assessment showed improvements in clinical accuracy and completeness, clarity and concision, and organization and logical coherence, while broader medical performance was maintained or improved in most comparisons. Together, these findings show that evidence-guided reinforcement learning can improve psychiatric reasoning in compact models without requiring response-level expert annotation during policy optimization, providing a practical basis for further evaluation in privacy-sensitive clinical settings. The overall data-construction, reward-modeling and policy-optimization workflow is summarized in Fig.~\ref{fig:1}.

\section{Results}\label{results}

\subsection{Validation of ClinRM against clinician consensus}
\label{clinrm-validation}

We compared ClinRM with clinician consensus on an independent set of 521 paired responses using the expert-developed CPTS rubric that evaluates clinical accuracy and completeness (C1), clarity and concision (C2), and organization and logical coherence (C3). This was the same 521-pair sample used for the clinician-rated rationale-quality analysis reported below. Because ClinMPO uses ClinRM scores to rank candidate responses within each group during policy optimization, this validation tested whether ClinRM assigned a higher score to the response preferred by clinicians. For each dimension, ClinRM and the clinician consensus classified the blinded, randomly ordered response pair as favoring response A, favoring response B or Tie. Conditional directional agreement measured whether ClinRM and clinician consensus selected the same response among pairs for which both expressed a non-tie A/B preference. Agreement was 71.7\% for C1 (147/205 pairs), 100.0\% for C2 (152/152 pairs), and 97.2\% for C3 (70/72 pairs). The number of pairs for which both ClinRM and the clinician consensus expressed an A/B preference differed across dimensions. Additional agreement metrics, expert-Tie recall and expert inter-rater reliability are reported separately (\suppfigref{supp-fig-clinrm-validation}; \supptableref{supp-tab-expert-interrater-reliability}).

\subsection{Performance across model scales and training conditions}
\label{clinmpo-improves-performance-across-model-scales}

\begin{figure}[tbp]
\centering
\includegraphics[width=.84\linewidth,keepaspectratio]{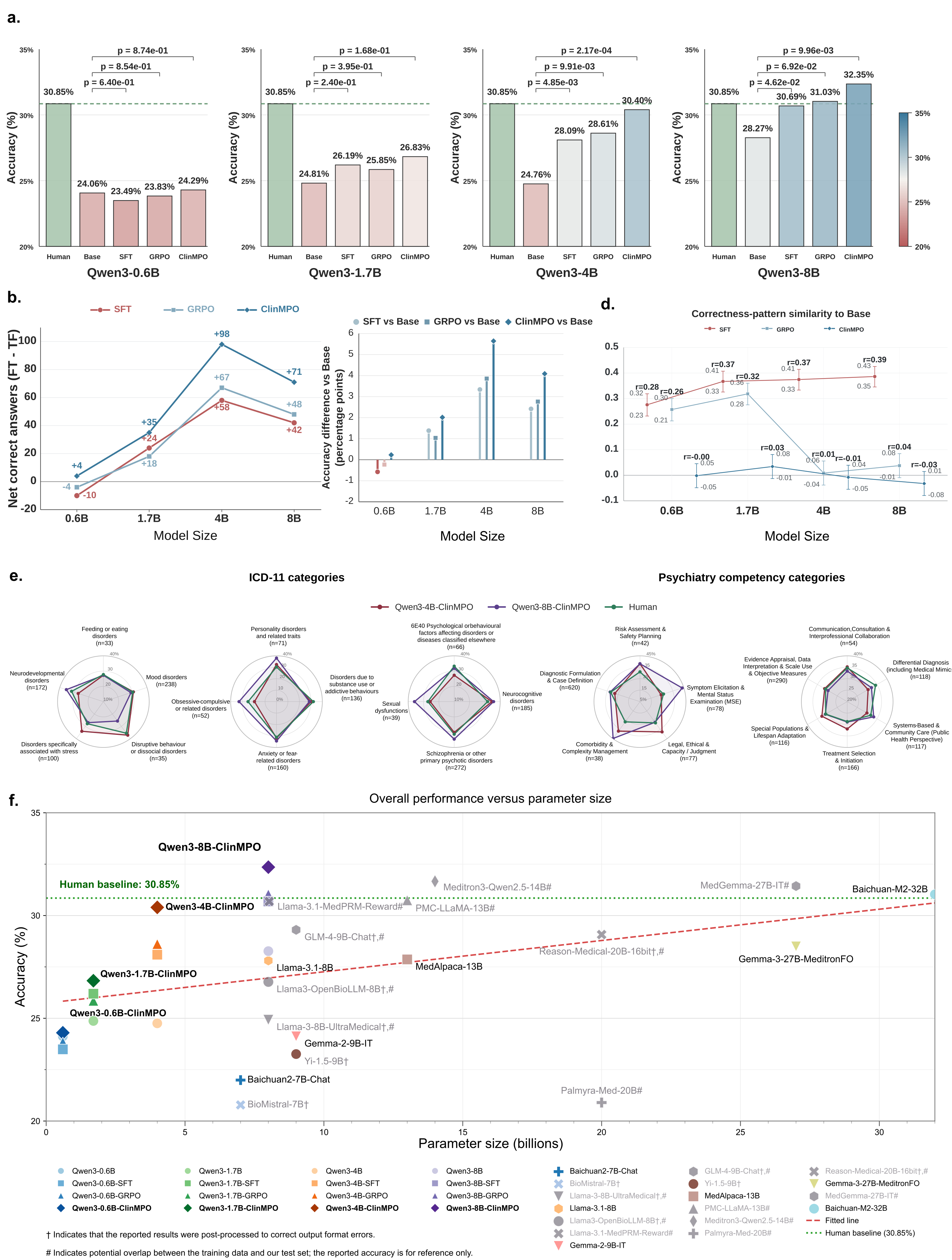}
\caption{\textbf{Performance of ClinMPO across model scales.} \textbf{a} Accuracy of the human baseline and Qwen3 Base, SFT, GRPO and ClinMPO models on 1,737 test questions. \textbf{b} Net change in correct answers relative to the size-matched Base model. \textbf{c} Accuracy difference from Base. \textbf{d} Pearson correlations and 95\% confidence intervals between question-level correctness patterns for each post-trained model and its size-matched Base model. \textbf{e} Profiles across the two classification dimensions for Qwen3-4B-ClinMPO, Qwen3-8B-ClinMPO and the human baseline. \textbf{f} Descriptive comparison of accuracy and parameter count across ClinMPO and external models. The green dotted line denotes the human baseline. The panel defines the symbols and possible training-data-overlap annotations.}
\label{fig:2}
\end{figure}

To determine whether ClinMPO improves psychiatric question answering across model capacities and relative to established post-training strategies, we evaluated Base, supervised fine-tuning (SFT), standard group relative policy optimization (GRPO) and ClinMPO variants for Qwen3 models ranging from 0.6B to 8B parameters on the same model-screened challenge test set of 1,737 psychiatric questions. We counted a response as correct only when its final option exactly matched the reference answer and used paired question-level analyses for comparisons with the size-matched Base models. Pre-split results from the 13-model panel used to construct the challenge test set appear separately (\supptableref{supp-tab-pre-split-model-accuracy}).

ClinMPO achieved the highest accuracy among the four evaluated variants at every model scale (Fig.~\ref{fig:2}a). The largest absolute improvement occurred at 4B parameters, where accuracy increased from 24.76\% for Base to 30.40\% for ClinMPO, a gain of 5.64 percentage points that exceeded the corresponding SFT and GRPO gains (Fig.~\ref{fig:2}c). This result approached the human baseline of 30.85\%. At 8B parameters, ClinMPO achieved the highest overall accuracy of 32.35\%, exceeding the same baseline by 1.50 percentage points. These results identify the 4B and 8B models as the strongest evaluated configurations for relative improvement and overall exact-match accuracy, respectively. Complete accuracy results and documentation of the student-facing interface and evaluation procedure are provided separately (\supptableref{supp-tab-all-model-accuracy}; \suppfigref{supp-fig-student-interface} and \suppfigref{supp-fig-student-evaluation}).

To quantify the question-level benefit underlying the accuracy gains, we analysed prediction transitions relative to the size-matched Base models (Fig.~\ref{fig:2}b). ClinMPO produced net increases of 98 correct responses at 4B and 71 at 8B, corresponding to gains of 5.64 and 4.09 percentage points, respectively. Paired McNemar tests supported both Base-to-ClinMPO improvements (4B, \(P=0.000217\); 8B, \(P=0.00996\)), indicating statistically significant differences in overall exact-match correctness for the two reported comparisons.

To assess whether these gains reflected broader changes in which questions were answered correctly, we compared question-level correctness patterns with those of the size-matched Base models. Correlations for ClinMPO were consistently close to zero across all four scales (\(r=-0.03\) to 0.03), whereas those for SFT ranged from \(r=0.28\) to 0.39 (Fig.~\ref{fig:2}d). Together with the positive net transitions, this pattern is consistent with ClinMPO substantially reorganizing the model's response profile rather than making only incremental changes to Base predictions. We further assessed parameter efficiency in a descriptive cross-system analysis comprising 31 open-source entries, including models and post-training variants, one closed-source model and the human baseline (Fig.~\ref{fig:2}f). Qwen3-8B-ClinMPO ranked first among 31 evaluated open-source entries, including models and post-training variants, on this fixed challenge set, while Qwen3-4B-ClinMPO outperformed several models with substantially more parameters.

\subsection{Performance across two complementary classification schemes}
\label{clinmpo-gains-span-domains-and-competencies}

To determine whether the overall gains extended across distinct areas of psychiatric knowledge and clinical reasoning, we evaluated ClinMPO using the two complementary classification schemes, ICD-11 diagnostic categories and psychiatric practice competencies. The former described the diagnostic content of each question, whereas the latter described the clinical task required to answer it. Separate tables define the complete categories, and accompanying figures show their cross-classification in the test and training sets (\supptableref{supp-tab-competencies} and \supptableref{supp-tab-icd11}; \suppfigref{supp-fig-test-set-category-cross-count} and \suppfigref{supp-fig-training-set-category-cross-count}). Category-level analyses compared ClinMPO with size-matched Base, SFT and GRPO models.

\begin{figure}[htbp]
\centering
\includegraphics[width=\linewidth,height=0.82\textheight,keepaspectratio]{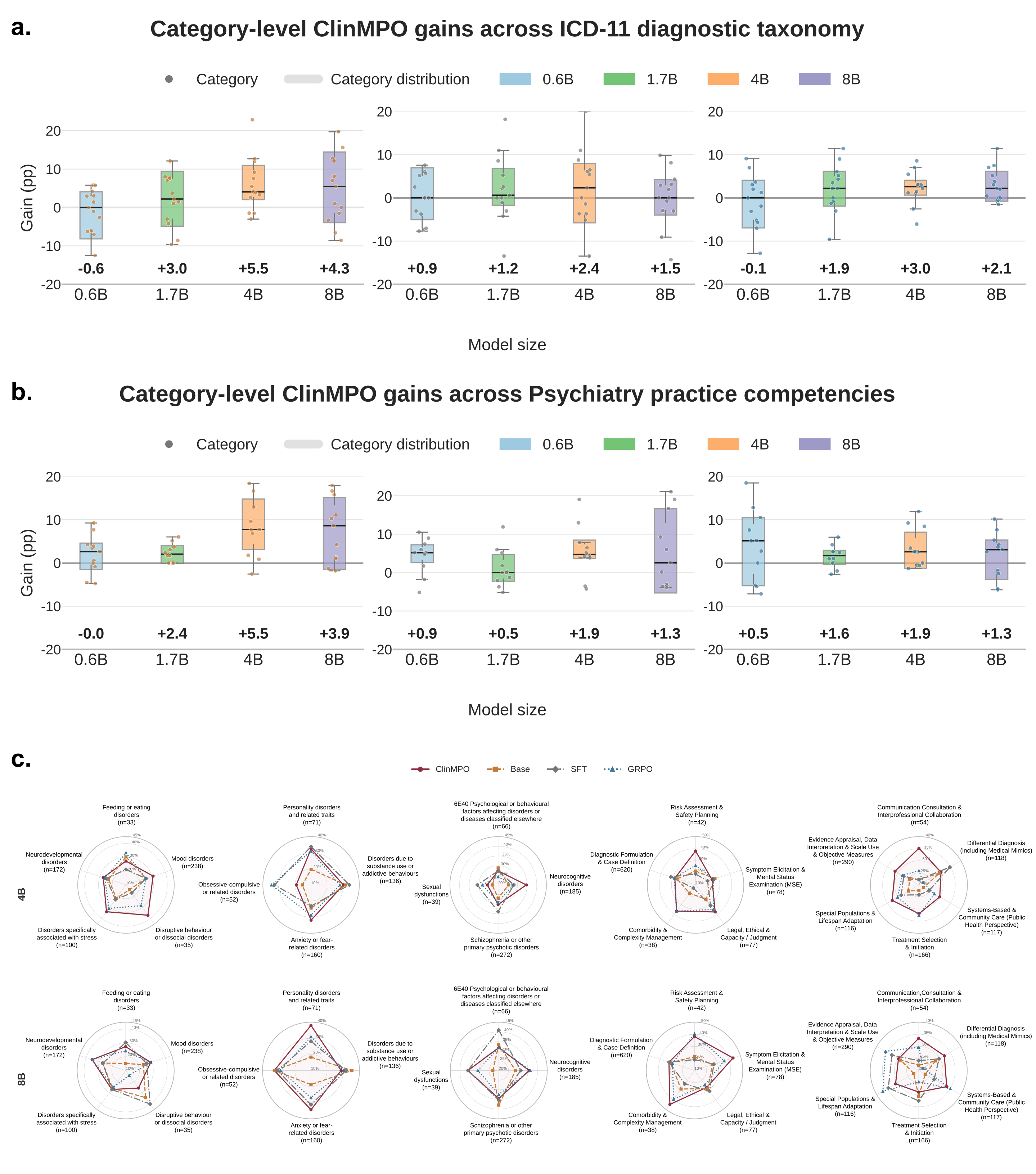}
\caption{\textbf{Category-level ClinMPO gains across the two classification dimensions.} We used the 1,737-item test set for all analyses. We counted predictions as correct only when they exactly matched the reference option; we counted empty or invalid outputs as incorrect. We computed category-level gains against size-matched Base, SFT and GRPO models. We excluded categories containing fewer than 30 questions from the primary category-level summaries to avoid emphasizing highly unstable estimates from sparsely represented categories. \textbf{a} ClinMPO accuracy gains across ICD-11 diagnostic categories. \textbf{b} ClinMPO accuracy gains across psychiatric practice competencies. Points represent retained categories, box plots summarize their distributions and labels below the boxes show question-weighted mean gains in percentage points. \textbf{c} Accuracy profiles of ClinMPO, Base, SFT and GRPO at 4B and 8B across the retained categories in the two classification dimensions. Radial ranges are shared between the 4B and 8B plots within each column but differ across columns.}
\label{fig:3}
\end{figure}

ClinMPO showed positive question-weighted gains across the two classification dimensions and against all three size-matched comparators at the 4B and 8B scales (Fig.~\ref{fig:3}a,b). Across the retained diagnostic categories, the 4B model gained 5.52, 2.44 and 2.99 percentage points over Base, SFT and GRPO, respectively; the corresponding 8B gains were 4.30, 1.54 and 2.05 percentage points. The same pattern extended across the practice competencies: gains at 4B were 5.54, 1.91 and 1.92 percentage points, respectively, and those at 8B were 3.90, 1.34 and 1.28 percentage points. The concordant results across the two classification dimensions indicate that ClinMPO's overall benefit extended beyond a single type of psychiatric knowledge.

To examine the breadth of this benefit, we compared category-level accuracy profiles across the retained categories in the two classification dimensions (Fig.~\ref{fig:3}c). At 4B, ClinMPO outperformed Base in 20 of 24 categories, SFT in 16 and GRPO in 17. At 8B, it outperformed the respective comparators in 17, 13 and 16 categories. These profiles show that ClinMPO gains extended across the two classification dimensions, with the broadest advantage observed for the 4B model relative to Base. Separate tables report the categories excluded by the minimum sample-size criterion and their descriptive accuracy results (\supptableref{supp-tab-unplotted-categories} and \supptableref{supp-tab-unplotted-category-accuracy}).

\subsection{Clinician assessment of generated rationales}
\label{clinician-review-identifies-stronger-rationales}

To determine whether the improvement in final-answer accuracy extended to the clinical and communicative quality of the underlying rationales, three clinicians independently assessed blinded response pairs from Qwen3-8B and Qwen3-8B-ClinMPO for the same 521 randomly selected questions used in the independent ClinRM validation. The rubric evaluated clinical accuracy and completeness (C1), clarity and concision (C2), and organization and logical coherence (C3). C1 further distinguished errors in pathogenesis, symptom identification, examination interpretation, differential diagnosis and treatment planning. Separate figures document the evaluation procedure and blinded scoring interfaces (\suppfigref{supp-fig-clinician-evaluation}, \suppfigref{supp-fig-clinician-model-a-interface} and \suppfigref{supp-fig-clinician-model-b-interface}).

\begin{figure}[htbp]
\centering
\includegraphics[width=\linewidth,height=0.82\textheight,keepaspectratio]{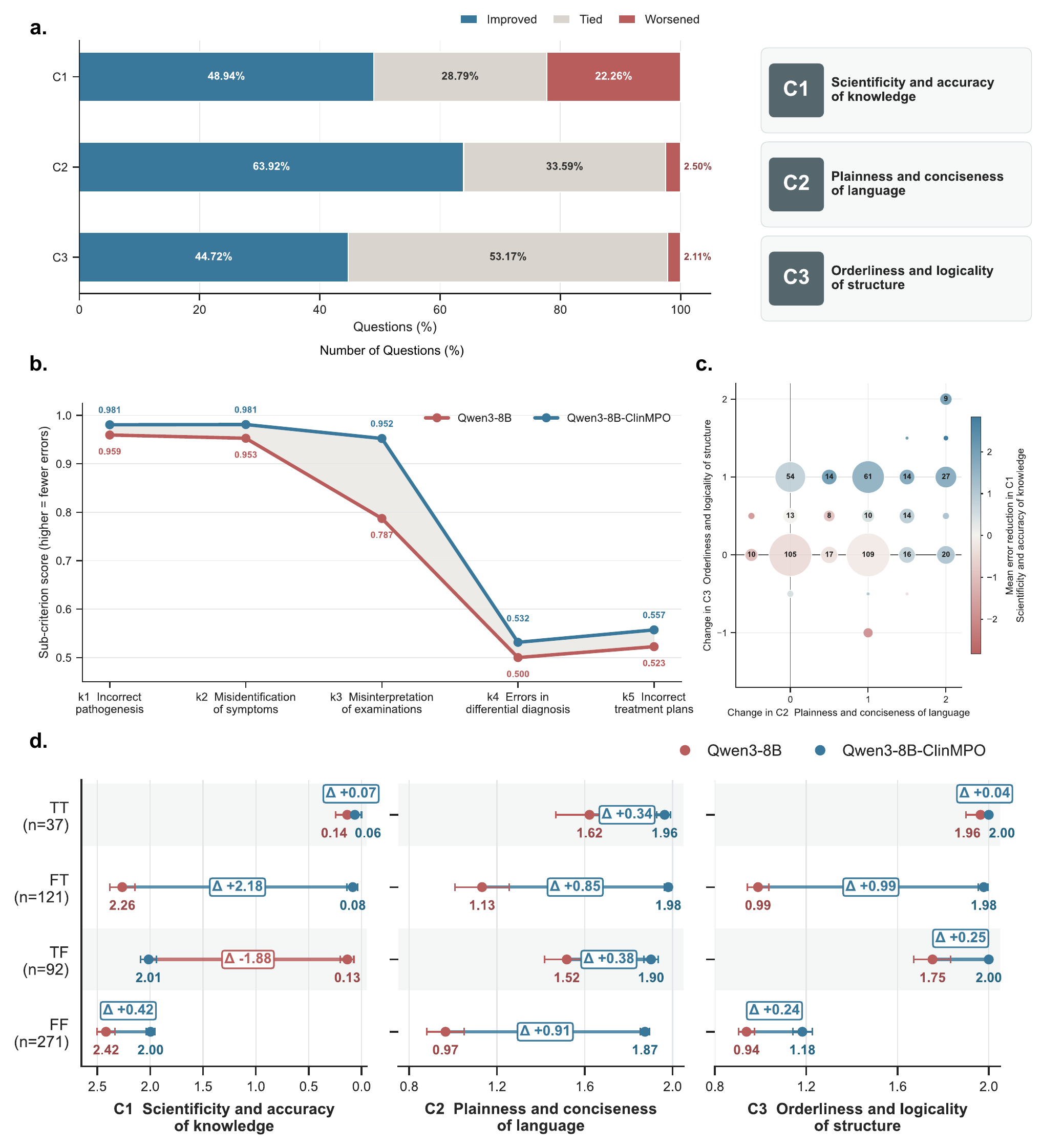}
\caption{\textbf{Blinded clinician evaluation of generated rationales.} Three clinicians independently assessed paired rationales from Qwen3-8B and Qwen3-8B-ClinMPO for the same 521 randomly sampled questions used in ClinRM validation, representing 30.0\% of the 1,737-question test set. We averaged scores across clinicians at the question level. \textbf{a} Question-level changes in clinical accuracy and completeness (C1), clarity and concision (C2), and organization and logical coherence (C3). \textbf{b} Scores across five C1 error categories. Higher values indicate fewer identified errors. \textbf{c} Joint changes in C2, C3 and C1. Bubble size denotes the number of questions and color denotes the mean change in C1. \textbf{d} Clinician scores stratified by final-answer transitions. TT, both models correct; FT, corrected by ClinMPO; TF, correct only for Qwen3-8B; FF, both models incorrect. Points denote means and error bars denote 95\% bootstrap confidence intervals.}
\label{fig:4}
\end{figure}

ClinMPO improved more responses than it worsened on all three dimensions (Fig.~\ref{fig:4}a). Clarity and concision improved in 63.92\% of questions and improved or remained unchanged in approximately 97.5\%. Organization and logical coherence improved in 44.72\% and improved or remained unchanged in 97.89\%. Clinical error burden decreased in 48.94\% of questions and decreased or remained unchanged in 77.73\%. Thus, clinician assessment showed the broadest improvement in clarity and concision, together with predominantly favorable or stable outcomes in organization and logical coherence and in clinical accuracy and completeness. Importantly, the gains extended beyond output presentation. The primary accuracy endpoint awarded credit solely for an exact match to the reference option and did not incorporate stylistic or formatting scores. Moreover, among questions for which ClinMPO corrected an incorrect Base-model answer, the mean improvement in clinical accuracy and completeness was +2.18, alongside improvements in clarity and concision and in organization and logical coherence. Together, these findings indicate that ClinMPO improved not only final-answer accuracy but also the clinician-rated clinical and communicative quality of the generated rationales.

To localize the clinical benefit, we compared clinician scores across the five predefined medical-error categories. Mean clinician scores were higher with ClinMPO in each of the five error categories (Fig.~\ref{fig:4}b), with examination interpretation showing the largest numerical increase (0.787 to 0.952). Joint analysis further identified response groups with concurrent improvements in clinical accuracy and completeness, clarity and concision, and organization and logical coherence (Fig.~\ref{fig:4}c). In the false-to-true (FT) subgroup, in which ClinMPO corrected an incorrect Base-model answer, mean changes were +2.18 for C1, +0.85 for C2 and +0.99 for C3; even in false-to-false (FF) cases, in which both models produced an incorrect final answer, the corresponding changes remained positive (+0.42, +0.91 and +0.24) (Fig.~\ref{fig:4}d). In the prespecified true-to-false (TF) subgroup, in which the Base model was correct but ClinMPO was incorrect, the direction of C1 followed the change in final-answer correctness (mean change, -1.88), whereas C2 and C3 remained higher (+0.38 and +0.25), distinguishing clinical accuracy and completeness from clarity and concision and from organization and logical coherence in the transition-stratified analysis.
An illustrative paired response with clinician annotations appears separately (\suppnoteref{supp-note-error-analysis}{11}; \suppfigref{supp-fig-baseline} and \suppfigref{supp-fig-clinmpo}).

\subsection{Performance on broader medical tasks}
\label{clinmpo-preserves-broader-medical-performance}

We evaluated performance in two settings that differed from the psychiatric questions used for post-training. First, we tested Base and ClinMPO models at four scales on item-disjoint, non-psychiatric subsets of CMB\cite{wang2024cmb}, MMLU-Pro\cite{wang2024mmlupro} and MedMCQA\cite{pal2022medmcqa}. Second, we evaluated Qwen3-8B and Qwen3-8B-ClinMPO on section summarization, dialogue-to-note generation and open-ended medical question answering. We used no evaluation samples for SFT, ClinMPO policy optimization or ClinRM training.

Across the 12 model-size-by-benchmark comparisons, ClinMPO achieved higher accuracy in ten, including every comparison from 1.7B to 8B (Table 1). At 4B, accuracy increased from 69.37\% to 69.48\% on CMB, 52.21\% to 53.42\% on MMLU-Pro and 52.44\% to 54.10\% on MedMCQA. At 8B, the corresponding increases were 74.37\% to 74.44\%, 59.67\% to 60.27\% and 60.64\% to 61.47\%. These results indicate that ClinMPO retained its psychiatric post-training gains across item-disjoint, non-psychiatric medical benchmarks at model scales from 1.7B to 8B.

{\small\setstretch{1.04}\renewcommand{\arraystretch}{1.28}
\begin{longtable}[]{@{}
  >{\raggedright\arraybackslash}p{(\linewidth - 6\tabcolsep) * \real{0.2500}}
  >{\raggedright\arraybackslash}p{(\linewidth - 6\tabcolsep) * \real{0.2500}}
  >{\raggedright\arraybackslash}p{(\linewidth - 6\tabcolsep) * \real{0.2500}}
  >{\raggedright\arraybackslash}p{(\linewidth - 6\tabcolsep) * \real{0.2500}}@{}}
\caption{Performance comparison of models across item-disjoint, non-psychiatric benchmark subsets.}\label{tab:ood-benchmarks}\\
\toprule\noalign{}
\begin{minipage}[b]{\linewidth}\raggedright
\textbf{Model}
\end{minipage} & \begin{minipage}[b]{\linewidth}\raggedright
\makecell[l]{\textbf{CMB accuracy}\\\textbf{(\%; correct/total)}}
\end{minipage} & \begin{minipage}[b]{\linewidth}\raggedright
\makecell[l]{\textbf{MMLU-Pro accuracy}\\\textbf{(\%; correct/total)}}
\end{minipage} & \begin{minipage}[b]{\linewidth}\raggedright
\makecell[l]{\textbf{MedMCQA accuracy}\\\textbf{(\%; correct/total)}}
\end{minipage} \\
\midrule\noalign{}
\endhead
\bottomrule\noalign{}
\endlastfoot
Qwen3-0.6B & 36.88\% (1049/2844) & 17.66\% (116/657) & 36.27\% (305/841) \\
Qwen3-0.6B-ClinMPO & 36.18\% (1029/2844) & 18.72\% (123/657) & 35.67\% (300/841) \\
Qwen3-1.7B & 51.90\% (1476/2844) & 34.09\% (224/657) & 44.59\% (375/841) \\
Qwen3-1.7B-ClinMPO & 53.41\% (1519/2844) & 34.55\% (227/657) & 45.66\% (384/841) \\
Qwen3-4B & 69.37\% (1973/2844) & 52.21\% (343/657) & 52.44\% (441/841) \\
Qwen3-4B-ClinMPO & 69.48\% (1976/2844) & 53.42\% (351/657) & 54.10\% (455/841) \\
Qwen3-8B & 74.37\% (2115/2844) & 59.67\% (392/657) & 60.64\% (510/841) \\
Qwen3-8B-ClinMPO & 74.44\% (2117/2844) & 60.27\% (396/657) & 61.47\% (517/841) \\
\end{longtable}
}

For cross-task evaluation, we used MTS-Dialog\cite{benabacha2023mtsdialog} for clinical section summarization, MEDIQA-Chat 2023 Task B\cite{benabacha2023mediqachat} with ACI-Bench\cite{yim2023acibench} for dialogue-to-note generation, and HealthCareMagic-100k-Chat-Format-en (HealthCareMagic), which follows the format used by ChatDoctor\cite{li2023chatdoctor}, for open-ended medical question answering. The same post-processing procedure converted the generated rationales into task-specific outputs for both models.

Qwen3-8B-ClinMPO achieved higher values for 11 of the 12 reported dataset--metric combinations and the same value for the remaining comparison at the reported precision (Table 2). The largest differences occurred for dialogue-to-note generation, where BERTScore recall increased from 0.652 to 0.666, ROUGE-1 from 0.442 to 0.482, ROUGE-L from 0.286 to 0.291 and ROUGE-Lsum from 0.401 to 0.443. MTS-Dialog and HealthCareMagic showed smaller, directionally consistent differences.

{\small\setstretch{1.04}\renewcommand{\arraystretch}{1.28}
\begin{longtable}[t]{@{}
  >{\raggedright\arraybackslash}p{(\linewidth - 8\tabcolsep) * \real{0.2200}}
  >{\raggedright\arraybackslash}p{(\linewidth - 8\tabcolsep) * \real{0.2000}}
  >{\centering\arraybackslash}p{(\linewidth - 8\tabcolsep) * \real{0.2000}}
  >{\centering\arraybackslash}p{(\linewidth - 8\tabcolsep) * \real{0.0950}}
  >{\centering\arraybackslash}p{(\linewidth - 8\tabcolsep) * \real{0.0950}}
  >{\centering\arraybackslash}p{(\linewidth - 8\tabcolsep) * \real{0.0900}}@{}}
\caption{Performance of formatted model outputs across external clinical text-generation tasks.}\label{tab:clinical-generation}\\
\toprule\noalign{}
\textbf{Task} & \textbf{Model} & \textbf{BERTScore} & \multicolumn{3}{c@{}}{\textbf{ROUGE}} \\
\cmidrule(lr){3-3}\cmidrule(l){4-6}
 & & \textbf{Recall} & \textbf{1} & \textbf{L} & \textbf{Lsum} \\
\midrule\noalign{}
\endhead
\bottomrule\noalign{}
\endlastfoot
MTS-Dialog & Qwen3-8B & 0.708 & 0.372 & 0.303 & 0.303 \\
 & Qwen3-8B-ClinMPO & 0.717 & 0.376 & 0.306 & 0.306 \\
\addlinespace[0.35em]
MEDIQA-Chat 2023 Task B & Qwen3-8B & 0.652 & 0.442 & 0.286 & 0.401 \\
 & Qwen3-8B-ClinMPO & 0.666 & 0.482 & 0.291 & 0.443 \\
\addlinespace[0.35em]
HealthCareMagic & Qwen3-8B & 0.564 & 0.258 & 0.128 & 0.135 \\
 & Qwen3-8B-ClinMPO & 0.568 & 0.260 & 0.128 & 0.138 \\
\end{longtable}
}

\section{Discussion}\label{discussion}

Small language models (SLMs) for psychiatric applications must support clinically structured reasoning while remaining suitable for institutionally governed deployment. We addressed this challenge with ClinMPO, which transforms literature-derived evidence and a psychiatrist-defined reasoning rubric into a scalable reward signal for GRPO-based policy optimization. This design avoids response-level expert annotation during policy training while retaining explicit supervision of clinical accuracy and completeness, clarity and concision, and organization and logical coherence. Across four Qwen3 model sizes, ClinMPO increased overall exact-match accuracy on a difficulty-enriched psychiatric question set and improved clinician-rated rationale quality. Descriptive category analyses suggested that these differences were not confined to a single diagnostic or competency category.

ClinMPO was the best-performing post-training method at every evaluated scale, with the 4B model gaining 5.64 percentage points over Base and the 8B model reaching the highest accuracy of 32.35\% (Fig.~\ref{fig:2}a--c). The 4B model approached the human baseline, whereas the 8B model modestly exceeded it at the aggregate level. ClinMPO also exceeded several larger open-source reference models in the parameter-efficiency comparison (Fig.~\ref{fig:2}f). The smaller marginal gain at 8B than at 4B may indicate that the fixed amount of training data and reward-guided optimization provided less adaptation signal relative to the capacity of the larger model. Alternatively, the response policy established during pretraining may be more resistant to modification at 8B, requiring greater data coverage or optimization exposure for the domain-specific reward to reshape it consistently. These explanations remain hypotheses and do not imply lower absolute performance or poorer generalization of the 8B model. Owing to computational-resource constraints, we could not perform controlled scaling experiments that varied training-data volume, candidate-sampling budget or optimization steps across model sizes. Such experiments will be needed to distinguish data-limited from optimization-limited effects. These scale-dependent patterns accord with evidence that model capacity, training-data volume and reward-data scale jointly shape optimization outcomes\cite{hoffmann2022empirical,gao2023scaling}.

The improvement extended across the two classification dimensions rather than concentrating in a single content area or task type (Fig.~\ref{fig:3}, \suppfigref{supp-fig-category-effects} and \suppfigref{supp-fig-training-category-heatmaps}). Blinded review showed that ClinMPO improved clarity and concision in 63.92\% of sampled responses, organization and logical coherence in 44.72\%, and clinical error burden in 48.94\% (Fig.~\ref{fig:4}a). Improvements across pathogenesis, symptom identification, examination interpretation, differential diagnosis and treatment planning indicate that the learned reward addressed several stages of psychiatric problem solving (Fig.~\ref{fig:4}b). Clinical error burden also decreased in some cases where both models selected an incorrect answer, suggesting that structured optimization can improve answer construction before the change is sufficient to alter the final choice (Fig.~\ref{fig:4}d).

At the level of the optimization design, ClinRM operationalizes the Clinical Psychiatry Thinking Strategy (CPTS) as criterion-resolved scores that separate clinical accuracy and completeness from clarity and concision and from organization and logical coherence. Independent validation showed substantial directional agreement between ClinRM and clinician-defined preferences, supporting its use as a scalable clinical reward. ClinMPO then uses this structured reward during GRPO-based policy optimization, prioritizing clinically consequential content while preserving readable and coherent responses. Its stronger performance than SFT and standard GRPO is consistent with the possibility that criterion-resolved supervision provides a richer learning signal than correctness-only optimization\cite{lightman2024verify,zhang2025process}. The diagnostic-by-competency analysis further suggests that optimization exposure across reasoning patterns contributes to category-level performance, consistent with evidence that data-mixture proportions influence language-model behavior\cite{xie2023doremi}. ClinMPO therefore links an expert-defined conceptual model of psychiatric reasoning to an optimization objective that can be applied repeatedly without repeated specialist scoring of individual policy outputs.

Psychiatry-specific optimization transferred to broader medical evaluation. ClinMPO improved ten of twelve model-size-by-benchmark comparisons on item-disjoint non-psychiatric subsets, with consistent gains from 1.7B parameters upward (Table~\ref{tab:ood-benchmarks}). On clinical generation, Qwen3-8B-ClinMPO improved eleven of twelve dataset--metric combinations and matched the remaining comparison across section summarization, dialogue-to-note generation and open-ended medical question answering (Table~\ref{tab:clinical-generation}). Together, these findings suggest that the effects of CPTS-guided reinforcement learning were not confined to reasoning in psychiatric multiple-choice tasks and may extend to broader medical knowledge assessment and clinical text generation.

This study has several limitations. First, although the training and test partitions of the Public QA Dataset were explicitly separated and semantically deduplicated, prior exposure to public content during model pretraining cannot be fully assessed because pretraining corpora are not completely observable\cite{deng2024investigating}. This is a general constraint in the evaluation of pretrained language models rather than one specific to ClinMPO. The model-screened challenge-test-set design reduced the contribution of questions readily solved through broadly shared model knowledge but could not exclude prior exposure or answer memorization. Comparing each ClinMPO model with a size-matched Base model sharing the same pretrained backbone and pretraining history further reduced the likelihood that shared exposure explained the observed post-training differences. Second, because inclusion in the in-domain challenge set depended on the performance of the 13-model screening panel, including Qwen3-8B, cross-model estimates and ranks on this fixed set are selection-dependent. Generalization was assessed separately using item-disjoint non-psychiatric benchmarks and external clinical text-generation datasets that were not used in performance-based partitioning. Although automated generation metrics do not directly measure factual completeness or clinical usefulness\cite{tang2023evaluating,agrawal2025evaluation}, blinded clinician assessment further evaluated whether the observed improvements extended to the clinical and communicative quality of generated rationales. Future evaluation on representative clinical cases is needed to determine whether these improvements translate to routine clinical tasks. Third, clinician review covered a random 30.0\% sample of the 8B-model outputs, and generated rationales, as with those of language models generally, are observable outputs rather than direct measurements of internal computation\cite{turpin2023language}. Further clinician review across model scales, together with perturbation-based analyses, is needed to characterize rationale faithfulness. In addition, end-to-end deployment remains to be evaluated. Further development will quantify efficiency on target hardware and assess performance for crisis content, medication contraindications and other safety-critical scenarios\cite{obradovich2024opportunities,zhang2025computational,lekadir2025future}. Clinical deployment would additionally require explicit uncertainty communication, refusal and escalation pathways, and clinician-facing interfaces that preserve independent clinical judgment.

In summary, ClinMPO provides an evidence-guided framework for aligning SLMs with psychiatrist-defined criteria for clinical accuracy and completeness, clarity and concision, and organization and logical coherence. Across four Qwen3 model sizes, CPTS-guided reinforcement learning improved exact-match performance on the fixed, difficulty-enriched psychiatric challenge set. The gains spanned the two classification dimensions. Clinician ratings also showed enhanced clinical and communicative quality of the generated rationales. Results on item-disjoint non-psychiatric benchmarks and external clinical text-generation tasks further suggest that these effects may extend beyond psychiatric multiple-choice reasoning. By converting literature-derived evidence and a one-time specialist-defined rubric into a scalable reward signal, ClinMPO reduces the need for response-level specialist annotation during policy optimization. ClinMPO therefore offers a reproducible approach to developing locally governed, domain-specialized SLMs for privacy-sensitive and resource-constrained clinical settings\cite{sokol2025artificial}.

\section{Methods}\label{methods}

\subsection{Ethical approval}\label{ethical-approval}

This study adhered to the principles of the Declaration of Helsinki. The human-participant components comprised a prospective evaluation involving senior pre-licensure medical students and a prospective blinded evaluation involving clinicians. The Survey and Behavioural Research Ethics Committee of The Chinese University of Hong Kong approved the medical student protocol (reference no. SBRE-25-0824), and the Research Ethics Committee of Qilu Hospital of Shandong University approved the clinician evaluation protocol (reference no. KYLL-2025-11-015-1). All participants provided informed consent before participation. The study collected no identifiable personal information and handled all responses in accordance with applicable privacy and data-protection requirements. Publicly available or literature-based sources supplied the model-development and benchmark datasets; the study accessed no retrospective patient-level clinical records.

\subsection{Clinical Psychiatry Thinking Strategy}\label{clinical-psychiatry-thinking-strategy}

Five psychiatry experts developed the Clinical Psychiatry Thinking Strategy (CPTS) with reference to the \textit{Shorter Oxford Textbook of Psychiatry}, 7th edition\cite{harrison2017shorter}. The strategy defines a structured rubric based on observable elements of psychiatric case assessment, diagnosis and management. These elements are formalized as explicit, auditable criteria used to construct training examples, supervise the reward model and evaluate generated rationales beyond final-answer correctness.

CPTS comprised three complementary dimensions. Clinical accuracy and completeness (C1) represented the core clinical reasoning pathway and assessed whether a response correctly addressed pathogenesis, identified relevant symptoms, interpreted examination findings, considered and discriminated among differential diagnoses, and formulated an appropriate treatment plan. Clarity and concision (C2) assessed whether the response communicated its reasoning precisely and efficiently without material ambiguity, redundancy or incomplete expression. Organization and logical coherence (C3) assessed whether the response followed a clinically interpretable sequence in which the available evidence supported the intermediate judgments and final conclusion. Separating these dimensions distinguished errors in clinical content from limitations in communication or reasoning structure while preserving their joint contribution to clinically structured psychiatric reasoning. The complete rubric, criterion definitions and scoring anchors appear separately (\suppnoteref{supp-note-cpts-development}{4}; \supptableref{supp-tab-cpts}).

The formal CPTS specification comprised the following criteria:

\begin{equation}
C=\begin{cases}
C_1: & \text{Clinical accuracy and completeness},\\
C_2: & \text{Clarity and concision},\\
C_3: & \text{Organization and logical coherence}.
\end{cases}
\label{eq:cpts-criteria}
\end{equation}

Within C1, the subcriteria covered pathogenesis, symptom identification, examination interpretation, differential diagnosis and treatment planning. This decomposition made the clinical content requirements explicit before ClinRM generated the criterion-level scores.

\subsection{Dataset preparation}\label{dataset-construction}

\subsubsection{Dataset construction}\label{dataset-collection-from-medical-evidence}

Two complementary question--answering datasets served distinct roles in the ClinMPO framework: the Public QA Dataset and the Evidence QA Dataset. The Public QA Dataset supported policy-model training and fixed held-out evaluation, whereas the Evidence QA Dataset supported ClinRM development. The datasets were constructed from different source pools: public medical question--answering datasets for policy training and evaluation, and psychiatry publications for ClinRM development.

The Public QA Dataset drew on six publicly available medical question--answering resources: CMB\cite{wang2024cmb}, MedBulletsQA\cite{langagilab2025medbullets}, MedMCQA\cite{pal2022medmcqa}, MedQA\cite{jin2021medqa}, MedXpertQA\cite{zuo2025medxpertqa} and MMLU-Pro\cite{wang2024mmlupro}. The dataset provided psychiatric questions for cold-start supervised fine-tuning, policy optimization and fixed held-out evaluation. GPT-5 screened the source questions for psychiatric relevance using the clinical topics, subspecialties and major psychiatric disorders represented in the \textit{Shorter Oxford Textbook of Psychiatry}, 7th edition\cite{harrison2017shorter}, yielding 9,649 candidates. Standardization added a question, answer options, a reference answer and an explanatory rationale to each retained item, while ClinicalBERT similarity at a 90\% threshold removed semantically redundant items\cite{alsentzer2019clinicalbert}. Five psychiatry experts independently reviewed a random 30\% subset for quality control, leaving 8,849 retained questions. Thirteen contemporary language models were evaluated on the retained questions to support performance-based dataset partitioning, with model-specific prediction coverage reported separately: GPT-4o, GPT-5, Gemini-2.5-Pro, GPT-5-Mini, Llama-3.1-8B, Mistral-7B, Claude-Sonnet-4, Claude-Sonnet-4-thinking, DeepSeek-V3, DeepSeek-R1, Grok-4, Qwen3-8B and Yi-1.5-9B. Questions answered correctly by at least 9 of the 13 models entered the 7,112-item model-screened higher-solvability training set, and the remaining 1,737 questions formed the fixed, difficulty-enriched test set. Missing model predictions were treated as incorrect for partitioning. All policy variants were evaluated on the same test questions, permitting paired comparisons. Separate materials provide source provenance, repository information and licensing, describe Public QA construction and overlap control, illustrate the complete construction workflow and report pre-partition model performance and prediction coverage (\suppnoteref{supp-note-dataset-provenance}{1}; \suppnoteref{supp-note-public-qa-construction}{2}; \supptableref{supp-tab-provenance}; \suppfigref{supp-fig-dataset-construction}; \supptableref{supp-tab-pre-split-model-accuracy}).

The Evidence QA Dataset followed the Oxford Centre for Evidence-Based Medicine grading framework and was constructed from psychiatry publications identified through MeSH-guided searches of OpenAlex\cite{priem2022openalex} and Europe PMC\cite{rosonovski2023europepmc}, covering January 2020 to October 2025. Eligible sources included practice guidelines, systematic reviews, controlled trials, observational studies and case reports. GPT-5 generated evidence-based four-option questions and CPTS-guided criterion-targeted variants, which yielded 18,569 question--answer pairs derived from 4,474 publications after SimHash-based deduplication\cite{charikar2002similarity}. The complete data-collection and processing workflow is described in \suppnoteref{supp-note-evidence-qa-construction}{3} and is openly available in the ClinMPO GitHub repository (\clinmporepo).

\subsubsection{Dataset categorization}\label{data-categorization-using-icd-11-diagnostic-categories-and-psychiatric-practice-competencies}

Two complementary schemes categorized the Public QA Dataset independently rather than as levels of a shared hierarchy. The first mapped each question to one of 26 ICD-11 diagnostic categories\cite{reed2019innovations}, representing the diagnostic content of the question. The second mapped each question to one of 12 psychiatric practice competencies, representing the reasoning task required to answer it. The competencies covered monitoring and follow-up, communication and consultation, risk assessment, differential diagnosis, systems-based care, treatment selection, symptom elicitation, legal and ethical assessment, comorbidity management, special populations, diagnostic formulation and evidence appraisal. The competency framework drew on clinical practice guidance from the American Psychiatric Association, NICE and the WHO mhGAP Intervention Guide, together with assessment and certification specifications from the United States Medical Licensing Examination, the National Board of Medical Examiners, the American Board of Psychiatry and Neurology and the Royal College of Psychiatrists. Five specialists refined the framework to ensure that it represented clinically relevant reasoning rather than topic labels alone.

The semi-automated categorization workflow first applied rule-based keyword matching tailored to ICD-11 psychiatric terminology and then used GPT-5 to assign items to diagnostic and competency categories. Five experts reviewed a random 10\% sample of the automatically categorized items. Separate materials define the complete categories, show cross-classifications of the test and training sets and document the classification prompt (\supptableref{supp-tab-competencies} and \supptableref{supp-tab-icd11}; \suppfigref{supp-fig-test-set-category-cross-count} and \suppfigref{supp-fig-training-set-category-cross-count}; \suppnoteref{supp-note-prompts}{12}).

\subsection{ClinMPO development}\label{clinmpo-development}

\subsubsection{ClinRM development}\label{reward-model-development}

ClinRM served as a scalable training-time reward model that operationalized CPTS. Following the generative-verifier formulation of reward modeling\cite{zhang2025generative}, ClinRM used Qwen3-8B\cite{yang2025qwen3} as its backbone and underwent SFT on the 18,569 Evidence QA pairs. Each input contained the clinical question, candidate answer options, reference answer and a generated rationale with its final selection. The reference answer allowed ClinRM to identify factual and clinical inconsistencies, while the rationale provided the text required to assess the C1--C3 dimensions. ClinRM returned a structured JSON record containing criterion-level scores and their associated assessment. This format enabled automated parsing of each dimension and aggregation into a scalar reward without requiring response-level psychiatrist annotation during policy optimization. ClinRM operated only during policy training and did not serve as a test-time predictor; additional development details appear in \suppnoteref{supp-note-clinrm-development}{5}.

For the five C1 subcriteria, a response received 0 when it failed to meet the corresponding clinical requirement and +2 when it addressed the requirement correctly. C2 and C3 received -1 for linguistic or organizational problems and +1 when the response satisfied the respective criterion. The lower range for C2 and C3 gave clinically grounded accuracy and completeness greater influence than generic presentation qualities. The sum of these criterion-level scores defined the CPTS reward:

\begin{equation}
R_{\mathrm{CPTS}}(s,a)=\sum_{j=1}^{5}s_j(s,a)+s_{C2}(s,a)+s_{C3}(s,a),
\label{eq:cpts-reward}
\end{equation}

where \(s\) denotes the input question context, \(a\) denotes a candidate response, \(s_j\) denotes the score for the \(j\)th C1 subcriterion and \(s_{C2}\) and \(s_{C3}\) denote the language and organization scores. Higher values therefore indicate fewer clinical errors and stronger presentation. Complete scoring anchors and the mathematical formulation appear separately (\supptableref{supp-tab-cpts}; \suppnoteref{supp-note-clinmpo-implementation}{7}).

\subsubsection{Cold-start supervised fine-tuning}\label{cold-start-supervised-fine-tuning}

Cold-start SFT used the 7,112-item higher-solvability training set and applied Low-Rank Adaptation (LoRA)\cite{hu2022lora} to Qwen3-0.6B, Qwen3-1.7B, Qwen3-4B and Qwen3-8B\cite{yang2025qwen3}. We selected the Qwen3 family because its open-weight availability permits reproducible local implementation and evaluation and because it provides strong general-purpose capabilities across compact model sizes. Evaluating four sizes from the same model family allowed us to assess the scale dependence of ClinMPO while limiting variation in model architecture and pretraining lineage. The resulting four SFT models initialized the policies for subsequent reinforcement learning and enabled evaluation of the same two-stage training design across model scales.

\subsubsection{ClinMPO policy optimization}\label{reinforcement-learning-via-grpo}

Group relative policy optimization (GRPO)\cite{shao2024deepseekmath} optimized the SFT-initialized policies on the same 7,112-item higher-solvability training set. For the standard GRPO comparator, candidate responses were scored using a rule-based reference-answer-consistency reward based on whether the extracted final option matched the reference answer. For each prompt, the policy generated a group of candidate responses and received this reward for each candidate. Centering each reward by the group mean and scaling it by the within-group standard deviation produced a relative advantage that indicated whether a response performed better or worse than the other candidates generated for the same prompt. This construction avoided the need for a separately trained value model and made each update depend on within-question ranking rather than the absolute scale of rewards across unrelated questions.

Policy updates followed the standard clipped GRPO objective. The objective multiplied the probability ratio between the updated and sampling policies by the normalized advantage and applied clipping to restrict changes outside a predefined interval. For positively advantaged responses, the update increased their probability without allowing an unconstrained step; for negatively advantaged responses, it reduced their probability under the same bound. A Kullback--Leibler penalty relative to the SFT-initialized reference policy further limited drift from the cold-start model. The objective averaged candidate-level terms within groups and then across training prompts. The full group-normalization and GRPO objective equations appear separately (\suppnoteref{supp-note-clinmpo-implementation}{7}).

ClinMPO retained the GRPO optimization framework but replaced the reference-answer-consistency reward with the ClinRM-derived CPTS reward, $R_{\mathrm{CPTS}}(s,a)$. This reward supplied criterion-resolved information about clinical accuracy and completeness, clarity and concision, and organization and logical coherence, allowing clinically consequential content to contribute more directly to policy updates than under a correctness-only or generic presentation reward.

ClinMPO normalized rewards within each candidate group using the same procedure as standard GRPO. Responses with rewards above the group mean received positive advantages, whereas responses below the group mean received negative advantages; the within-group standard deviation placed these differences on a comparable scale. The resulting clinically adjusted advantages replaced the standard GRPO advantages in the clipped objective, while the sampling ratio, clipping operation and Kullback--Leibler regularization remained unchanged. The SFT-initialized model served as the reference policy. Thus, ClinMPO modified the source and clinical structure of the learning signal while preserving the underlying group-based optimization procedure, enabling direct comparison with standard GRPO. Separate materials detail the advantage calculations, objective functions and available training settings (\suppnoteref{supp-note-clinmpo-implementation}{7}; \supptableref{supp-tab-training-configuration}). The reproducible training implementation and complete run configurations are publicly available in the ClinMPO GitHub repository (\clinmporepo).

\subsection{Human baseline}\label{human-cohort-medical-students-baseline-evaluation}

Participants received modest compensation and remained blinded to model performance and the study hypotheses. The protocol involved no deception.

The human baseline, a medical-student reference, was derived from 300 senior pre-licensure medical students in Shandong who had completed formal psychiatry training as part of their medical curriculum. Participants answered assigned questions from the same 1,737-item test set used for model evaluation through a supervised protocol combining paper-based testing and an online platform. Randomized question assignment ensured that every question received six independent student responses. For each question, the proportion of students selecting the reference answer defined the item-level correct-response proportion; its mean across the test set defined the human baseline. The cohort provided an independent human educational anchor on these test items. The students' formal psychiatry training and relatively uniform educational stage allowed model accuracy to be interpreted in relation to a standardized learner cohort and the evaluated models. Specialist input complemented this baseline by targeting clinical reasoning dimensions not captured by final-option correctness through CPTS development, ClinRM validation and blinded assessment of clinical accuracy and completeness, clarity and concision, and organization and logical coherence in generated rationales. The protocol and participant-facing interfaces are documented separately (\suppnoteref{supp-note-medical-student-evaluation}{9}; \suppfigref{supp-fig-student-interface} and \suppfigref{supp-fig-student-evaluation}).

\subsection{Evaluation}\label{model-evaluation-metrics}

\subsubsection{Independent ClinRM validation}
\label{clinrm-independent-validation}

Independent ClinRM validation used 521 paired responses drawn from the fixed 1,737-question test set. This was the same 521-pair sample used for the blinded clinician analysis of rationale quality. These questions contributed neither to policy SFT or reinforcement learning nor to ClinRM development. Each pair comprised Qwen3-8B and Qwen3-8B-ClinMPO responses to the same question, answer options and reference answer. The evaluators and ClinRM received the reference answer to support identification of clinical errors, but the analysis did not treat it as a model-generated preference label. Deterministic randomization by question identifier with a fixed seed reduced response-order bias. The validation dataset contained complete expert assessments for all 521 pairs and valid ClinRM outputs for all 1,042 individual responses.

Three clinicians independently assessed both responses using C1, C2 and C3. The analysis aggregated the five C1 subcriteria into a clinical-error burden and analyzed C2 and C3 directly. Clinician preferences were converted to A, B or Tie labels, with the majority label across clinicians defining the expert consensus. Inter-rater reliability quantified agreement for the clinical-error and presentation dimensions.

ClinRM scores for each pair were converted to the same three-class preference format. The analysis calculated conditional directional agreement when both ClinRM and the expert consensus selected A or B, conservative directional agreement using all expert-decisive cases, exact three-class agreement and expert-Tie recall. Question-level bootstrap resampling provided point estimates and 95\% confidence intervals, and the same procedure evaluated a Qwen3-8B judge as a comparator. This validation assessed empirical alignment with clinician consensus rather than probability calibration or equivalence to clinical expertise. Full definitions, reliability estimates and comparative results appear separately (\suppnoteref{supp-note-clinrm-validation}{6}; \suppfigref{supp-fig-clinrm-validation}; \supptableref{supp-tab-expert-interrater-reliability}).

\subsubsection{In-domain evaluation and statistical analysis}

The in-domain evaluation compared Base, SFT, standard GRPO and ClinMPO variants of Qwen3-0.6B, 1.7B, 4B and 8B on the fixed 1,737-item test set. Exact final-option matching defined correctness, with missing or invalid outputs counted as incorrect. These four variants formed a controlled comparison within each model size. Base represented the pretrained model, SFT represented cold-start supervision, and GRPO and ClinMPO represented alternative policy-optimization procedures initialized from the corresponding SFT model. Overall performance was evaluated against the human baseline by averaging item-level student correct-response proportions across the same test questions. This baseline provided an independent human educational anchor for the model-screened challenge set. Question-level transitions relative to each size-matched Base model distinguished corrected responses from errors introduced by post-training: TT denoted both models correct; TF, Base correct and post-trained model incorrect; FT, Base incorrect and post-trained model correct; and FF, both incorrect. Net change in correct answers equalled FT minus TF. Two-sided McNemar tests assessed paired changes in correctness for the reported 4B and 8B Base--ClinMPO comparisons. Pearson correlations and 95\% confidence intervals quantified similarity between the question-level correctness patterns of each post-trained model and its Base model. A separate two-sided Pearson correlation described the association between parameter count and accuracy among directly comparable systems; differences in architecture, training data and possible benchmark exposure precluded a confirmatory interpretation. Detailed statistical methods appear in \suppnoteref{supp-sec-statistical-analysis}{8}.

A descriptive cross-system analysis evaluated 33 entries on the same test set, comprising 31 open-source entries (models and post-training variants), one closed-source model and the human baseline. The set included general-purpose open-source models, medically specialized systems and post-training variants. Fig.~\ref{fig:2}f displays systems that allowed direct comparison of parameter scale and marks potential training-data overlap or post-processed output formats. Differences in architectures, training resources and evaluation conditions limited these cross-model comparisons to descriptive rather than controlled inferential interpretation. Complete model-level results appear separately (\supptableref{supp-tab-all-model-accuracy}).

Category-level analyses calculated accuracy independently for the 26 ICD-11 diagnostic categories and 12 psychiatric practice competencies. We excluded categories containing fewer than 30 test questions from the primary summaries to avoid emphasizing highly unstable estimates from sparsely represented categories. Descriptive accuracy estimates appear in \suppfigref{supp-fig-category-effects}, \suppfigref{supp-fig-training-category-heatmaps}, \supptableref{supp-tab-unplotted-categories} and \supptableref{supp-tab-unplotted-category-accuracy}.

Using the same 521 response pairs and clinician ratings described for ClinRM validation, the expert evaluation compared blinded Qwen3-8B and Qwen3-8B-ClinMPO rationales. The sample represented 30.0\% of the test set. Clinicians scored each rationale on C1 clinical accuracy and completeness, C2 clarity and concision and C3 organization and logical coherence. Question-level analysis averaged scores across clinicians, summarized changes as better, tied or worse, and estimated mean differences with 95\% confidence intervals through bootstrap resampling. The protocol and interfaces are documented separately (\suppnoteref{supp-note-clinician-evaluation}{10}; \suppfigref{supp-fig-clinician-evaluation}, \suppfigref{supp-fig-clinician-model-a-interface} and \suppfigref{supp-fig-clinician-model-b-interface}).

\subsubsection{Out-of-domain and cross-task evaluation}

Out-of-domain evaluation used item-disjoint, non-psychiatric subsets of CMB, MMLU-Pro and MedMCQA. These comprised 2,844 CMB questions, 657 MMLU-Pro questions and 841 MedMCQA questions, for 4,342 questions in total. None contributed to psychiatric SFT, policy optimization or in-domain testing. The analysis calculated accuracy separately for each benchmark and compared each ClinMPO model with its size-matched Base model. Dataset-specific results remained separate rather than contributing to a pooled score because the three benchmarks differed in composition, difficulty and question format. This design descriptively assessed changes in broader medical-benchmark performance after psychiatric post-training without allowing the larger benchmark to dominate the interpretation.

Cross-task evaluation compared Qwen3-8B and Qwen3-8B-ClinMPO on three external clinical-generation tasks beyond psychiatric multiple-choice questions. Both models received identical prompts, chat templates and inference settings. MTS-Dialog assessed clinical section summarization using its complete 200-item test set\cite{benabacha2023mtsdialog}; MEDIQA-Chat 2023 Task B assessed dialogue-to-note generation using the 40-item full-dialogue ACI-Bench test set\cite{benabacha2023mediqachat,yim2023acibench}; and HealthCareMagic assessed open-ended medical question answering using 1,000 examples sampled from the public chat-format dataset with a fixed random seed of 42\cite{li2023chatdoctor}. Task-specific formatting preceded calculation of BERTScore recall, ROUGE-1, ROUGE-L and ROUGE-Lsum, while identical predefined rules assessed format validity, empty outputs and inference failures for both models.

\section{Data availability}\label{data-availability}

The six public datasets used to construct the Public QA Dataset are available from their respective public repositories: CMB, \url{https://github.com/FreedomIntelligence/CMB}; MedBulletsQA, \url{https://huggingface.co/datasets/LangAGI-Lab/medbullets}; MedMCQA, \url{https://github.com/medmcqa/medmcqa}; MedQA, \url{https://github.com/jind11/MedQA}; MedXpertQA, \url{https://github.com/TsinghuaC3I/MedXpertQA}; and MMLU-Pro, \url{https://github.com/TIGER-AI-Lab/MMLU-Pro}. A separate table summarizes the source-specific terms governing access and reuse (\supptableref{supp-tab-provenance}).

The external task datasets are publicly available from their original sources. The 200-item MTS-Dialog test set used for section summarization is available from \url{https://github.com/abachaa/MTS-Dialog} under the Creative Commons Attribution 4.0 International license (CC BY 4.0). The MEDIQA-Chat 2023 Task B definition and evaluation resources are available from \url{https://github.com/abachaa/MEDIQA-Chat-2023}; its 40-item test set is distributed as ACI-Bench TEST1 at \url{https://github.com/wyim/aci-bench} under CC BY 4.0. The HealthCareMagic-100k-Chat-Format-en dataset used for medical question answering is available from \url{https://huggingface.co/datasets/RafaelMPereira/HealthCareMagic-100k-Chat-Format-en}, whose dataset card declares the Apache License 2.0. The identifiers of the 1,000 HealthCareMagic examples sampled with random seed 42 are provided with the source data.

\section{Code availability}\label{code-availability}

Code for data processing, policy training and clinician-annotation analysis is available at \url{https://github.com/PAI-CUHK/ClinMPO}.

\section{Acknowledgments}
This work was supported by the Chinese University of Hong Kong Vice-Chancellor Early Career Professorship Scheme under Grant 441, the National Key Research and Development Program of China under Grant 2023YFB4706100, the National Natural Science Foundation of China under Grant 62573271, and the Major Basic Research Project of Shandong Provincial Natural Science Foundation under Grant ZR2025ZD24.

\backmatter
\bibliography{ref}

\section*{Author information}

\subsection*{Authors and Affiliations}

\noindent\textbf{Department of Psychiatry, The Chinese University of Hong Kong, Hong Kong SAR, China:} Xinxin Lin, Xiang Li, Jian Liu, Zhengdong Wu, Joey W. Y. Chan, Ngan Yin Chan, Sijing Chen, Yun Kwok Wing and Lizhou Fan.

\noindent\textbf{School of Control Science and Engineering, Shandong University, Jinan, Shandong, China:} Guangxin Dai, Shuo Wu, Jun Zhao and Xin Ma.

\noindent\textbf{Key Engineering Research Center of Intelligent Unmanned System of Ministry of Education, Shandong University, Jinan, Shandong, China:} Guangxin Dai, Shuo Wu, Jun Zhao and Xin Ma.

\noindent\textbf{Peking University Sixth Hospital, Peking University Institute of Mental Health, NHC Key Laboratory of Mental Health (Peking University), National Center for Mental Disorders, National Clinical Research Center for Mental Disorders (Peking University Sixth Hospital), Peking University, Beijing, China:} Yi Zhong and Lin Lu.

\noindent\textbf{Inspur Cloud Information Technology Co., Ltd., Jinan, China:} Xue Xiao.

\noindent\textbf{Department of Medicine, Shanghai Medical College Fudan University, Shanghai 200032, China:} Yixin Zhang.

\noindent\textbf{Department of Psychology, Shandong Provincial Hospital Affiliated to Shandong First Medical University, Jinan, China:} Lingming Hu, Yongbo Zheng and Yumei Wang.

\noindent\textbf{Li Chiu Kong Family Sleep Assessment Unit, Faculty of Medicine, The Chinese University of Hong Kong, Hong Kong SAR, China:} Zhengdong Wu, Joey W. Y. Chan, Ngan Yin Chan, Sijing Chen and Yun Kwok Wing.

\noindent\textbf{Institute of Brain Science and Brain-inspired Research, Shandong First Medical University and Shandong Academy of Medical Sciences, Jinan, Shandong, China:} Yongbo Zheng, Yi Zhang and Lin Lu.

\noindent\textbf{Department of Electrical and Computer Engineering, National University of Singapore, Singapore:} Runchuan Zhu.

\noindent\textbf{School of Intelligence Science and Technology, Nanjing University, Suzhou, Jiangsu, China:} Ming Zhao.

\noindent\textbf{Department of Medicine and Therapeutics, The Chinese University of Hong Kong, Hong Kong SAR, China:} Huizi Yu.

\noindent\textbf{Department of Neurology, The First Affiliated Hospital of Anhui Medical University, Anhui Medical University, Hefei 230022, China:} Fangting Lu.

\noindent\textbf{Hong Kong ICI Cloud Service Limited, Hong Kong SAR, China:} Ping Yin.

\noindent\textbf{Department of Psychology, Qilu Hospital of Shandong University, Jinan, China:} Lejin Yang.

\noindent\textbf{Department of Geriatric Medicine, Qilu Hospital of Shandong University, Jinan, Shandong, China:} Yanqiu Xing.

\noindent\textbf{Li Ka Shing Institute of Health Sciences, Faculty of Medicine, The Chinese University of Hong Kong, Hong Kong SAR, China:} Yun Kwok Wing and Lizhou Fan.

\noindent\textbf{Gerald Choa Neuroscience Institute, The Chinese University of Hong Kong, Hong Kong SAR, China:} Lin Lu.

\noindent\textbf{National Institute on Drug Dependence and Beijing Key Laboratory of Innovative Intervention Techniques and Translational Research for Mental Disorders Rehabilitation, Peking University, Beijing, China:} Lin Lu.

\subsection*{Author notes}
Xinxin Lin, Guangxin Dai, Yi Zhong and Xiang Li contributed equally to this work.

\subsection*{Author contributions}

\textbf{X.X.L.}: Conceptualization, Methodology, Software, Investigation, Formal analysis, Writing -- original draft, Writing -- review \& editing.
\textbf{G.X.D.}: Investigation, Project administration, Writing -- review \& editing.
\textbf{Y. Zhong}: Methodology, Data curation, Validation, Writing -- review \& editing.
\textbf{X.L.}: Data curation, Software, Writing -- review \& editing.
\textbf{X.X.}: Data curation, Software, Writing -- review \& editing.
\textbf{J.L.}: Validation, Writing -- review \& editing.
\textbf{Y.X.Z.}: Data curation, Validation, Writing -- review \& editing.
\textbf{L.M.H.}: Methodology, Data curation, Validation, Writing -- review \& editing.
\textbf{Z.D.W.}: Data curation, Software, Writing -- review \& editing.
\textbf{Y.B.Z.}: Data curation, Software, Validation, Writing -- review \& editing.
\textbf{R.C.Z.}: Data curation, Software, Writing -- review \& editing.
\textbf{M.Z.}: Data curation, Software, Writing -- review \& editing.
\textbf{H.Y.}: Data curation, Software, Writing -- review \& editing.
\textbf{Y. Zhang}: Methodology, Data curation, Validation, Writing -- review \& editing.
\textbf{F.T.L.}: Data curation, Validation, Writing -- review \& editing.
\textbf{S.W.}: Visualization, Writing -- review \& editing.
\textbf{J.Z.}: Visualization, Writing -- review \& editing.
\textbf{P.Y.}: Data curation, Software, Writing -- review \& editing.
\textbf{J.W.Y.C.}: Validation, Writing -- review \& editing.
\textbf{N.Y.C.}: Validation, Writing -- review \& editing.
\textbf{L.Y.}: Methodology, Data curation, Validation, Writing -- review \& editing.
\textbf{Y.M.W.}: Methodology, Data curation, Validation, Writing -- review \& editing.
\textbf{Y.Q.X.}: Validation, Writing -- review \& editing.
\textbf{S.J.C.}: Methodology, Data curation, Validation, Writing -- review \& editing.
\textbf{Y.K.W.}: Resources, Supervision, Project administration, Funding acquisition, Writing -- review \& editing.
\textbf{L.L.}: Resources, Supervision, Project administration, Writing -- review \& editing.
\textbf{X.M.}: Methodology, Resources, Supervision, Project administration, Funding acquisition, Writing -- review \& editing.
\textbf{L.F.}: Conceptualization, Methodology, Validation, Supervision, Resources, Project administration, Funding acquisition, Writing -- original draft, Writing -- review \& editing.

\subsection*{Corresponding authors}
Correspondence to Yun Kwok Wing, Lin Lu, Xin Ma or Lizhou Fan.

\section*{Ethics declarations}

\subsection*{Ethics approval}
The medical student study was approved by the Survey and Behavioural Research Ethics Committee of The Chinese University of Hong Kong (reference no. SBRE-25-0824). All participants provided electronic informed consent. The clinician evaluation study was approved by the Research Ethics Committee of Qilu Hospital of Shandong University (reference no. KYLL-2025-11-015-1), and all participants provided informed consent.

\end{document}


\begin{center}
  {\Large\bfseries Supplementary Information\par}
\end{center}

\vspace{1em}
\tableofcontents
\clearpage

\newcommand{\placeholder}[1]{\textcolor{red}{\textbf{[PLACEHOLDER: #1]}}}
\newcommand{\suppfigref}[1]{Supplementary Fig.~\hyperref[#1]{\textbf{\ref*{#1}}}}
\newcommand{\suppfigsref}[2]{Supplementary Figs.~\hyperref[#1]{\textbf{\ref*{#1}}} and \hyperref[#2]{\textbf{\ref*{#2}}}}
\newcommand{\suppfigrange}[2]{Supplementary Figs.~\hyperref[#1]{\textbf{\ref*{#1}}}--\hyperref[#2]{\textbf{\ref*{#2}}}}
\newcommand{\clinmporepo}{\url{https://github.com/PAI-CUHK/ClinMPO}}

\section{Supplementary Notes}

\subsection{Supplementary Note 1 | Dataset provenance, licensing and governance}
\label{supp-note-dataset-provenance}
The Public QA Dataset incorporated CMB, MedBulletsQA, MedMCQA, MedQA, MedXpertQA and MMLU-Pro. Repository and licensing details are provided in Supplementary Table~\ref{supp-tab-provenance}. No new identifiable patient data were collected. Evidence QA Dataset records were linked to source metadata from OpenAlex and Europe PMC.

\subsection{Supplementary Note 2 | Public QA construction and overlap control}
\label{supp-note-public-qa-construction}
GPT-5 screened the six sources for psychiatric relevance, after which records were standardized and semantically deduplicated using ClinicalBERT at a 90\% similarity threshold. Five psychiatry experts reviewed a random 30\% subset, yielding 8,849 questions. Questions answered correctly by at least 9 of the 13 evaluated models formed the 7,112-item model-screened higher-solvability training set. The remaining 1,737 questions formed the fixed, difficulty-enriched test set (\suppfigref{supp-fig-dataset-construction}a; Supplementary Table~\ref{supp-tab-pre-split-model-accuracy}). Missing model predictions were treated as incorrect for partitioning. All policy variants were evaluated on the same test questions, permitting paired comparisons. Model-specific prediction coverage is reported in Supplementary Table~\ref{supp-tab-pre-split-model-accuracy}.

\subsection{Supplementary Note 3 | Evidence QA construction and expert review}
\label{supp-note-evidence-qa-construction}
The Evidence QA Dataset was constructed according to the Oxford Centre for Evidence-Based Medicine grading framework, with source retrieval focused on critical-appraisal materials, experimental studies and observational studies. MeSH-guided searches of OpenAlex and Europe PMC identified psychiatry publications issued between January 2020 and October 2025. Eligible sources included practice guidelines, systematic reviews, controlled trials, observational studies and case reports. GPT-5 extracted the principal findings from the eligible publications and generated one or more self-contained four-option questions with reference answers and explanatory rationales. For each retained question, the original response and CPTS-guided criterion-targeted variants were generated to provide criterion-resolved training examples for ClinRM. Following SimHash-based deduplication, the retained examples yielded 18,569 Evidence QA question--answer pairs derived from 4,474 publications. Each pair included criterion-level scores and rationales defined in Supplementary Table~\ref{supp-tab-cpts}. Five clinical experts conducted a consensus-based binary quality assessment of 1,857 randomly selected training instructions (10.0\% of the retained instructions) against the source evidence and prespecified quality criteria. The experts reached a consensus classification of acceptable or unacceptable for each instruction. Of the 1,857 instructions assessed, 1,543 were classified as acceptable, yielding an acceptability rate of 83.09\% (95\% Wilson confidence interval, 81.32--84.73\%). The complete and reproducible data-collection and processing workflow is described in this Supplementary Note and is openly available in the ClinMPO GitHub repository (\clinmporepo).

\subsection{Supplementary Note 4 | CPTS development and scoring anchors}
\label{supp-note-cpts-development}
Five psychiatry experts developed the Clinical Psychiatry Thinking Strategy (CPTS). C1 assessed clinical accuracy and completeness across pathogenesis, symptoms, examinations, differential diagnosis and treatment; C2 assessed clarity and concision; and C3 assessed organization and logical coherence. C1 subcriteria were scored 0 or +2, whereas C2 and C3 were scored -1 or +1 (Supplementary Table~\ref{supp-tab-cpts}).

\subsection{Supplementary Note 5 | ClinRM development}
\label{supp-note-clinrm-development}
ClinRM was initialized from Qwen3-8B and fine-tuned on 18,569 Evidence QA pairs. Given a question, options, reference answer and model response, it generated structured CPTS scores whose sum defined the reward used for policy optimization. The rubric is provided in Supplementary Table~\ref{supp-tab-cpts}, and available ClinRM training details are summarized in Supplementary Table~\ref{supp-tab-training-configuration}.

\subsection{Supplementary Note 6 | Independent held-out validation of ClinRM}
\label{supp-note-clinrm-validation}
ClinRM was evaluated using the same 521 randomly selected Qwen3-8B and Qwen3-8B-ClinMPO response pairs and clinician ratings used for the blinded rationale-quality analysis. These questions were excluded from policy training and ClinRM development. Three clinicians scored each pair using C1--C3; majority A/B/Tie labels defined expert consensus, with reliability reported in Supplementary Table~\ref{supp-tab-expert-interrater-reliability}.

Conditional directional agreement was 71.7\% (95\% CI, 65.5--78.2\%) across 205 jointly decisive pairs for C1, 100.0\% (95\% CI, 100.0--100.0\%) across 152 jointly decisive pairs for C2 and 97.2\% (95\% CI, 92.6--100.0\%) across 72 jointly decisive pairs for C3. The denominators differed because this analysis excluded pairs for which either the clinicians or ClinRM assigned a Tie, rather than because of missing assessments. Bootstrap confidence intervals and comparison with a Qwen3-8B judge are reported (\suppfigref{supp-fig-clinrm-validation}a--e).

\subsection{Supplementary Note 7 | ClinMPO implementation and mathematical formulation}
\label{supp-note-clinmpo-implementation}
Qwen3 models at 0.6B, 1.7B, 4B and 8B parameters underwent Low-Rank Adaptation (LoRA) supervised fine-tuning (SFT) on the 7,112-item higher-solvability training set before policy optimization. ClinMPO used the group relative policy optimization (GRPO) objective with group-normalized advantages and the ClinRM-derived reward $R_{\mathrm{CPTS}}(s,a)$. Hardware, software and training settings are reported in Supplementary Table~\ref{supp-tab-training-configuration}.

ClinMPO retained the GRPO policy-update mechanism but replaced the standard GRPO reward with the ClinRM-derived CPTS reward. For the standard GRPO comparator, the reward was a rule-based indicator of whether the extracted final option matched the reference answer. The SFT initialization, policy-training data and group-relative optimization procedure were otherwise held constant between the GRPO and ClinMPO variants.

\label{supp-sec-mathematical-formulation}
\textbf{CPTS criteria.} Let $C$ denote the evaluation criteria:
\begin{equation}
C=\begin{cases}
C_1: & \text{Clinical accuracy and completeness},\\
C_2: & \text{Clarity and concision},\\
C_3: & \text{Organization and logical coherence}.
\end{cases}\label{supp-eq:cpts-criteria}
\end{equation}

For $C_1$, the subcriteria are:
\begin{equation}
C_1=\begin{cases}
k_1: & \text{Incorrect pathogenesis},\\
k_2: & \text{Misidentification of symptoms},\\
k_3: & \text{Misinterpretation of examinations},\\
k_4: & \text{Errors in differential diagnosis},\\
k_5: & \text{Incorrect treatment plans}.
\end{cases}\label{supp-eq:c1-subcriteria}
\end{equation}

For each $C_1$ subcriterion $k_j$, ClinRM assigns a score:
\begin{equation}
s_j(s,a)\in\{0,+2\},\qquad j=1,\ldots,5.
\label{supp-eq:clinical-score}
\end{equation}

For the remaining criteria $C_2$ and $C_3$, scalar scores are assigned as:
\begin{equation}
s_{C2}(s,a),s_{C3}(s,a)\in\{-1,+1\}.
\label{supp-eq:language-structure-score}
\end{equation}

The CPTS reward is computed by aggregating all criterion-level scores:
\begin{equation}
R_{\mathrm{CPTS}}(s,a)=\sum_{j=1}^{5}s_j(s,a)+s_{C2}(s,a)+s_{C3}(s,a).
\label{supp-eq:cpts-reward}
\end{equation}

\textbf{CPTS reward and ClinMPO optimization.}
A clinically adjusted advantage is computed using the same group-normalization scheme:
\begin{subequations}\label{supp-eq:clinmpo-advantage}
\begin{align}
\widehat{A}^{\mathrm{ClinMPO}}_i &= \frac{R_{\mathrm{CPTS}}(s,a_i)-\mu_{\mathrm{CPTS}}}{\sigma_{\mathrm{CPTS}}+\epsilon_{\mathrm{norm}}}, \\
\mu_{\mathrm{CPTS}} &= \frac{1}{G}\sum_{i=1}^{G}R_{\mathrm{CPTS}}(s,a_i), \\
\sigma_{\mathrm{CPTS}} &= \sqrt{\frac{1}{G}\sum_{i=1}^{G}\left(R_{\mathrm{CPTS}}(s,a_i)-\mu_{\mathrm{CPTS}}\right)^2}.
\end{align}
\end{subequations}

Here, $s$ denotes the input prompt, $a_i$ denotes the $i$th candidate response, $G$ is the number of candidate responses generated per prompt and $\epsilon_{\mathrm{norm}}$ is a small positive constant added to the denominator for numerical stability.

The resulting objective is:
\begin{subequations}\label{supp-eq:clinmpo-objective}
\begin{align}
L_{\mathrm{ClinMPO}}(\theta) &= \mathbb{E}\!\Bigg[\frac{1}{G}\sum_{i=1}^{G}\min\!\Big(\rho_i(\theta)\widehat{A}^{\mathrm{ClinMPO}}_i, \nonumber\\
&\qquad \operatorname{clip}(\rho_i(\theta),1-\epsilon_{\mathrm{clip}},1+\epsilon_{\mathrm{clip}})\widehat{A}^{\mathrm{ClinMPO}}_i\Big) \nonumber\\
&\qquad -\beta D_{\mathrm{KL}}\!\left(\pi_\theta\|\pi_{\mathrm{ref}}\right)\Bigg], \\
\rho_i(\theta) &= \frac{\pi_\theta(a_i\mid s)}{\pi_{\theta_{\mathrm{old}}}(a_i\mid s)}.
\end{align}
\end{subequations}

Here, $\pi_\theta$ is the current trainable policy, $\pi_{\theta_{\mathrm{old}}}$ is the rollout policy used to generate the candidate responses and $\pi_{\mathrm{ref}}$ is the fixed SFT-initialized reference policy. The probability ratio $\rho_i(\theta)$ follows the standard GRPO formulation. The clipping threshold $\epsilon_{\mathrm{clip}}$ limits policy updates, $\beta$ is the coefficient of the Kullback--Leibler penalty and $D_{\mathrm{KL}}$ denotes the Kullback--Leibler divergence.

\subsection{Supplementary Note 8 | Statistical analysis}
\label{supp-sec-statistical-analysis}
Exact final-option match on the 1,737-item test set defined accuracy; invalid outputs were counted as incorrect. Paired transitions were classified as TT, TF, FT or FF, with net gain defined as FT minus TF. Two-sided McNemar tests compared paired Base and ClinMPO correctness outcomes at 4B and 8B. Pearson correlations quantified similarity between binary question-level correctness patterns and, separately, the descriptive association between parameter count and accuracy. Clinician-score confidence intervals used question-level bootstrap resampling.

Category analyses used 26 ICD-11 diagnostic categories and 12 psychiatric practice competencies (Supplementary Tables~\ref{supp-tab-competencies} and \ref{supp-tab-icd11}). Categories containing fewer than 30 test questions were excluded from the primary category-level summaries to avoid emphasizing highly unstable estimates from sparsely represented categories. Complete category-level accuracy estimates for all categories are reported descriptively in the Supplementary Information. OOD accuracy was calculated separately for CMB, MMLU-Pro and MedMCQA.

\subsection{Supplementary Note 9 | Medical-student evaluation protocol}
\label{supp-note-medical-student-evaluation}
The evaluation included 300 senior pre-licensure medical students from Jining Medical University who had completed formal psychiatry training as part of their medical curriculum. The Chinese University of Hong Kong approved the protocol (SBRE-25-0824); participants provided electronic consent and were blinded to model performance and study hypotheses. Analyses used de-identified responses.

Each of the 1,737 test questions received six independently assigned student responses. The mean item-level correct-response proportion defined the human baseline, a medical-student reference (Supplementary Table~\ref{supp-tab-all-model-accuracy}). The interface and assessment procedure are documented (\suppfigsref{supp-fig-student-interface}{supp-fig-student-evaluation}).

The human baseline provided an independent human educational anchor on the same test items used for model evaluation. Senior pre-licensure students had completed formal psychiatry training and were at a relatively uniform educational stage, allowing model accuracy to be interpreted in relation to a standardized learner cohort and the evaluated models. Specialist input complemented this baseline by targeting clinical reasoning dimensions not captured by final-option correctness: psychiatrists developed CPTS, contributed to dataset quality control and categorization, supported independent ClinRM validation and assessed clinical accuracy and completeness, clarity and concision, and organization and logical coherence in generated rationales.

\subsection{Supplementary Note 10 | Clinician evaluation protocol}
\label{supp-note-clinician-evaluation}
The blinded rationale-quality analysis drew on the same three clinicians and 521 response pairs as ClinRM validation; these pairs represented 30.0\% of the test set. Scoring followed CPTS (Supplementary Table~\ref{supp-tab-cpts}). Clinician scores were averaged at the question level, and pairwise differences were classified as better, tied or worse. The evaluation procedure and interfaces are documented (\suppfigrange{supp-fig-clinician-evaluation}{supp-fig-clinician-model-b-interface}).

\subsection{Supplementary Note 11 | Illustrative paired error analysis}
\label{supp-note-error-analysis}
The following test-set example illustrates the clinician-assessed differences reported in Fig.~4.

\clearpage
\subsubsection*{Illustrative test item}
\begin{lstlisting}[style=prompt]
Question 1.

After three sessions with a therapy client, Dr. Leonard Lykowski realizes
that he is feeling somewhat hostile toward the client because she reminds
him of his wife, whom he is currently divorcing. Dr. Lykowski recognizes
that his emotional reaction may interfere with effective therapy. What is
the most appropriate professional action?

A. Consult with another psychologist to determine whether to continue
   seeing the client in therapy.
B. Seek personal therapy to manage his feelings and continue working with
   the client.
C. Refer the client to another therapist after discussing the reason with
   her.
D. Recognize that his feelings result from countertransference and continue
   working with the client.
E. Express his hostility toward the client during the session.
F. Terminate the therapy sessions without explanation.
G. Discuss his personal situation with the client to create transparency.
H. Ignore his feelings and continue working with the client.
I. Refer the client to another therapist without explaining the reason for
   the referral.
J. Ask the client to find another therapist without providing a reason.

Reference answer: A.
\end{lstlisting}

\begin{figure}[H]
  \centering
  \includegraphics[width=0.90\linewidth,height=0.66\textheight,keepaspectratio]{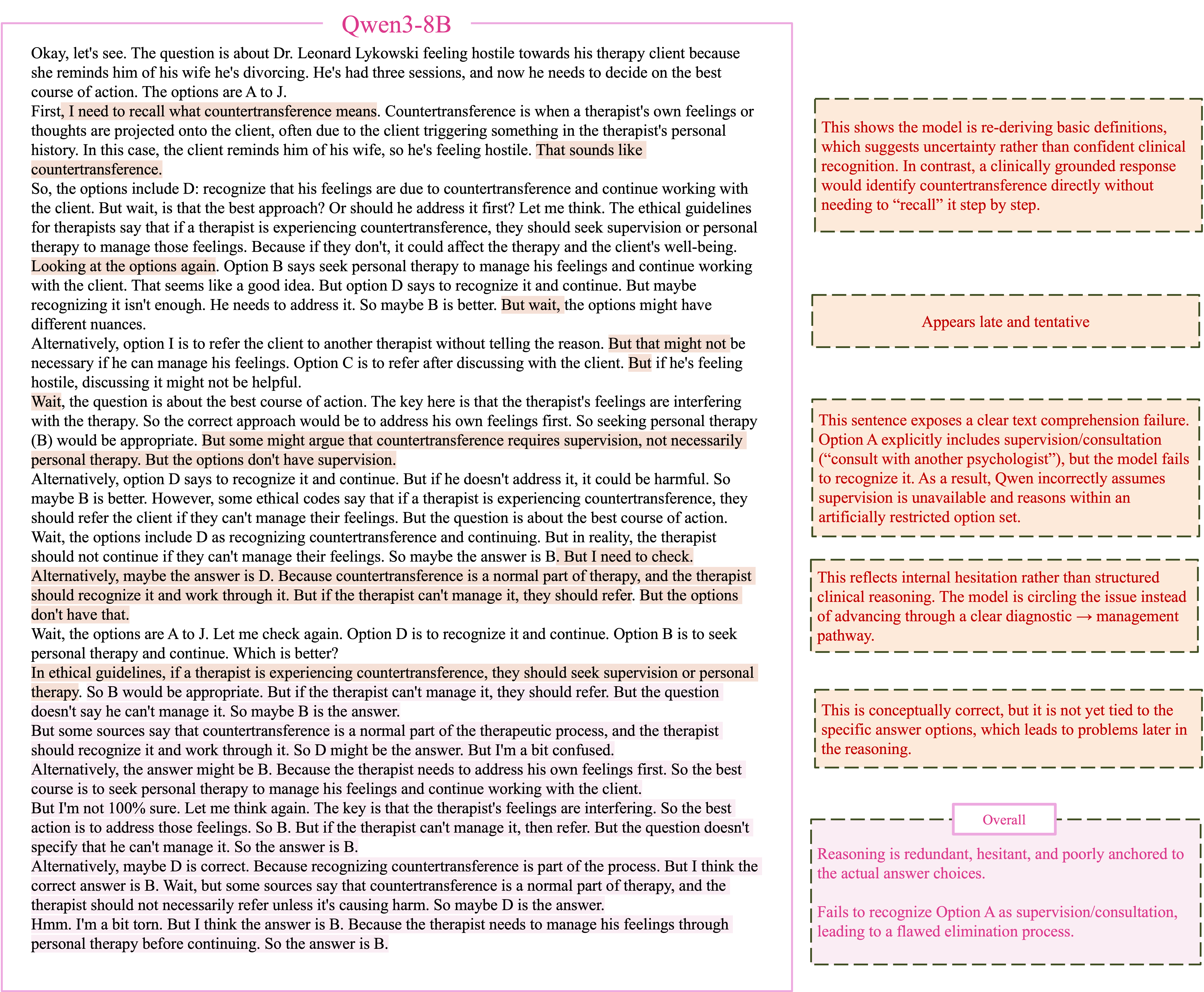}
  \caption{\textbf{Highlighted response from the Qwen3-8B Base model.} The response repeatedly revisits the definition of countertransference and alternates between answer options. Clinician assessment identified hesitation, redundancy and failure to recognize that option A provides consultation with another psychologist.}
  \label{supp-fig-baseline}
\end{figure}

The Base-model response was hesitant and did not select the consultation option; ClinMPO identified countertransference and selected consultation directly (\suppfigsref{supp-fig-baseline}{supp-fig-clinmpo}).

\subsection{Supplementary Note 12 | Documented prompt templates}
\label{supp-note-prompts}
The following templates were used for question generation and ICD-11 classification; programmatic variables appear in braces.

\subsubsection*{Evidence-based multiple-choice question generation}
\begin{lstlisting}[style=prompt]
You are a senior clinical medical educator. Carefully read the following
psychiatric article and generate three high-quality, independent
multiple-choice questions (MCQs) that assess clinical reasoning and medical
knowledge.

Requirements:
1. Each question must be independent and self-contained. Include all
   information required to answer it, including relevant patient background,
   clinical course, laboratory or imaging findings and treatment details.
2. Do not use phrases such as "this patient" or "the above case" that create
   dependencies between questions.
3. Provide four medically plausible options (A-D), with one correct answer.
4. Mark the correct answer and provide a professional explanation of no more
   than 100 words.
5. Questions may address clinical background, pathogenesis, diagnosis,
   differential diagnosis, investigations, treatment or prognosis.
6. Return only a JSON array in the following structure:

[
  {
    "question": "{self-contained question}",
    "options": {
      "A": "{option A}",
      "B": "{option B}",
      "C": "{option C}",
      "D": "{option D}"
    },
    "answer": "{correct option label}",
    "explanation": "{clinical rationale, <=100 words}"
  }
]

Source article:
{full_article_content}
\end{lstlisting}

\subsubsection*{Classification across the two classification dimensions}
The following template was used after initial rule-based screening to assign an ICD-11 diagnostic category as the major category and a psychiatric practice competency as the minor category.

\begin{lstlisting}[style=prompt]
You are a medical classification expert. Analyze the medical question and
classify it using the ICD-11 diagnostic categories and psychiatric practice
competencies supplied with the task.

1. Select the ICD-11 diagnostic category that best matches the question as
   the major category.
2. Select the psychiatric practice competency that best describes the
   reasoning task required to answer the question as the minor category.

Return only the raw JSON object, without Markdown or explanatory text:

{
  "major_category": "{selected ICD-11 diagnostic category}",
  "minor_category": "{selected psychiatric practice competency}"
}

Question:
{question}
\end{lstlisting}

\clearpage
\section{Supplementary Figures}

\phantomsection
\addcontentsline{toc}{subsection}{Supplementary Fig. 1 | Qwen3-8B Base response with clinician annotations}

\clearpage
\phantomsection
\addcontentsline{toc}{subsection}{Supplementary Fig. 2 | Qwen3-8B-ClinMPO response with clinician annotations}
\begin{figure}[H]
  \centering
  \includegraphics[width=0.90\linewidth,height=0.66\textheight,keepaspectratio]{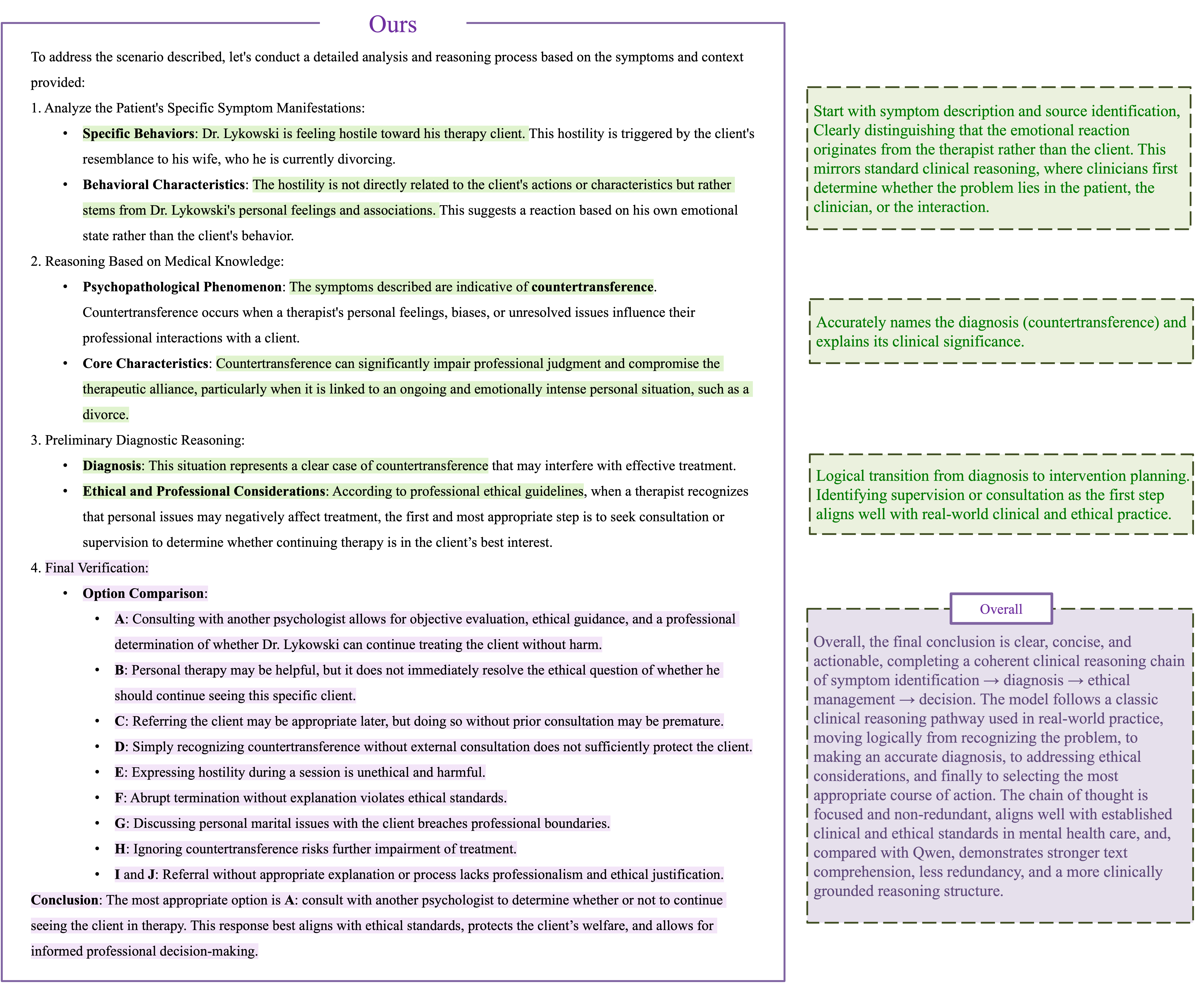}
  \caption{\textbf{Highlighted response from Qwen3-8B-ClinMPO.} The response identifies countertransference, explains its clinical and ethical relevance, compares the available options and selects professional consultation. Clinician assessment highlighted the more direct progression from problem identification to management and final answer selection.}
  \label{supp-fig-clinmpo}
\end{figure}

\clearpage
\phantomsection
\addcontentsline{toc}{subsection}{Supplementary Fig. 3 | Dataset construction and partitioning workflow}
\begin{figure}[H]
  \centering
  \includegraphics[width=\linewidth,height=0.82\textheight,keepaspectratio]{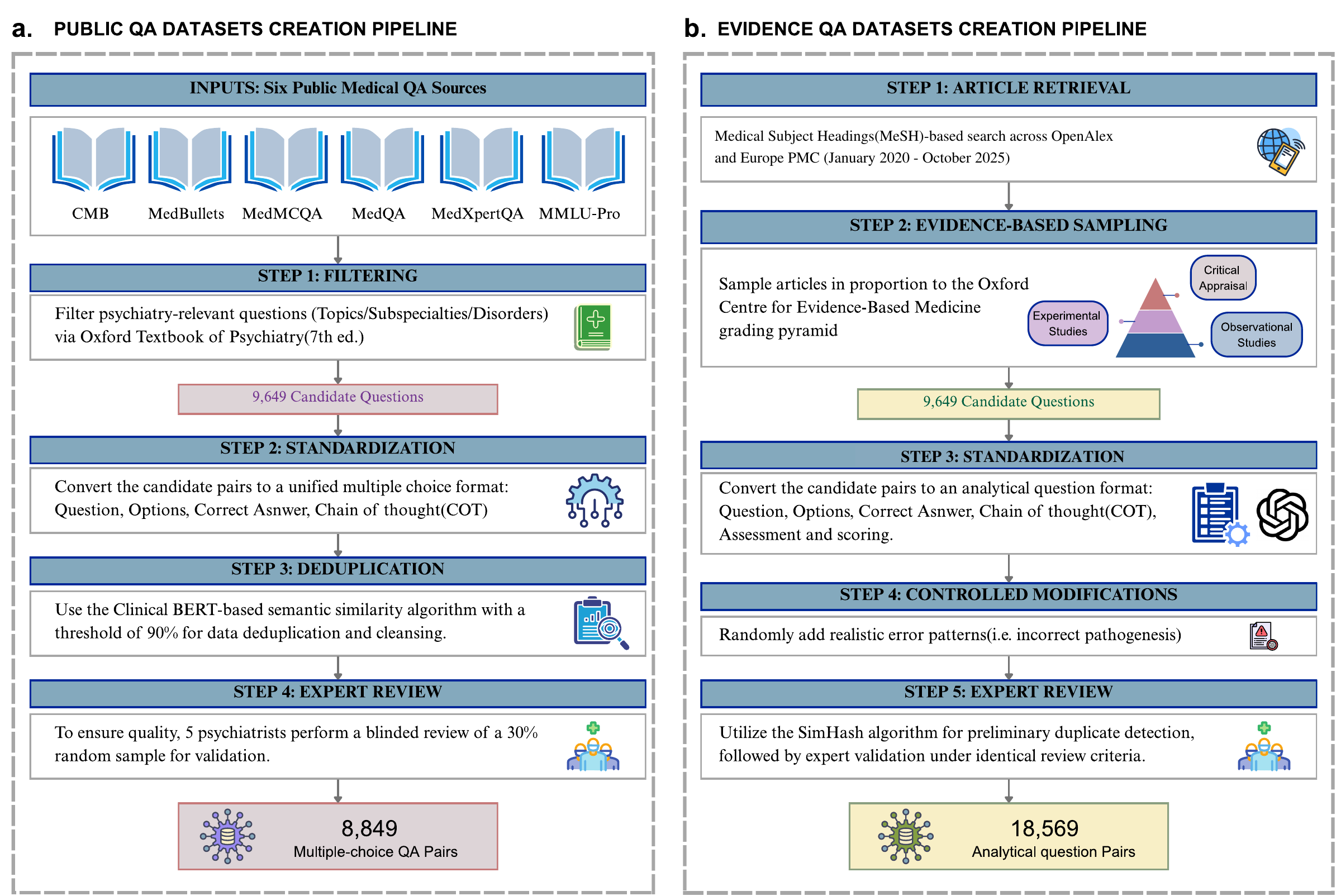}
  \caption{\textbf{Dataset construction and partitioning workflow.} \textbf{a}, Construction of the Public QA Dataset from six public medical question--answering resources through psychiatric-relevance screening, format standardization, semantic deduplication and expert quality review. Questions answered correctly by at least 9 of the 13 models entered the 7,112-item model-screened higher-solvability training set; the remaining 1,737 questions formed the fixed, difficulty-enriched test set. Missing model predictions were treated as incorrect for partitioning. \textbf{b}, Construction of the Evidence QA Dataset from 4,474 psychiatry publications identified through OpenAlex and Europe PMC. Evidence-derived questions and criterion-resolved score records were reviewed and deduplicated, producing 18,569 question--answer pairs for ClinRM development.}
  \label{supp-fig-dataset-construction}
\end{figure}

\clearpage
\phantomsection
\addcontentsline{toc}{subsection}{Supplementary Fig. 4 | Medical-student online question interface}
\begin{figure}[H]
  \centering
  \includegraphics[width=0.68\linewidth,height=0.76\textheight,keepaspectratio]{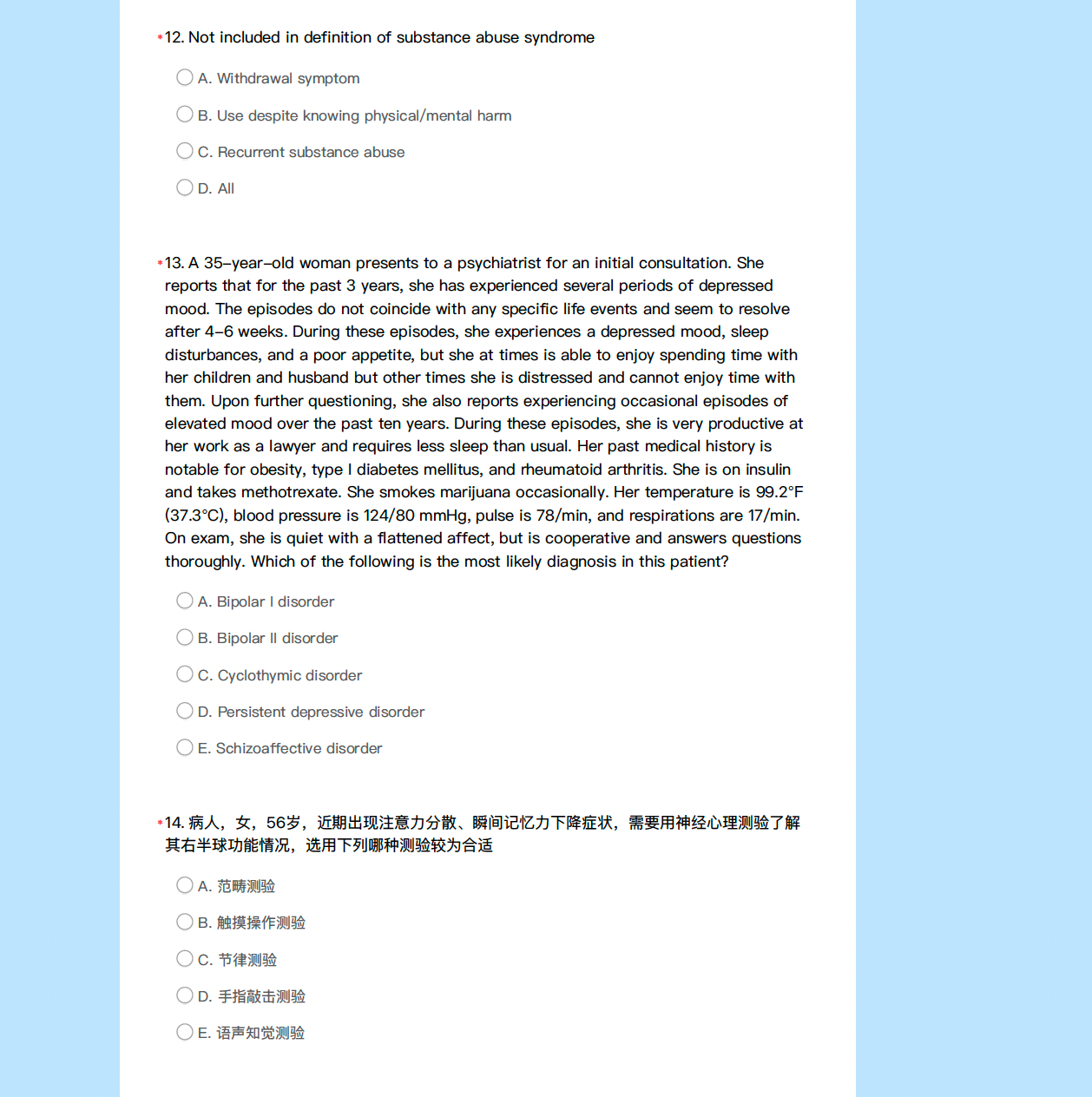}
  \caption{\textbf{Online question interface for the medical-student evaluation at Jining Medical University.} Example of the multiple-choice interface used to present psychiatric test items and answer options. The screenshot illustrates the participant-facing question format used during electronic response collection.}
  \label{supp-fig-student-interface}
\end{figure}

\clearpage
\phantomsection
\addcontentsline{toc}{subsection}{Supplementary Fig. 5 | Medical-student evaluation procedure}
\begin{figure}[H]
  \centering
  \textbf{a}\par\vspace{2pt}
  \includegraphics[width=0.40\linewidth]{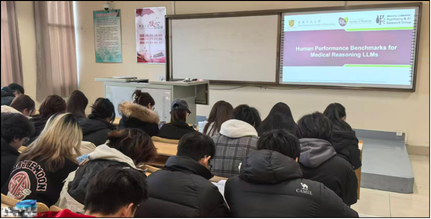}

  \vspace{4pt}
  \textbf{b}\par\vspace{2pt}
  \includegraphics[width=0.40\linewidth]{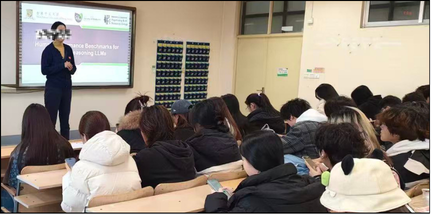}

  \vspace{4pt}
  \textbf{c}\par\vspace{2pt}
  \includegraphics[width=0.40\linewidth]{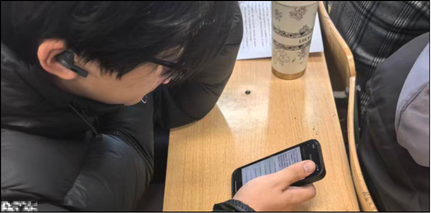}

  \vspace{4pt}
  \textbf{d}\par\vspace{2pt}
  \includegraphics[width=0.40\linewidth]{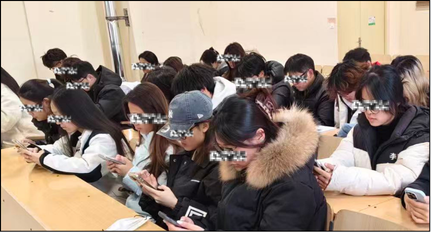}
  \caption{\textbf{Medical-student evaluation procedure at Jining Medical University.} \textbf{a,b}, Classroom briefing and administration of the medical-reasoning assessment. \textbf{c}, Example of a participant submitting responses through the online interface on a mobile device. \textbf{d}, Participants completing assigned questions during a supervised classroom session. The photographs document the administration procedure; faces have been obscured to protect participant privacy.}
  \label{supp-fig-student-evaluation}
\end{figure}

\clearpage
\phantomsection
\addcontentsline{toc}{subsection}{Supplementary Fig. 6 | Clinician evaluation procedure}
\begin{figure}[H]
  \centering
  \textbf{a}\par\vspace{2pt}
  \includegraphics[width=0.55\linewidth]{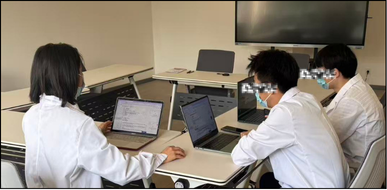}

  \vspace{8pt}
  \textbf{b}\par\vspace{2pt}
  \includegraphics[width=0.55\linewidth]{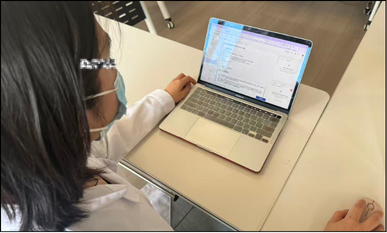}

  \vspace{8pt}
  \textbf{c}\par\vspace{2pt}
  \includegraphics[width=0.55\linewidth]{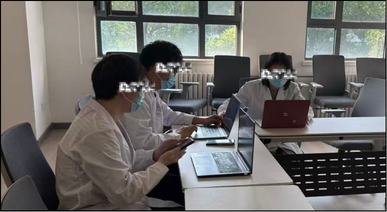}
  \caption{\textbf{Computer-assisted clinician evaluation of generated rationales.} \textbf{a}, Clinicians independently reviewing study materials at separate workstations. \textbf{b}, Example of the electronic scoring workspace used to inspect a generated response and record criterion-level ratings. \textbf{c}, Clinicians completing the blinded evaluation in the assessment setting. The photographs document the evaluation procedure; faces have been obscured to protect evaluator privacy.}
  \label{supp-fig-clinician-evaluation}
\end{figure}

\clearpage
\phantomsection
\addcontentsline{toc}{subsection}{Supplementary Fig. 7 | Clinician-scoring interface for Model A}
\begin{figure}[H]
  \centering
  \includegraphics[width=0.90\linewidth,height=0.68\textheight,keepaspectratio]{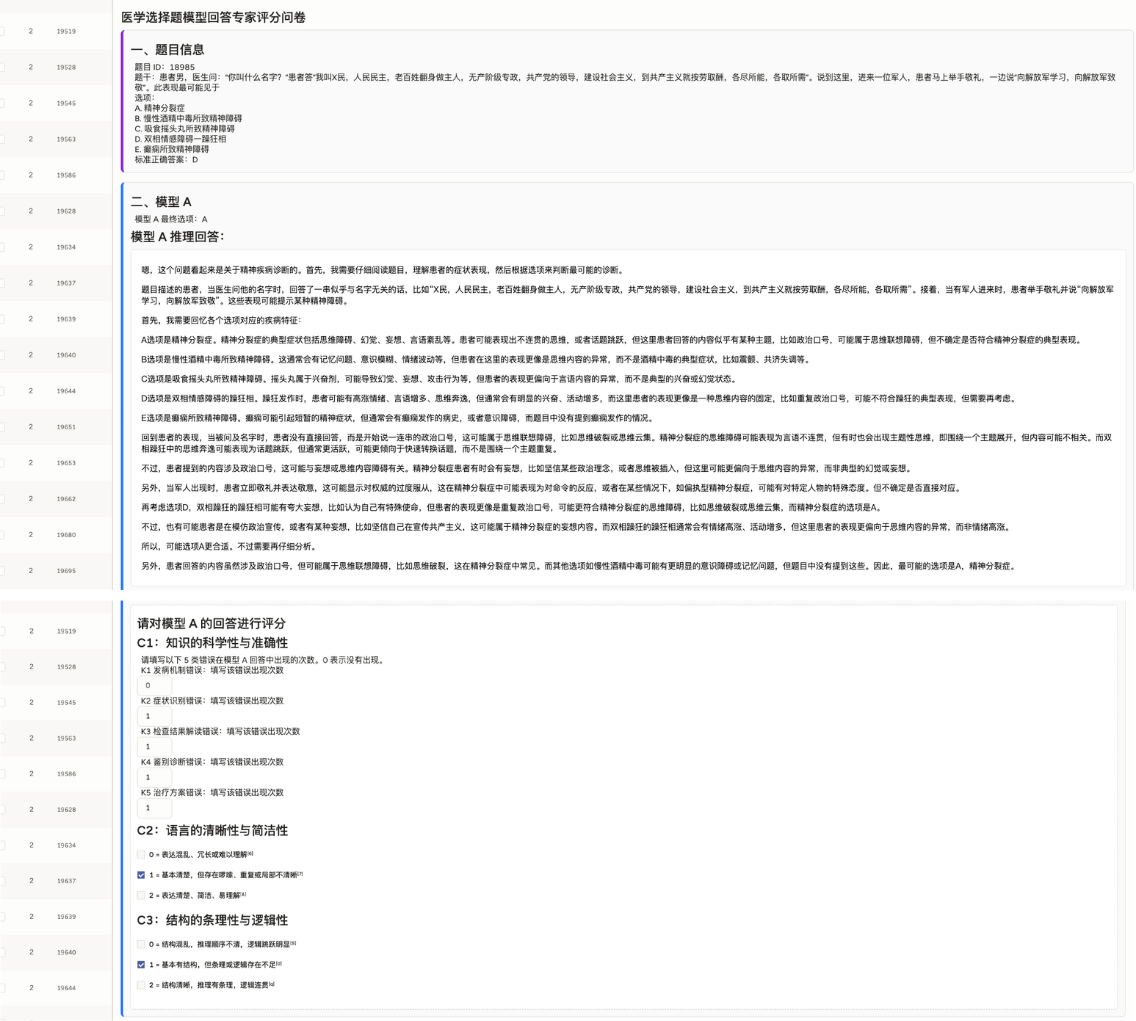}
  \caption{\textbf{Representative electronic clinician-scoring interface for an anonymized Model A response.} The interface displays the psychiatric multiple-choice question, answer options, reference answer and the generated rationale evaluated by the clinician. Criterion-level fields were used to record C1 clinical accuracy and completeness, C2 clarity and concision, and C3 organization and logical coherence. Model labels were anonymized during clinician evaluation.}
  \label{supp-fig-clinician-model-a-interface}
\end{figure}

\clearpage
\phantomsection
\addcontentsline{toc}{subsection}{Supplementary Fig. 8 | Clinician-scoring interface for Model B}
\begin{figure}[H]
  \centering
  \includegraphics[width=0.90\linewidth,height=0.68\textheight,keepaspectratio]{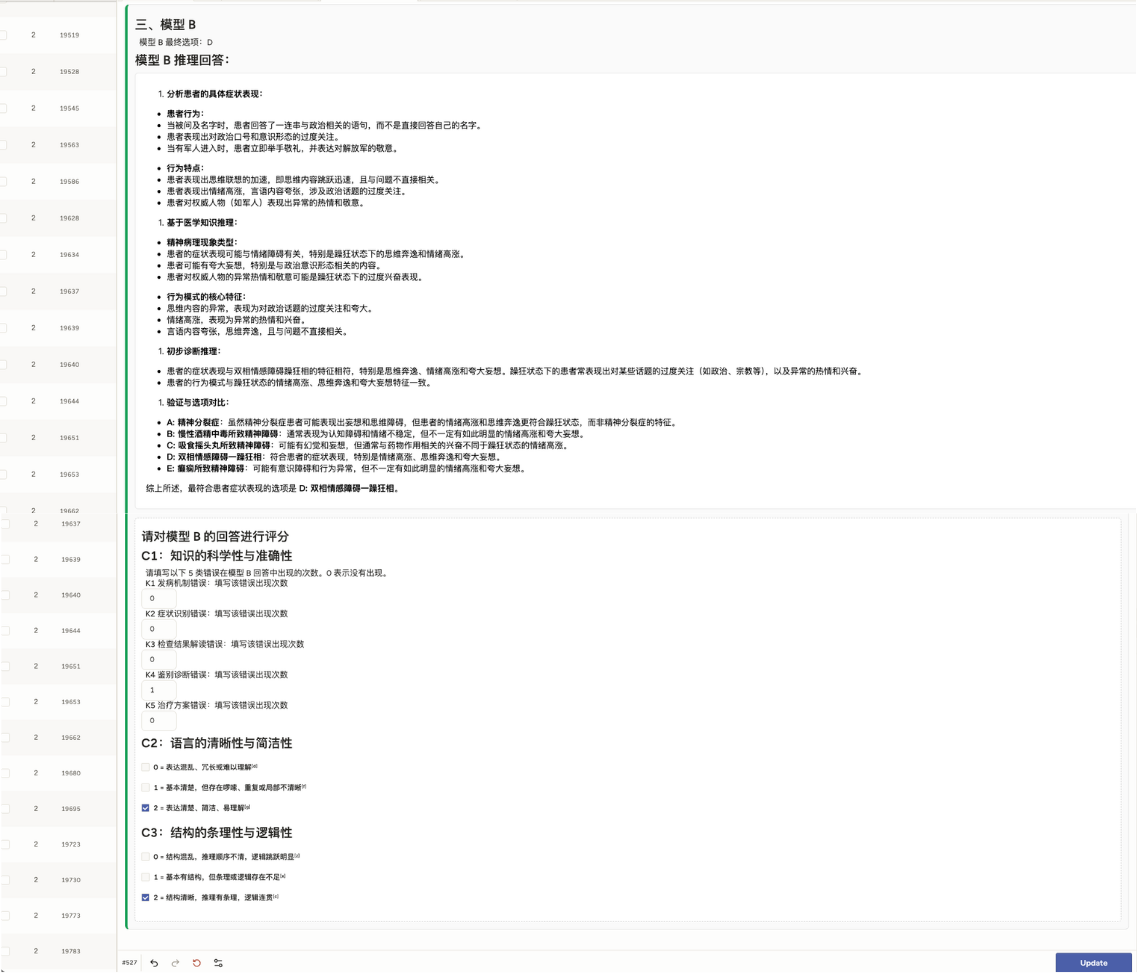}
  \caption{\textbf{Representative electronic clinician-scoring interface for an anonymized Model B response.} The interface displays the psychiatric multiple-choice question, answer options, reference answer and the generated rationale evaluated by the clinician. The same criterion-level fields were used to record C1 clinical accuracy and completeness, C2 clarity and concision, and C3 organization and logical coherence. Model labels were anonymized during clinician evaluation.}
  \label{supp-fig-clinician-model-b-interface}
\end{figure}

\clearpage
\phantomsection
\addcontentsline{toc}{subsection}{Supplementary Fig. 9 | Independent ClinRM validation against clinician consensus}
\begin{figure}[H]
  \centering
  \includegraphics[width=\linewidth,height=0.82\textheight,keepaspectratio]{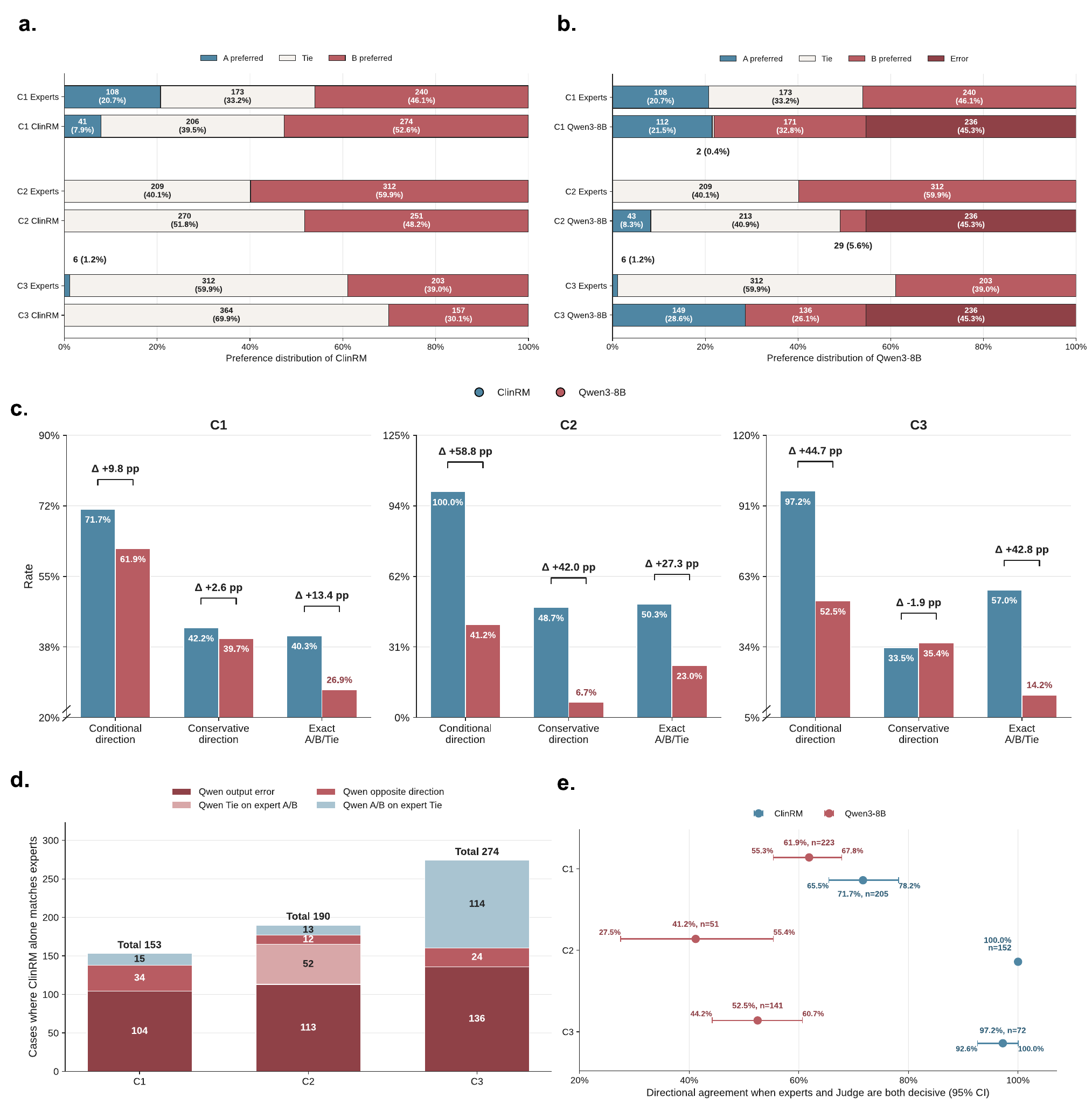}
  \caption{\textbf{Independent validation of ClinRM against clinician consensus.} Three clinicians independently assessed paired Qwen3-8B and Qwen3-8B-ClinMPO responses for 521 held-out questions using the C1, C2 and C3 rubric. \textbf{a}, Preference distributions for clinician consensus and ClinRM. \textbf{b}, Preference and output-validity distributions for the Qwen3-8B judge. \textbf{c}, Conditional directional agreement, conservative directional agreement and exact three-class agreement for ClinRM and the Qwen3-8B judge; points and error bars denote estimates and 95\% bootstrap confidence intervals. \textbf{d}, Sources of cases in which ClinRM agreed with clinicians and the Qwen3-8B judge did not. \textbf{e}, Directional agreement with clinician consensus, restricted to questions for which both the judge and clinicians made explicit A/B choices.}
  \label{supp-fig-clinrm-validation}
\end{figure}

\clearpage
\phantomsection
\addcontentsline{toc}{subsection}{Supplementary Fig. 10 | Category-level accuracy profiles after sparse-category exclusion}
\setcounter{figure}{9}
\begin{figure}[H]
  \centering
  \textbf{a}\par\vspace{2pt}
  \includegraphics[width=\linewidth,height=0.70\textheight,keepaspectratio]{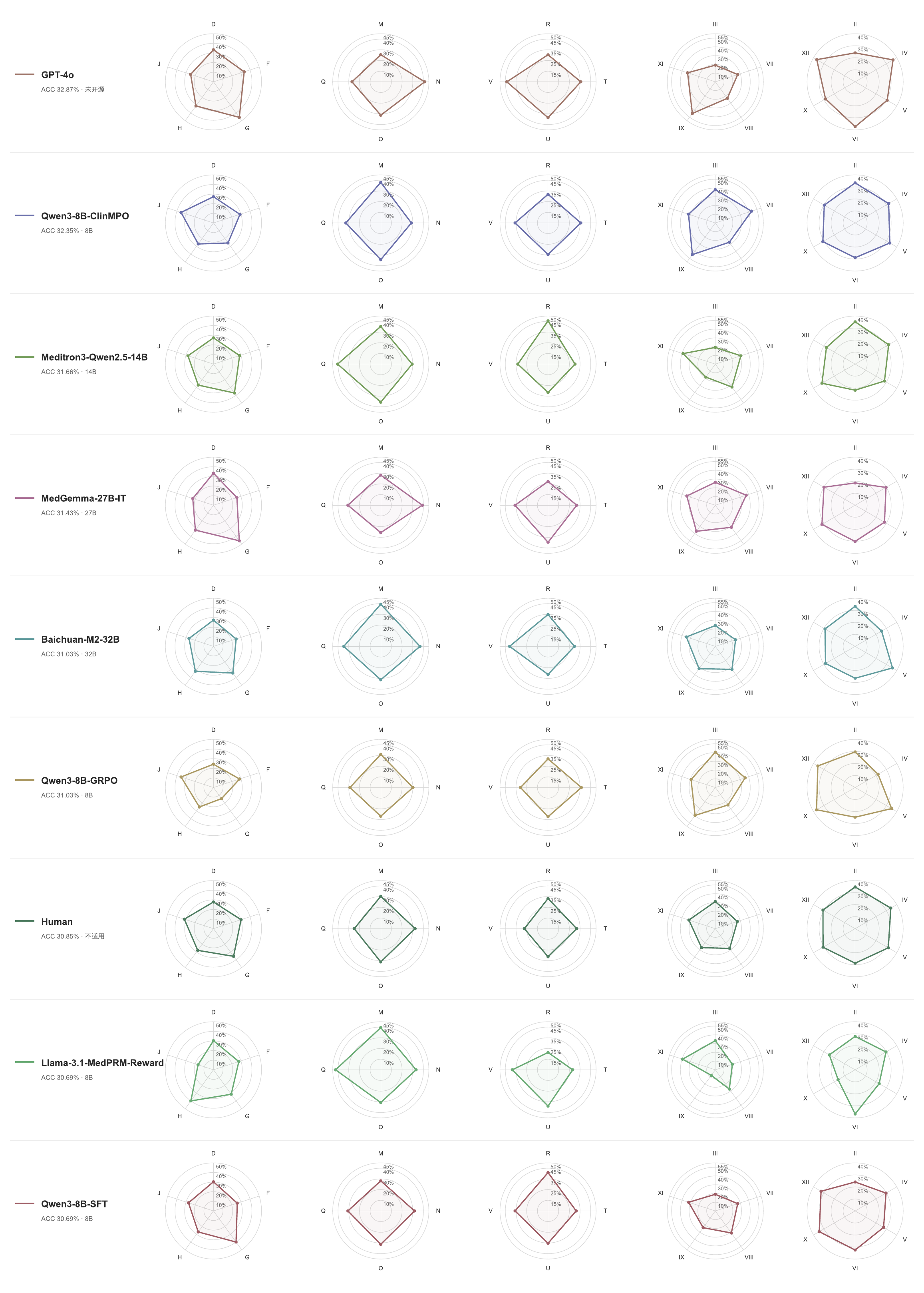}
  \caption{\textbf{Category-level accuracy profiles after sparse-category exclusion.} \textbf{a--d}, Evaluated models ordered by overall test-set accuracy and displayed across four consecutive pages. Each row gives the model's overall accuracy and parameter count, followed by radar plots of accuracy across the retained categories in the two classification dimensions. Categories represented by fewer than 30 questions were excluded from the primary summaries to avoid emphasizing highly unstable estimates from sparsely represented categories. Category codes are defined in Supplementary Tables 5 and 6.}
  \label{supp-fig-category-effects}
\end{figure}

\clearpage
\begin{figure}[H]
  \centering
  \textbf{b}\par\vspace{2pt}
  \includegraphics[width=\linewidth,height=0.82\textheight,keepaspectratio]{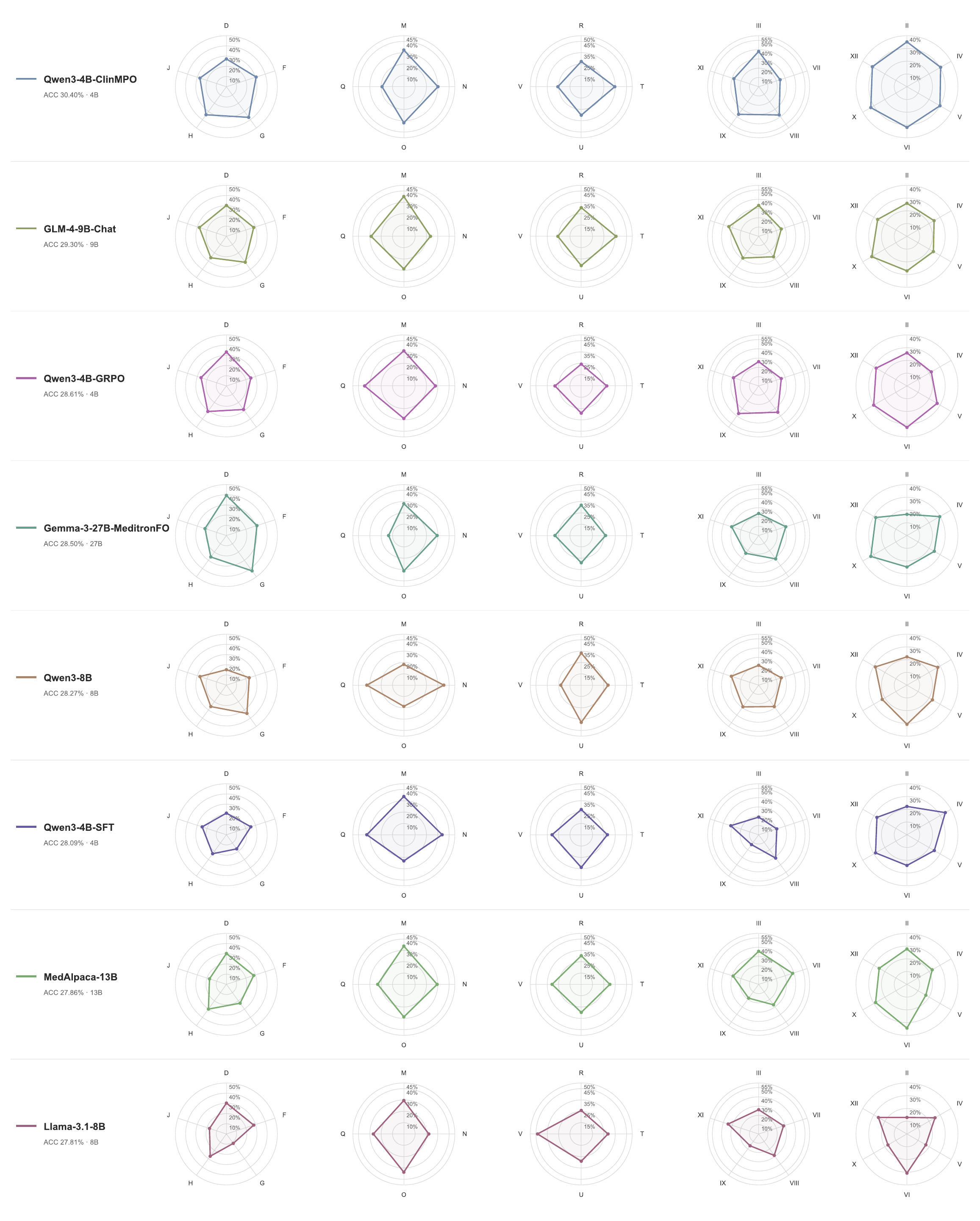}
\end{figure}

\clearpage
\begin{figure}[H]
  \centering
  \textbf{c}\par\vspace{2pt}
  \includegraphics[width=\linewidth,height=0.82\textheight,keepaspectratio]{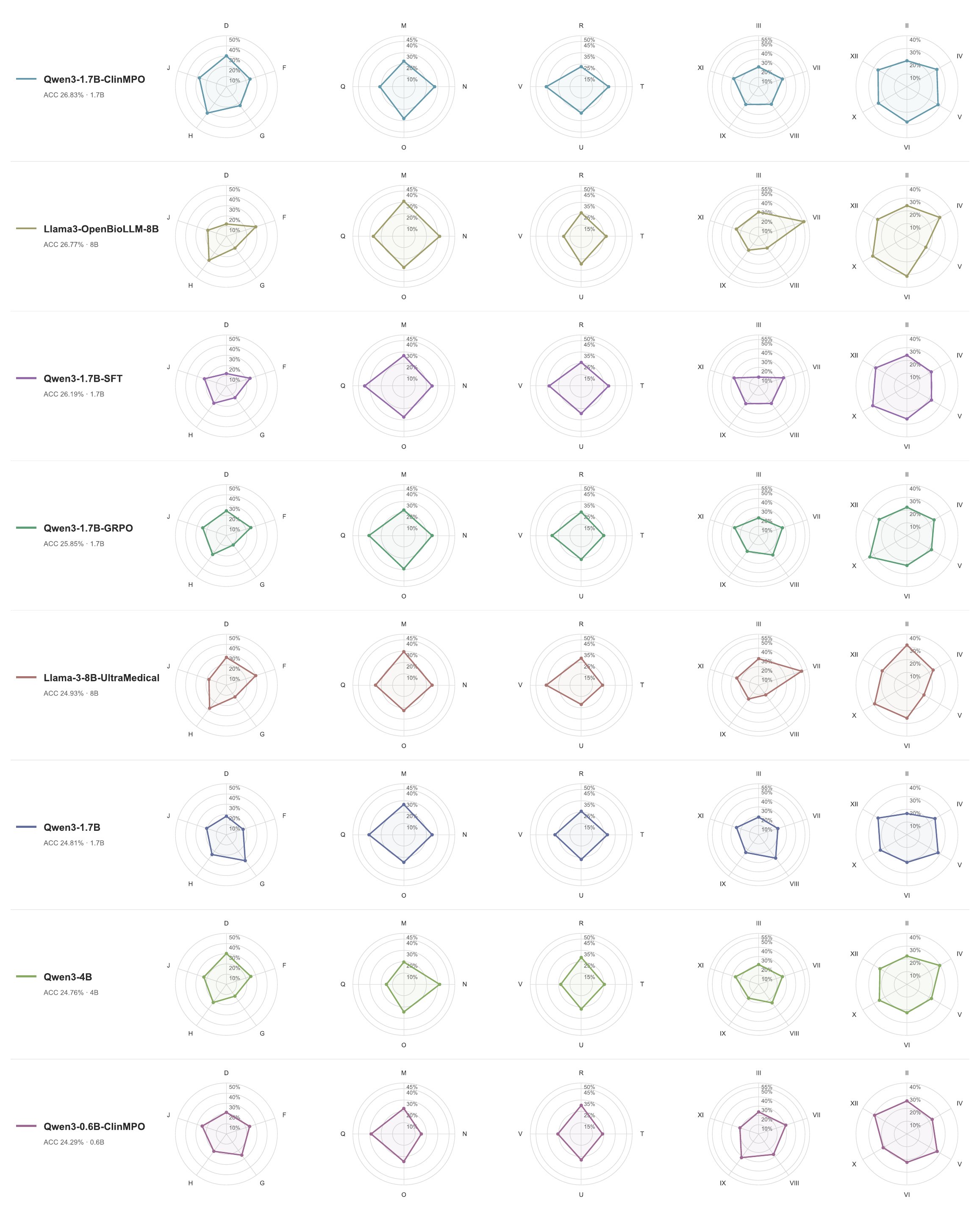}
\end{figure}

\clearpage
\begin{figure}[H]
  \centering
  \textbf{d}\par\vspace{2pt}
  \includegraphics[width=\linewidth,height=0.82\textheight,keepaspectratio]{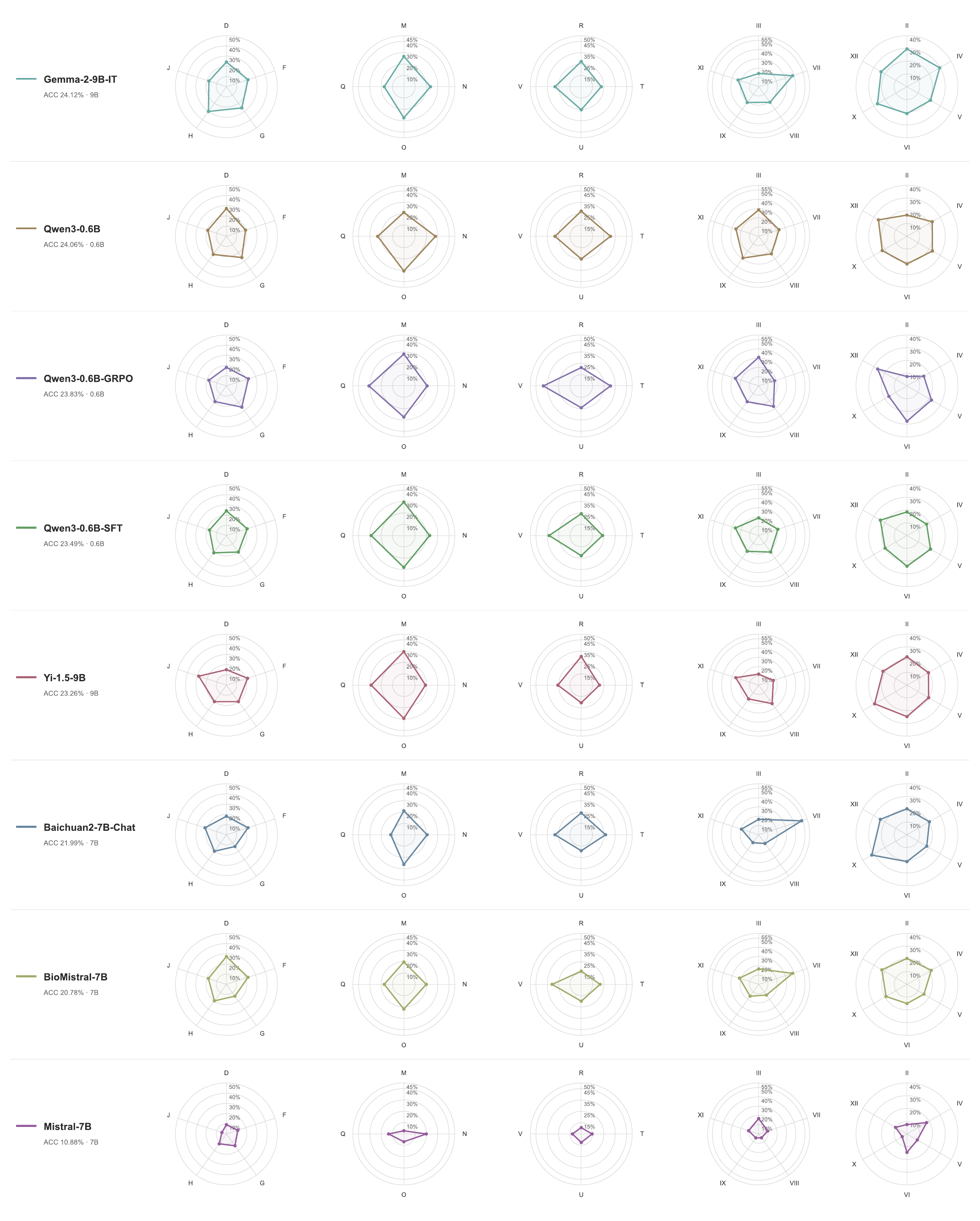}
\end{figure}

\clearpage
\phantomsection
\addcontentsline{toc}{subsection}{Supplementary Fig. 11 | Category-level accuracy across training strategies}
\begin{figure}[H]
  \centering
  \textbf{a}\par\vspace{2pt}
  \includegraphics[width=\linewidth,height=0.70\textheight,keepaspectratio]{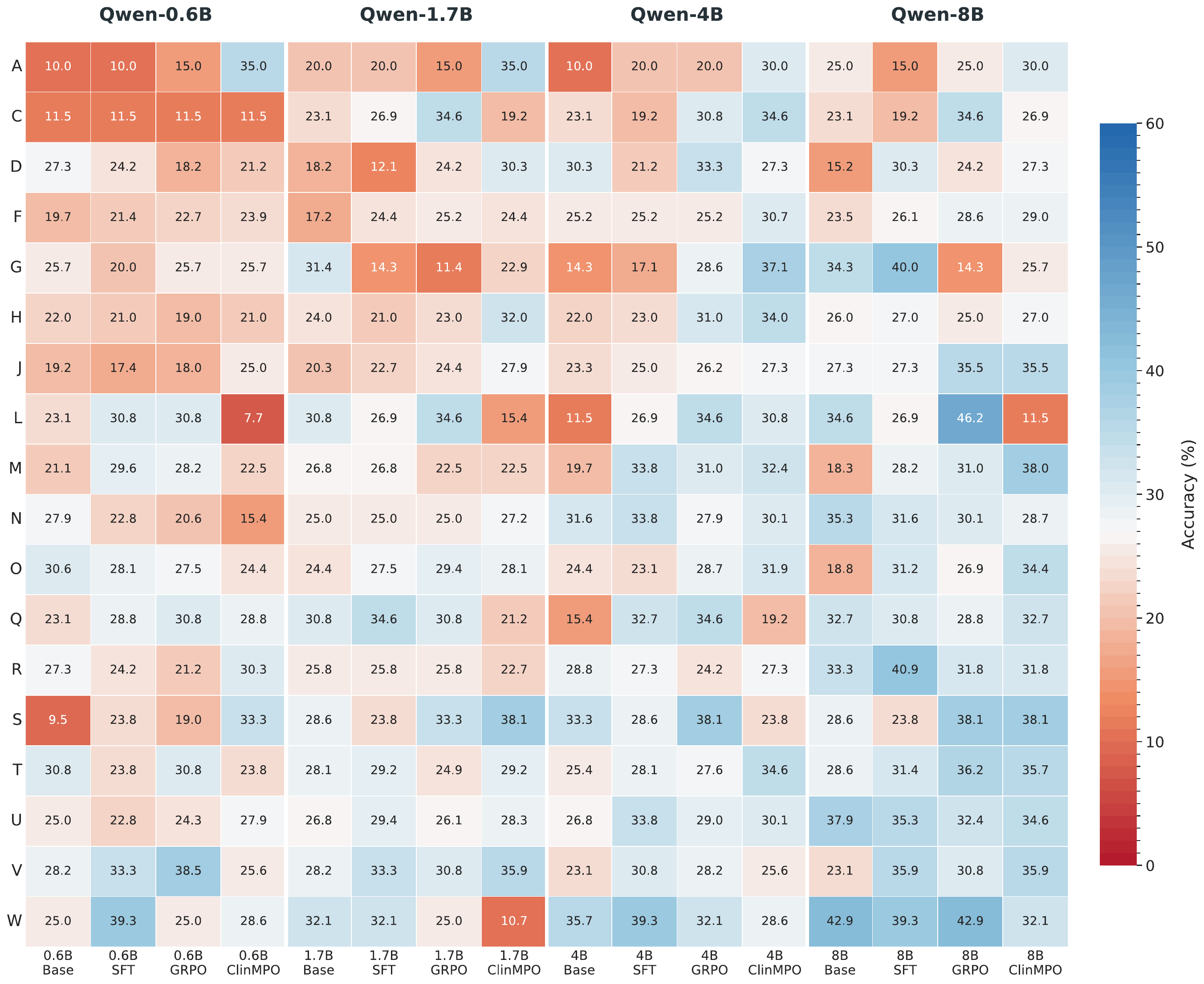}
  \caption{\textbf{Category-level accuracy across model sizes and training strategies.} \textbf{a}, Accuracy across ICD-11 diagnostic categories. \textbf{b}, Accuracy across psychiatric practice competencies. Columns are grouped by Qwen model size (0.6B, 1.7B, 4B and 8B); within each size, Base, supervised fine-tuning (SFT), group relative policy optimization (GRPO) and ClinMPO are shown from left to right. Cell values and color intensity denote accuracy in percent. Row codes correspond to the category definitions in Supplementary Tables 5 and 6.}
  \label{supp-fig-training-category-heatmaps}
\end{figure}

\clearpage
\begin{figure}[H]
  \centering
  \vspace*{0.08\textheight}
  \textbf{b}\par\vspace{2pt}
  \includegraphics[width=\linewidth,height=0.72\textheight,keepaspectratio]{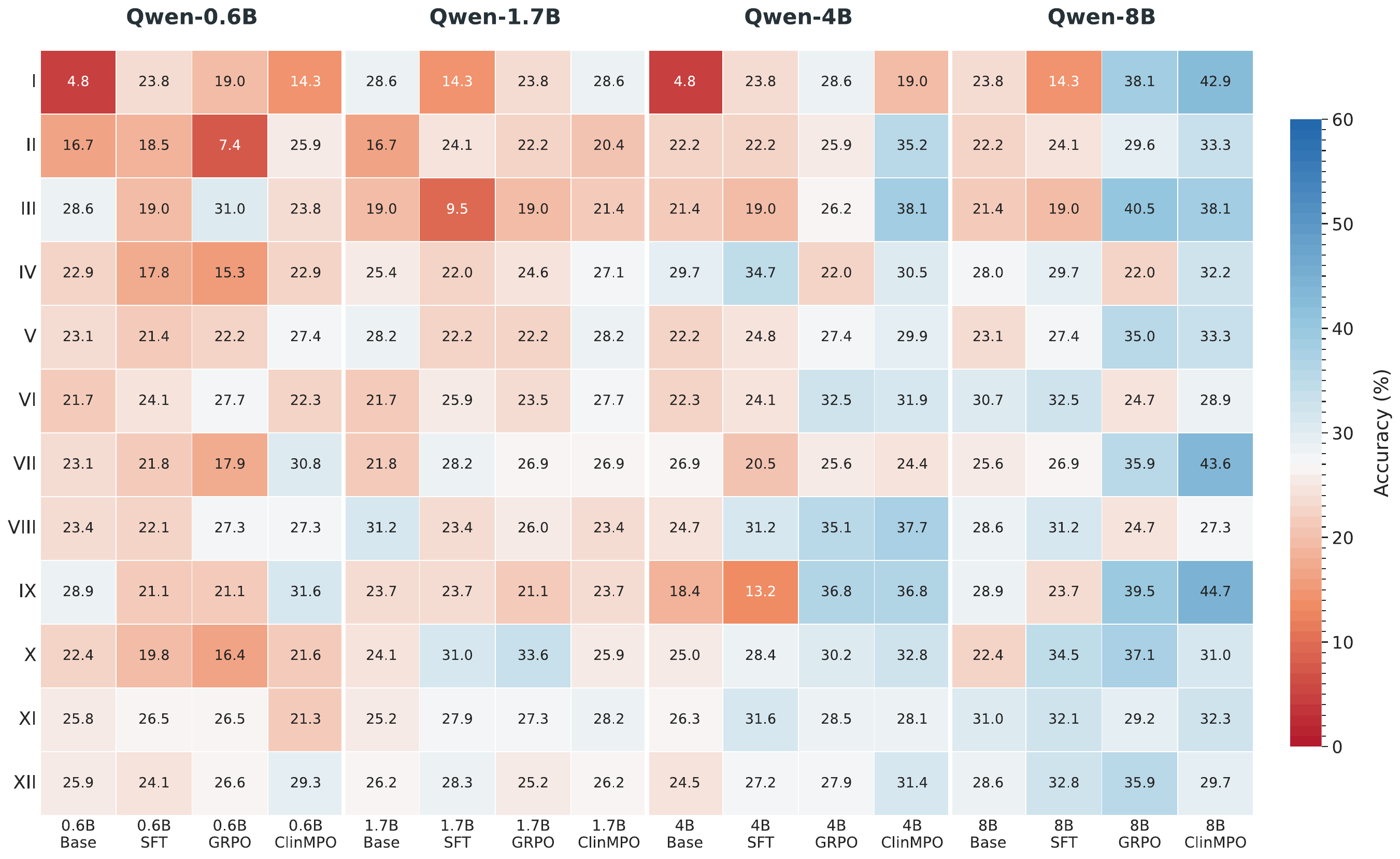}
  \caption*{\textbf{Supplementary Fig. 11 (continued).}}
\end{figure}

\clearpage
\phantomsection
\addcontentsline{toc}{subsection}{Supplementary Fig. 12 | Cross-classification of test-set questions}
\begin{figure}[H]
  \centering
  \vspace*{0.08\textheight}
  \includegraphics[width=\linewidth,height=0.68\textheight,keepaspectratio]{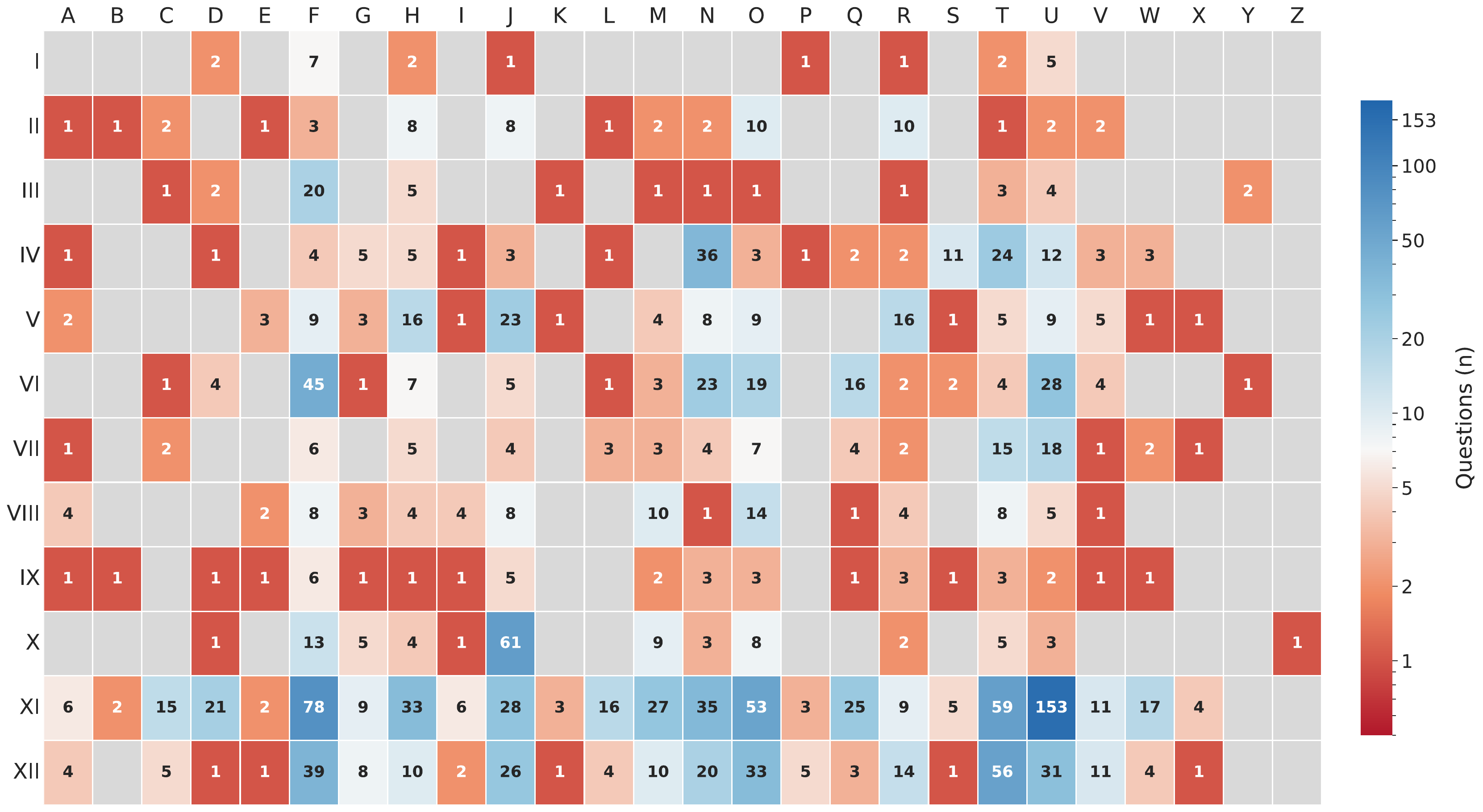}
  \caption{\textbf{Cross-classification of questions in the fixed 1,737-item test set.} Columns denote the ICD-11 diagnostic categories (codes A--Z), and rows denote the psychiatric practice competencies (codes I--XII). Each annotated cell gives the number of test questions assigned jointly to the corresponding diagnostic category and competency. Gray cells indicate no questions for that combination; the color bar reports the observed question count, \(n\). Category codes are defined in Supplementary Tables 5 and 6.}
  \label{supp-fig-test-set-category-cross-count}
\end{figure}

\clearpage
\phantomsection
\addcontentsline{toc}{subsection}{Supplementary Fig. 13 | Cross-classification of training-set questions}
\begin{figure}[H]
  \centering
  \vspace*{0.08\textheight}
  \includegraphics[width=\linewidth,height=0.68\textheight,keepaspectratio]{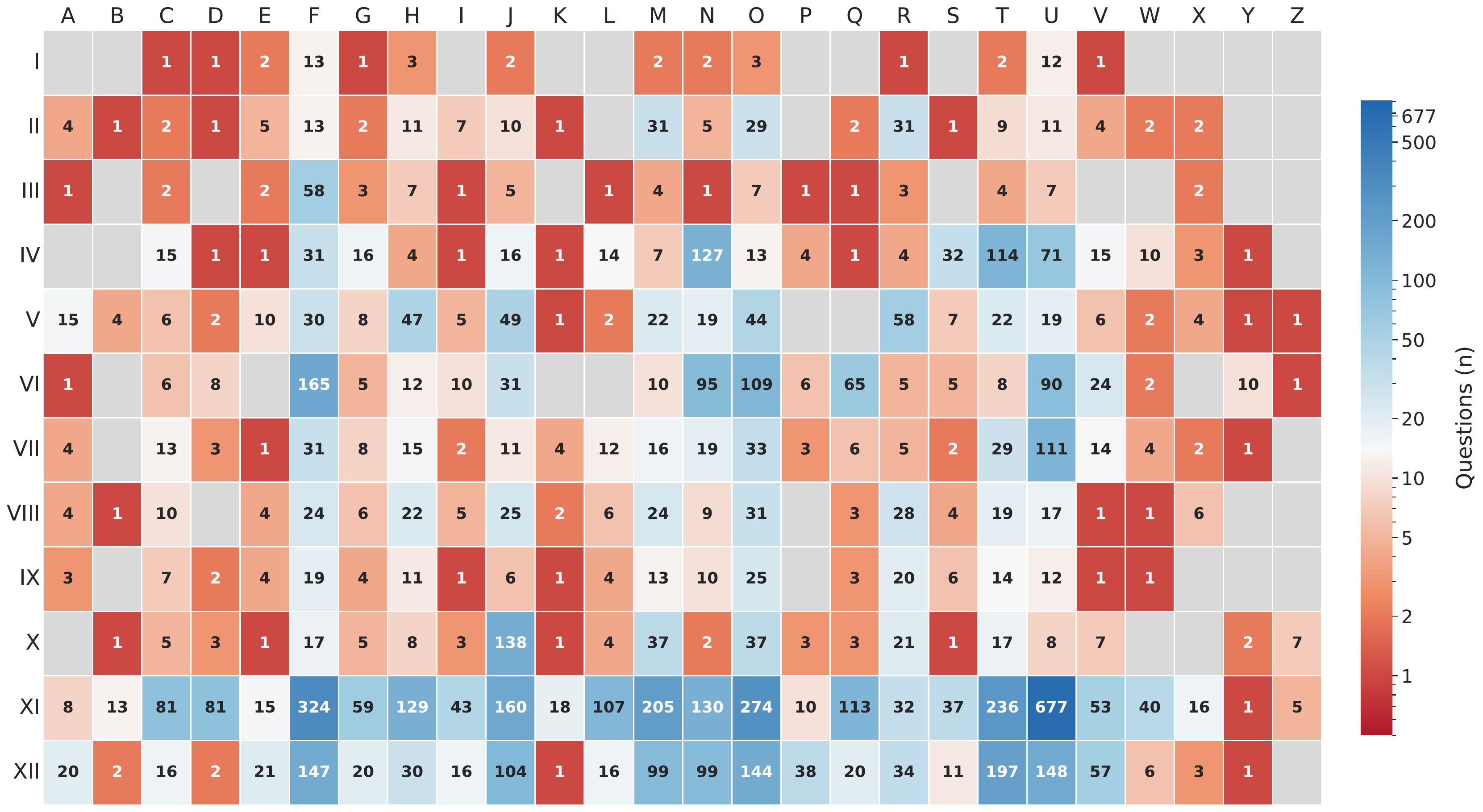}
  \caption{\textbf{Cross-classification of questions in the fixed 7,112-item higher-solvability training set.} Columns denote the ICD-11 diagnostic categories (codes A--Z), and rows denote the psychiatric practice competencies (codes I--XII). Each annotated cell gives the number of training questions assigned jointly to the corresponding diagnostic category and competency. Gray cells indicate no questions for that combination; the color bar reports the observed question count, \(n\). Category codes are defined in Supplementary Tables 5 and 6.}
  \label{supp-fig-training-set-category-cross-count}
\end{figure}

\clearpage
\section{Supplementary Tables}

\phantomsection
\addcontentsline{toc}{subsection}{Supplementary Table 1 | Public QA provenance and licensing}

\begingroup
\small
\singlespacing
\begin{longtable}{@{}p{0.14\textwidth}p{0.24\textwidth}p{0.14\textwidth}p{0.37\textwidth}@{}}
\caption{\textbf{Provenance and licensing of the public question--answering resources.}}\label{supp-tab-provenance}\\
\toprule
\textbf{Resource} & \textbf{Official repository} & \textbf{License} & \textbf{Complete source description} \\
\midrule
\endfirsthead
\multicolumn{4}{l}{\tablename~\thetable\ continued}\\
\toprule
\textbf{Resource} & \textbf{Official repository} & \textbf{License} & \textbf{Complete source description} \\
\midrule
\endhead
\bottomrule
\endfoot
CMB & FreedomIntelligence/CMB & Apache-2.0 & 269,359 training, 280 validation and 11,200 test questions in CMB-Exam; six major categories and 28 subcategories. \\
MedBulletsQA & LangAGI-Lab/medbullets & Original provider and source-site terms & 492 training and 124 test records at the time of access. \\
MedMCQA & medmcqa/medmcqa & MIT & More than 194,000 questions covering approximately 2,400 healthcare topics and 21 medical subjects. \\
MedQA & jind11/MedQA & MIT & Medical-examination question collections from the United States, mainland China and Taiwan, with official train, development and test splits. \\
MedXpertQA & TsinghuaC3I/MedXpertQA & MIT & 4,460 questions spanning 17 specialties and 11 body systems. \\
MMLU-Pro & TIGER-AI-Lab/MMLU-Pro & Apache-2.0 & Official enhanced multi-task language-understanding benchmark; the psychiatry-relevant subset was selected by the study pipeline. \\
\end{longtable}
\endgroup

\phantomsection
\addcontentsline{toc}{subsection}{Supplementary Table 2 | CPTS rubric and scoring anchors}
The Clinical Psychiatry Thinking Strategy (CPTS) separates clinically consequential content requirements from general presentation characteristics. Five psychiatry experts developed the rubric with reference to the \textit{Shorter Oxford Textbook of Psychiatry}, 7th edition. Each criterion was applied to the generated rationale in the context of the question, answer options and reference answer.

\begingroup
\footnotesize
\singlespacing
\renewcommand{\arraystretch}{1.18}
\begin{longtable}{@{}p{0.088\textwidth}p{0.185\textwidth}p{0.205\textwidth}p{0.205\textwidth}p{0.205\textwidth}@{}}
\caption{\textbf{Complete CPTS rubric, criterion definitions and scoring anchors.} C1 criteria receive 0 or +2; C2 and C3 receive -1 or +1. Examples are schematic scoring anchors rather than additional study observations.}\label{supp-tab-cpts}\\
\toprule
\textbf{Code} & \textbf{Construct assessed} & \textbf{Lower-score anchor} & \textbf{Higher-score anchor} & \textbf{Illustrative negative and positive examples} \\
\midrule
\endfirsthead
\multicolumn{5}{l}{\tablename~\thetable\ continued}\\
\toprule
\textbf{Code} & \textbf{Construct assessed} & \textbf{Lower-score anchor} & \textbf{Higher-score anchor} & \textbf{Illustrative negative and positive examples} \\
\midrule
\endhead
\bottomrule
\endfoot
C1-$k_1$ & Pathogenesis: whether the causal or mechanistic account is clinically accurate and relevant. & 0: the required mechanism is absent, contradicted or described incorrectly. & +2: the response correctly identifies and applies the relevant mechanism without a material causal error. & Negative: attributes a medication adverse effect to the underlying disorder. Positive: distinguishes the treatment-related mechanism from disease progression. \\
C1-$k_2$ & Symptom identification: recognition and characterization of clinically relevant symptoms and signs. & 0: a key symptom is omitted, misidentified or assigned the wrong clinical significance. & +2: the response identifies the relevant symptom pattern and interprets it consistently with the case. & Negative: treats reduced need for sleep as ordinary insomnia. Positive: recognizes reduced need for sleep with increased activity as a manic symptom pattern. \\
C1-$k_3$ & Examination interpretation: interpretation of mental-state, physical, laboratory, imaging or other examination findings. & 0: a material finding is ignored, reversed or interpreted outside its clinical context. & +2: the response interprets the relevant finding accurately and links it appropriately to diagnosis or management. & Negative: describes a normal result as confirming pathology. Positive: uses the result to support or narrow the differential diagnosis. \\
C1-$k_4$ & Differential diagnosis: comparison of plausible alternatives and discrimination using case evidence. & 0: important alternatives are omitted, an implausible diagnosis is prioritized or distinguishing evidence is misused. & +2: the response considers relevant alternatives and explains why the evidence favors the selected diagnosis. & Negative: names a diagnosis without addressing a close medical or psychiatric mimic. Positive: compares the leading alternatives using onset, course and examination findings. \\
C1-$k_5$ & Treatment planning: appropriateness, safety, sequence and justification of management. & 0: the plan is incorrect, unsafe, contraindicated, inadequately prioritized or inconsistent with the case. & +2: the response selects an appropriate and safe intervention and explains its timing or priority. & Negative: recommends treatment despite a stated contraindication. Positive: selects the indicated first-line action while accounting for immediate risk and follow-up. \\
C2 & Clarity and concision of language. This criterion captures redundancy, incomplete expression, unnecessary repetition and avoidable ambiguity. & -1: a linguistic problem is present, including repetitive reconsideration, incomplete statements, excessive verbosity or unclear wording. & +1: the rationale is direct, complete and concise, with no material redundancy or ambiguity. & Negative: repeatedly revisits the same options without resolving the comparison. Positive: states the relevant evidence, comparison and conclusion once in precise language. \\
C3 & Organization and logical coherence. This criterion captures ordering, internal consistency and progression from evidence to conclusion. & -1: a structural problem is present, including illogical ordering, contradiction, disconnected claims or a conclusion unsupported by the preceding reasoning. & +1: the rationale follows a coherent clinical sequence and the conclusion is supported by the preceding analysis. & Negative: gives a conclusion before presenting contradictory evidence. Positive: proceeds from findings to interpretation, differential diagnosis, management and final answer. \\
\end{longtable}
\endgroup

For ClinRM training and policy optimization, the CPTS reward was the sum of all criterion-level scores, $R_{\mathrm{CPTS}}(s,a)=\sum_{j=1}^{5}s_j(s,a)+s_{C2}(s,a)+s_{C3}(s,a)$. The five C1 dimensions therefore contributed a possible total of 0 to +10, whereas C2 and C3 each contributed -1 or +1. This weighting gave clinically grounded accuracy and completeness greater influence than generic linguistic or structural qualities.

\phantomsection
\addcontentsline{toc}{subsection}{Supplementary Table 3 | Inter-rater reliability of clinician reference ratings}
\begingroup
\small
\singlespacing
\setlength{\tabcolsep}{7pt}
\renewcommand{\arraystretch}{1.15}
\begin{longtable}{@{}>{\centering\arraybackslash}p{0.10\textwidth}>{\raggedright\arraybackslash}p{0.42\textwidth}>{\centering\arraybackslash}p{0.16\textwidth}>{\centering\arraybackslash}p{0.20\textwidth}@{}}
\caption{\textbf{Inter-rater reliability of the three clinicians providing reference ratings for ClinRM validation.} Fleiss' kappa was calculated for the binary K1--K5 error labels, and ordinal Krippendorff's alpha was calculated for C2 and C3. Confidence intervals were obtained from cluster-bootstrap resampling using \texttt{question\_id} as the sampling unit. Agreement coefficients ranged from 0.632 to 0.908. These estimates support the use of the aggregated clinician ratings as the reference for ClinRM validation.}\label{supp-tab-expert-interrater-reliability}\\
\toprule
\textbf{Dimension} & \textbf{Agreement statistic} & \textbf{Coefficient} & \textbf{95\% CI} \\
\midrule
\endfirsthead
\multicolumn{4}{l}{\tablename~\thetable\ continued}\\
\toprule
\textbf{Dimension} & \textbf{Agreement statistic} & \textbf{Coefficient} & \textbf{95\% CI} \\
\midrule
\endhead
\bottomrule
\endfoot
K1 & Fleiss' kappa (binary) & 0.729 & 0.623--0.817 \\
K2 & Fleiss' kappa (binary) & 0.632 & 0.515--0.737 \\
K3 & Fleiss' kappa (binary) & 0.794 & 0.743--0.839 \\
K4 & Fleiss' kappa (binary) & 0.894 & 0.856--0.927 \\
K5 & Fleiss' kappa (binary) & 0.768 & 0.716--0.816 \\
C2 & Krippendorff's alpha (ordinal) & 0.867 & 0.834--0.897 \\
C3 & Krippendorff's alpha (ordinal) & 0.908 & 0.877--0.934 \\
\end{longtable}
\endgroup

\phantomsection
\addcontentsline{toc}{subsection}{Supplementary Table 4 | Training and inference configuration}
\begingroup
\footnotesize
\singlespacing
\renewcommand{\arraystretch}{1.16}
\setcounter{table}{3}
\begin{longtable}{@{}p{0.18\textwidth}p{0.23\textwidth}p{0.22\textwidth}p{0.29\textwidth}@{}}
\caption{\textbf{Training and inference configuration.}}\label{supp-tab-training-configuration}\\
\toprule
\textbf{Stage} & \textbf{Item} & \textbf{Recorded value} & \textbf{Implementation detail or reporting note} \\
\midrule
\endfirsthead
\multicolumn{4}{l}{\tablename~\thetable\ continued}\\
\toprule
\textbf{Stage} & \textbf{Item} & \textbf{Recorded value} & \textbf{Implementation detail or reporting note} \\
\midrule
\endhead
\bottomrule
\endfoot
Hardware & Accelerator server & Eight NVIDIA A100 GPUs; 80 GB memory per GPU & Single computing server running Ubuntu 24.04.1 LTS; the complete hardware and system configuration is available in the ClinMPO repository (\clinmporepo). \\
Environment & SFT software & Python 3.10; PyTorch 2.2.2; LLaMA-Factory & Separate Conda environment used for supervised fine-tuning. \\
Environment & Policy-optimization software & Python 3.10; PyTorch 2.7.1+cu126; Verl & Separate Conda environment used for GRPO and ClinMPO policy optimization. \\
Environment & Inference software & Python 3.10; PyTorch 2.7.1+cu126; vLLM & Separate Conda environment used for model evaluation and response generation. \\
ClinRM & Backbone and training data & Qwen3-8B; 18,569 Evidence QA pairs & ClinRM was trained by supervised fine-tuning to emit structured criterion-resolved CPTS scores from $(q,A,a^*,r)$. \\
ClinRM & Optimizer settings & See the ClinMPO repository & The complete optimizer and run configuration is available in the ClinMPO repository (\clinmporepo). \\
Cold-start SFT & Policy backbones & Qwen3-0.6B, 1.7B, 4B and 8B & Each backbone was adapted on the 7,112-item higher-solvability training set from the Public QA Dataset and used to initialize the corresponding RL policy. \\
Cold-start SFT & Adaptation method & LoRA & The complete adaptation configuration, including target modules, rank, alpha, dropout, trainable-parameter count and bias setting, is available in the ClinMPO repository (\clinmporepo). \\
Cold-start SFT & Optimization settings & See the ClinMPO repository & The complete optimizer and run configuration is available in the ClinMPO repository (\clinmporepo). \\
GRPO & Training data and initialization & Predefined higher-solvability training set from the Public QA Dataset; initialized from the size-matched cold-start SFT policy & Candidate responses were rewarded and standardized within each prompt group before optimization. \\
GRPO & Objective & Clipped policy-ratio objective with KL regularization & The fixed reference policy was initialized from the corresponding SFT model; complete objective and optimization settings are available in the ClinMPO repository (\clinmporepo). \\
ClinMPO & Reward & $R_{\mathrm{CPTS}}(s,a)$ & The operational reward definition and policy-optimization settings are available in the ClinMPO repository (\clinmporepo). \\
RL rollout & Generation settings & See the ClinMPO repository & The complete rollout and generation configuration is available in the ClinMPO repository (\clinmporepo). \\
Evaluation & Fixed in-domain test set & 1,737-item model-screened challenge test set from the Public QA Dataset & Base, SFT, GRPO and ClinMPO variants were evaluated on the same item set; exact final-option match defined correctness. \\
Training time & Wall-clock duration & Hardware- and bandwidth-dependent & Training time depends on available hardware, GPU utilization and effective inter-GPU and data-transfer bandwidth; the relevant system and run configuration is available in the ClinMPO repository (\clinmporepo). \\
\end{longtable}
\endgroup

\phantomsection
\addcontentsline{toc}{subsection}{Supplementary Table 5 | psychiatric practice competencies}
The study used two complementary classification schemes: ICD-11 diagnostic categories and psychiatric practice competencies. The former describe the diagnostic domain of a question, whereas the latter describe the reasoning task required to answer it. The schemes were applied independently and were not tiers of a single hierarchy.

\setcounter{table}{4}
\input{supplementary-information/sections/supplementary-tables-categories}

\clearpage
\phantomsection
\addcontentsline{toc}{subsection}{Supplementary Table 7 | Accuracy across all evaluated models}

\begingroup
\fontsize{8.5}{9}\selectfont
\singlespacing
\setlength{\tabcolsep}{2pt}
\renewcommand{\arraystretch}{1.00}
\begin{longtable}{@{}r p{0.292\textwidth}@{\hspace{1pt}}r@{\hspace{8pt}}r r@{}}
\caption{\textbf{Accuracy across the 33 evaluated entries: 31 open-source entries (models and post-training variants), one closed-source model and the human baseline.} Accuracy was calculated on all 1,737 test items.}\label{supp-tab-all-model-accuracy}\\
\toprule
\textbf{Rank} & \textbf{Model or reference} & \textbf{Parameters} & \textbf{Correct/total} & \textbf{Accuracy (\%)} \\
\midrule
\endfirsthead
\multicolumn{5}{l}{\tablename~\thetable\ continued}\\
\toprule
\textbf{Rank} & \textbf{Model or reference} & \textbf{Parameters} & \textbf{Correct/total} & \textbf{Accuracy (\%)} \\
\midrule
\endhead
\bottomrule
\endfoot
1 & GPT-4o & Not disclosed & 571/1,737 & 32.87 \\
2 & Qwen3-8B-ClinMPO & 8B & 562/1,737 & 32.35 \\
3 & Meditron3-Qwen2.5-14B\textsuperscript{\#} & 14B & 550/1,737 & 31.66 \\
4 & MedGemma-27B-IT\textsuperscript{\#} & 27B & 546/1,737 & 31.43 \\
5 & Baichuan-M2-32B & 32B & 539/1,737 & 31.03 \\
6 & Qwen3-8B-GRPO & 8B & 539/1,737 & 31.03 \\
7 & Human baseline & N/A & 535.85/1,737 & 30.85 \\
8 & Llama-3.1-MedPRM-Reward\textsuperscript{\#} & 8B & 533/1,737 & 30.69 \\
9 & Qwen3-8B-SFT & 8B & 533/1,737 & 30.69 \\
10 & Qwen3-4B-ClinMPO & 4B & 528/1,737 & 30.40 \\
11 & GLM-4-9B-Chat\textsuperscript{\(\dagger,\#\)} & 9B & 509/1,737 & 29.30 \\
12 & Qwen3-4B-GRPO & 4B & 497/1,737 & 28.61 \\
13 & Gemma-3-27B-MeditronFO & 27B & 495/1,737 & 28.50 \\
14 & Qwen3-8B & 8B & 491/1,737 & 28.27 \\
15 & Qwen3-4B-SFT & 4B & 488/1,737 & 28.09 \\
16 & MedAlpaca-13B & 13B & 484/1,737 & 27.86 \\
17 & Llama-3.1-8B & 8B & 483/1,737 & 27.81 \\
18 & Qwen3-1.7B-ClinMPO & 1.7B & 466/1,737 & 26.83 \\
19 & Llama3-OpenBioLLM-8B\textsuperscript{\(\dagger,\#\)} & 8B & 465/1,737 & 26.77 \\
20 & Qwen3-1.7B-SFT & 1.7B & 455/1,737 & 26.19 \\
21 & Qwen3-1.7B-GRPO & 1.7B & 449/1,737 & 25.85 \\
22 & Llama-3-8B-UltraMedical\textsuperscript{\(\dagger,\#\)} & 8B & 433/1,737 & 24.93 \\
23 & Qwen3-1.7B & 1.7B & 431/1,737 & 24.81 \\
24 & Qwen3-4B & 4B & 430/1,737 & 24.76 \\
25 & Qwen3-0.6B-ClinMPO & 0.6B & 422/1,737 & 24.29 \\
26 & Gemma-2-9B-IT & 9B & 419/1,737 & 24.12 \\
27 & Qwen3-0.6B & 0.6B & 418/1,737 & 24.06 \\
28 & Qwen3-0.6B-GRPO & 0.6B & 414/1,737 & 23.83 \\
29 & Qwen3-0.6B-SFT & 0.6B & 408/1,737 & 23.49 \\
30 & Yi-1.5-9B\textsuperscript{\(\dagger\)} & 9B & 404/1,737 & 23.26 \\
31 & Baichuan2-7B-Chat & 7B & 382/1,737 & 21.99 \\
32 & BioMistral-7B\textsuperscript{\(\dagger\)} & 7B & 361/1,737 & 20.78 \\
33 & Mistral-7B & 7B & 189/1,737 & 10.88 \\
\end{longtable}
\vspace{-0.4em}
\textsuperscript{\(\dagger\)} Output-format errors were corrected by predefined post-processing.\par
\textsuperscript{\#} Possible training-data overlap; accuracy should be interpreted descriptively.
\endgroup

\clearpage
\phantomsection
\addcontentsline{toc}{subsection}{Supplementary Table 8 | Categories excluded from category-level plots}

\begingroup
\small
\singlespacing
\setlength{\tabcolsep}{4pt}
\renewcommand{\arraystretch}{1.08}
\begin{longtable}{@{}>{\raggedright\arraybackslash}p{0.15\textwidth}>{\centering\arraybackslash}p{0.05\textwidth}>{\raggedright\arraybackslash}p{0.51\textwidth}r >{\raggedright\arraybackslash}p{0.11\textwidth}@{}}
\caption{\textbf{Categories excluded from the category-level radar plots.} The 14 categories listed here were not plotted because they contained fewer than 30 test questions. Sample size denotes the number of questions assigned to the category before application of the prespecified minimum-size criterion. Category codes correspond to Supplementary Tables 5 and 6.}\label{supp-tab-unplotted-categories}\\
\toprule
\textbf{Dimension} & \textbf{Code} & \textbf{Category} & \shortstack{\textbf{Sample}\\\textbf{size}} & \textbf{Exclusion reason} \\
\midrule
\endfirsthead
\multicolumn{5}{l}{\tablename~\thetable\ continued}\\
\toprule
\textbf{Dimension} & \textbf{Code} & \textbf{Category} & \shortstack{\textbf{Sample}\\\textbf{size}} & \textbf{Exclusion reason} \\
\midrule
\endhead
\bottomrule
\endfoot
ICD-11 & A & Other specified mental, behavioural or neurodevelopmental disorders [6E8Y] & 20 & \(n < 30\) \\
ICD-11 & B & Gender incongruence & 4 & \(n < 30\) \\
ICD-11 & C & Catatonia & 26 & \(n < 30\) \\
ICD-11 & E & Mental, behavioural or neurodevelopmental disorders, unspecified [6E8Z] & 10 & \(n < 30\) \\
ICD-11 & I & Paraphilic disorders & 16 & \(n < 30\) \\
ICD-11 & K & Impulse control disorders & 6 & \(n < 30\) \\
ICD-11 & L & Dissociative disorders & 26 & \(n < 30\) \\
ICD-11 & P & Sleep-wake disorders [07] & 10 & \(n < 30\) \\
ICD-11 & S & Secondary mental or behavioural syndromes associated with disorders or diseases classified elsewhere & 21 & \(n < 30\) \\
ICD-11 & W & Disorders of bodily distress or bodily experience & 28 & \(n < 30\) \\
ICD-11 & X & Factitious disorders & 7 & \(n < 30\) \\
ICD-11 & Y & Elimination disorders & 3 & \(n < 30\) \\
ICD-11 & Z & Mental or behavioural disorders associated with pregnancy, childbirth or the puerperium & 1 & \(n < 30\) \\
psychiatric practice competency & I & Monitoring, follow-up and measurement-based care & 21 & \(n < 30\) \\
\end{longtable}
\endgroup

\clearpage
\phantomsection
\addcontentsline{toc}{subsection}{Supplementary Table 9 | Model accuracy and medical-student mean correct-response proportions}

\begingroup
\fontsize{6.8}{7.4}\selectfont
\singlespacing
\setlength{\tabcolsep}{1.5pt}
\renewcommand{\arraystretch}{0.90}
\begin{center}
\resizebox{0.95\textwidth}{!}{\rotatebox{90}{\begin{minipage}{0.92\textheight}
\captionsetup{font=footnotesize}
\captionof{table}{\textbf{Model accuracy and medical-student mean correct-response proportions in the 14 categories excluded from the category-level radar plots.} Values are percentages of all test questions assigned to each category. ICD-11 diagnostic category codes and the psychiatric practice competency I code correspond to the full category names and sample sizes in Supplementary Table 8. Missing or unparseable model predictions were counted as incorrect; the medical-student mean correct-response proportions are weighted estimates. Because each category contained fewer than 30 questions, the estimates are reported descriptively rather than used for between-category inference.}\label{supp-tab-unplotted-category-accuracy}
\begin{tabular}{@{}>{\raggedright\arraybackslash}p{0.25\linewidth}*{14}{>{\centering\arraybackslash}p{0.047\linewidth}}@{}}
\toprule
& \multicolumn{13}{c}{\textbf{ICD-11 diagnostic category}} & \textbf{psychiatric practice competency} \\
\cmidrule(lr){2-14}\cmidrule(l){15-15}
\textbf{Model} & \textbf{A} & \textbf{B} & \textbf{C} & \textbf{E} & \textbf{I} & \textbf{K} & \textbf{L} & \textbf{P} & \textbf{S} & \textbf{W} & \textbf{X} & \textbf{Y} & \textbf{Z} & \textbf{I} \\
\midrule
GPT-4o & 20.0 & 25.0 & 19.2 & 10.0 & 18.8 & 16.7 & 19.2 & 30.0 & 38.1 & 25.0 & 42.9 & 100.0 & 0.0 & 28.6 \\
Qwen3-8B-ClinMPO & 30.0 & 50.0 & 26.9 & 20.0 & 31.2 & 50.0 & 11.5 & 20.0 & 38.1 & 32.1 & 42.9 & 100.0 & 100.0 & 42.9 \\
Meditron3-Qwen2.5-14B & 55.0 & 0.0 & 30.8 & 20.0 & 25.0 & 16.7 & 23.1 & 50.0 & 38.1 & 32.1 & 28.6 & 33.3 & 0.0 & 9.5 \\
MedGemma-27B-IT & 10.0 & 0.0 & 26.9 & 30.0 & 43.8 & 16.7 & 30.8 & 30.0 & 38.1 & 50.0 & 57.1 & 33.3 & 0.0 & 14.3 \\
Baichuan-M2-32B & 35.0 & 0.0 & 26.9 & 40.0 & 31.2 & 16.7 & 23.1 & 50.0 & 38.1 & 39.3 & 0.0 & 66.7 & 0.0 & 28.6 \\
Qwen3-8B-GRPO & 25.0 & 0.0 & 34.6 & 30.0 & 25.0 & 16.7 & 46.2 & 60.0 & 38.1 & 42.9 & 28.6 & 33.3 & 0.0 & 38.1 \\
Human baseline & 37.7 & 28.9 & 28.1 & 27.0 & 32.2 & 20.3 & 36.7 & 29.2 & 31.4 & 31.6 & 25.5 & 63.0 & 100.0 & 29.9 \\
Llama-3.1-MedPRM-Reward & 25.0 & 25.0 & 34.6 & 30.0 & 12.5 & 16.7 & 15.4 & 20.0 & 19.0 & 35.7 & 42.9 & 66.7 & 0.0 & 28.6 \\
Qwen3-8B-SFT & 15.0 & 50.0 & 19.2 & 20.0 & 25.0 & 16.7 & 26.9 & 40.0 & 23.8 & 39.3 & 28.6 & 66.7 & 100.0 & 14.3 \\
Qwen3-4B-ClinMPO & 30.0 & 25.0 & 34.6 & 50.0 & 37.5 & 0.0 & 30.8 & 10.0 & 23.8 & 28.6 & 28.6 & 33.3 & 100.0 & 19.0 \\
GLM-4-9B-Chat & 35.0 & 50.0 & 26.9 & 10.0 & 18.8 & 0.0 & 23.1 & 20.0 & 19.0 & 42.9 & 42.9 & 66.7 & 0.0 & 14.3 \\
Qwen3-4B-GRPO & 20.0 & 50.0 & 30.8 & 40.0 & 31.2 & 33.3 & 34.6 & 40.0 & 38.1 & 32.1 & 28.6 & 66.7 & 0.0 & 28.6 \\
Gemma-3-27B-MeditronFO & 20.0 & 50.0 & 26.9 & 10.0 & 43.8 & 16.7 & 23.1 & 40.0 & 19.0 & 35.7 & 57.1 & 33.3 & 0.0 & 19.0 \\
Qwen3-8B & 25.0 & 0.0 & 23.1 & 10.0 & 12.5 & 16.7 & 34.6 & 50.0 & 28.6 & 42.9 & 14.3 & 33.3 & 100.0 & 23.8 \\
Qwen3-4B-SFT & 20.0 & 50.0 & 19.2 & 30.0 & 18.8 & 0.0 & 26.9 & 60.0 & 28.6 & 39.3 & 28.6 & 33.3 & 100.0 & 23.8 \\
MedAlpaca-13B & 25.0 & 25.0 & 19.2 & 20.0 & 18.8 & 33.3 & 42.3 & 20.0 & 23.8 & 28.6 & 28.6 & 66.7 & 0.0 & 28.6 \\
Llama-3.1-8B & 40.0 & 50.0 & 34.6 & 20.0 & 31.2 & 16.7 & 46.2 & 0.0 & 23.8 & 39.3 & 42.9 & 66.7 & 0.0 & 19.0 \\
Qwen3-1.7B-ClinMPO & 35.0 & 0.0 & 19.2 & 30.0 & 37.5 & 16.7 & 15.4 & 10.0 & 38.1 & 10.7 & 14.3 & 33.3 & 100.0 & 28.6 \\
Llama3-OpenBioLLM-8B & 10.0 & 25.0 & 15.4 & 20.0 & 31.2 & 0.0 & 38.5 & 30.0 & 28.6 & 28.6 & 28.6 & 33.3 & 0.0 & 23.8 \\
Qwen3-1.7B-SFT & 20.0 & 50.0 & 26.9 & 20.0 & 18.8 & 33.3 & 26.9 & 10.0 & 23.8 & 32.1 & 57.1 & 66.7 & 100.0 & 14.3 \\
Qwen3-1.7B-GRPO & 15.0 & 50.0 & 34.6 & 30.0 & 25.0 & 0.0 & 34.6 & 20.0 & 33.3 & 25.0 & 57.1 & 66.7 & 100.0 & 23.8 \\
Llama-3-8B-UltraMedical & 20.0 & 25.0 & 23.1 & 20.0 & 31.2 & 0.0 & 34.6 & 30.0 & 23.8 & 35.7 & 14.3 & 0.0 & 100.0 & 33.3 \\
Qwen3-1.7B & 20.0 & 0.0 & 23.1 & 20.0 & 37.5 & 66.7 & 30.8 & 30.0 & 28.6 & 32.1 & 28.6 & 100.0 & 100.0 & 28.6 \\
Qwen3-4B & 10.0 & 0.0 & 23.1 & 10.0 & 31.2 & 33.3 & 11.5 & 40.0 & 33.3 & 35.7 & 0.0 & 33.3 & 0.0 & 4.8 \\
Qwen3-0.6B-ClinMPO & 35.0 & 0.0 & 11.5 & 30.0 & 25.0 & 16.7 & 7.7 & 50.0 & 33.3 & 28.6 & 57.1 & 0.0 & 0.0 & 14.3 \\
Gemma-2-9B-IT & 5.0 & 25.0 & 19.2 & 30.0 & 25.0 & 16.7 & 26.9 & 10.0 & 33.3 & 35.7 & 28.6 & 66.7 & 0.0 & 19.0 \\
Qwen3-0.6B & 10.0 & 0.0 & 11.5 & 30.0 & 12.5 & 50.0 & 23.1 & 0.0 & 9.5 & 25.0 & 28.6 & 66.7 & 0.0 & 4.8 \\
Qwen3-0.6B-GRPO & 15.0 & 0.0 & 11.5 & 10.0 & 18.8 & 16.7 & 30.8 & 10.0 & 19.0 & 25.0 & 57.1 & 0.0 & 0.0 & 19.0 \\
Qwen3-0.6B-SFT & 10.0 & 25.0 & 11.5 & 20.0 & 18.8 & 66.7 & 30.8 & 0.0 & 23.8 & 39.3 & 42.9 & 66.7 & 0.0 & 23.8 \\
Yi-1.5-9B & 15.0 & 25.0 & 23.1 & 40.0 & 31.2 & 16.7 & 26.9 & 0.0 & 19.0 & 14.3 & 28.6 & 0.0 & 0.0 & 14.3 \\
Baichuan2-7B-Chat & 10.0 & 25.0 & 15.4 & 20.0 & 25.0 & 0.0 & 26.9 & 10.0 & 28.6 & 32.1 & 57.1 & 33.3 & 0.0 & 23.8 \\
BioMistral-7B & 25.0 & 25.0 & 23.1 & 10.0 & 31.2 & 50.0 & 26.9 & 10.0 & 14.3 & 21.4 & 28.6 & 0.0 & 0.0 & 19.0 \\
Mistral-7B & 0.0 & 0.0 & 15.4 & 0.0 & 0.0 & 0.0 & 15.4 & 0.0 & 0.0 & 10.7 & 14.3 & 33.3 & 0.0 & 9.5 \\
\bottomrule
\end{tabular}
\end{minipage}}}
\end{center}
\endgroup

\clearpage
\phantomsection
\addcontentsline{toc}{subsection}{Supplementary Table 10 | Pre-split evaluation of 13 models}

\begingroup
\small
\singlespacing
\setlength{\tabcolsep}{5pt}
\renewcommand{\arraystretch}{1.10}
\begin{longtable}{@{}>{\raggedright\arraybackslash}p{0.39\textwidth}rrrr@{}}
\caption{\textbf{Evaluation of 13 models before the retained questions were divided into training and test sets.} All models were evaluated on the same 8,849 retained questions. Questions answered correctly by at least 9 models entered the 7,112-item model-screened higher-solvability training set; the remaining 1,737 questions formed the fixed, difficulty-enriched test set. Missing model predictions were treated as incorrect for partitioning. Predictions denotes the number of questions for which a model prediction was available; coverage was calculated as predictions divided by 8,849. Accuracy was calculated conditionally as correct predictions divided by available predictions and should therefore be interpreted together with coverage.}\label{supp-tab-pre-split-model-accuracy}\\
\toprule
\textbf{Model identifier} & \textbf{Predictions} & \textbf{Correct} & \shortstack{\textbf{Coverage}\\\textbf{(\%)}} & \shortstack{\textbf{Accuracy}\\\textbf{(\%)}} \\
\midrule
\endfirsthead
\multicolumn{5}{l}{\tablename~\thetable\ continued}\\
\toprule
\textbf{Model identifier} & \textbf{Predictions} & \textbf{Correct} & \shortstack{\textbf{Coverage}\\\textbf{(\%)}} & \shortstack{\textbf{Accuracy}\\\textbf{(\%)}} \\
\midrule
\endhead
\bottomrule
\endfoot
claude-sonnet-4-20250514 & 8,849 & 7,707 & 100.00 & 87.09 \\
claude-sonnet-4-20250514-thinking & 8,849 & 7,728 & 100.00 & 87.33 \\
deepseek-r1 & 8,848 & 7,834 & 99.99 & 88.54 \\
deepseek-v3 & 8,847 & 7,708 & 99.98 & 87.13 \\
gemini-2.5-pro & 8,474 & 7,578 & 95.76 & 89.43 \\
gpt-4o & 8,848 & 7,428 & 99.99 & 83.95 \\
gpt-5 & 8,849 & 7,919 & 100.00 & 89.49 \\
gpt-5-mini & 8,849 & 7,701 & 100.00 & 87.03 \\
grok-4 & 2,967 & 2,752 & 33.53 & 92.75 \\
llama-3.1-8b & 8,848 & 6,083 & 99.99 & 68.75 \\
mistral-7b & 5,576 & 3,251 & 63.01 & 58.30 \\
qwen3-8b & 8,558 & 7,011 & 96.71 & 81.92 \\
yi-1.5-9b & 8,848 & 6,051 & 99.99 & 68.39 \\
\end{longtable}
\endgroup

%% file: supplementary-information/sections/supplementary-tables-categories.tex
\begingroup
\small
\singlespacing
\renewcommand{\arraystretch}{1.12}

\begin{longtable}{@{}p{0.10\textwidth}p{0.84\textwidth}@{}}
\caption{\textbf{Psychiatric practice competencies used in the study.} The 12 competencies operationalize clinically relevant reasoning tasks for category-level evaluation.}\label{supp-tab-competencies}\\
\toprule
\textbf{Code} & \textbf{Psychiatric practice competency} \\
\midrule
\endfirsthead
\multicolumn{2}{l}{\tablename~\thetable\ continued}\\
\toprule
\textbf{Code} & \textbf{Psychiatric practice competency} \\
\midrule
\endhead
\bottomrule
\endfoot
I & Monitoring, follow-up and measurement-based care \\
II & Communication, consultation and interprofessional collaboration \\
III & Risk assessment and safety planning \\
IV & Differential diagnosis, including medical mimics \\
V & Systems-based and community care from a public-health perspective \\
VI & Treatment selection and initiation \\
VII & Symptom elicitation and mental status examination \\
VIII & Legal, ethical, capacity and judgment assessment \\
IX & Comorbidity and complexity management \\
X & Special populations and lifespan adaptation \\
XI & Diagnostic formulation and case definition \\
XII & Evidence appraisal, data interpretation and scale use \\
\end{longtable}

\phantomsection
\addcontentsline{toc}{subsection}{Supplementary Table 6 | ICD-11 diagnostic categories}
The following table lists the 26 ICD-11 diagnostic categories used in the study. These categories support domain-level analysis and do not replace the complete ICD-11 hierarchy.

\begin{longtable}{@{}p{0.10\textwidth}p{0.84\textwidth}@{}}
\caption{\textbf{ICD-11 diagnostic categories used in the study.} These operational categories were used to group questions for analysis and do not replace the full ICD-11 hierarchy.}\label{supp-tab-icd11}\\
\toprule
\textbf{Code} & \textbf{Diagnostic category} \\
\midrule
\endfirsthead
\multicolumn{2}{l}{\tablename~\thetable\ continued}\\
\toprule
\textbf{Code} & \textbf{Diagnostic category} \\
\midrule
\endhead
\bottomrule
\endfoot
A & Other specified mental, behavioural or neurodevelopmental disorders [6E8Y] \\
B & Gender incongruence \\
C & Catatonia \\
D & Feeding or eating disorders \\
E & Mental, behavioural or neurodevelopmental disorders, unspecified [6E8Z] \\
F & Mood disorders \\
G & Disruptive behaviour or dissocial disorders \\
H & Disorders specifically associated with stress \\
I & Paraphilic disorders \\
J & Neurodevelopmental disorders \\
K & Impulse control disorders \\
L & Dissociative disorders \\
M & Personality disorders and related traits \\
N & Disorders due to substance use or addictive behaviours \\
O & Anxiety or fear-related disorders \\
P & Sleep-wake disorders [07] \\
Q & Obsessive-compulsive or related disorders \\
R & Psychological or behavioural factors affecting disorders or diseases classified elsewhere [6E40] \\
S & Secondary mental or behavioural syndromes associated with disorders or diseases classified elsewhere \\
T & Neurocognitive disorders \\
U & Schizophrenia or other primary psychotic disorders \\
V & Sexual dysfunctions \\
W & Disorders of bodily distress or bodily experience \\
X & Factitious disorders \\
Y & Elimination disorders \\
Z & Mental or behavioural disorders associated with pregnancy, childbirth or the puerperium \\
\end{longtable}

\endgroup